\pdfoutput=1

\documentclass{docclass_edit}

\usepackage{jmlr2e_style_edit}
\usepackage{bm}
\usepackage{rotating,tabularx,array,multirow,graphicx,graphics,mdwlist}
\usepackage{amsmath}
\usepackage{times}
\usepackage{natbib,natbibspacing}
\usepackage{subfigure}
\usepackage{paralist}
\usepackage[small,it]{caption}
\newcommand{\uw}{\mbox{$\underline{w}$}}

\begin{document}

\title{Statistical Topic Models for Multi-Label Document Classification}

\author{\name Timothy Rubin \email trubin@uci.edu \\
       \addr Department of Cognitive Sciences\\
       University of California, Irvine\\
       Irvine, CA 92697, USA
       \AND
       \name America Chambers  \email ahollowa@uci.edu \\
       \addr Department of Computer Science\\
       University of California\\
       Irvine, CA 92697, USA
              \AND
       \name Padhraic Smyth \email smyth@ics.uci.edu \\
       \addr Department of Computer Science\\
       University of California\\
       Irvine, CA 92697, USA
              \AND
       \name Mark Steyvers\email mark.steyvers@uci.edu \\
       \addr Department of Cognitive Sciences\\
       University of California, Irvine\\
       Irvine, CA 92697, USA
       }

\maketitle

\vspace{-6mm}

\begin{abstract}%   <- trailing '%' for backward compatibility of .sty file
Machine learning approaches to multi-label document classification have to date largely relied
on discriminative modeling techniques such as support vector machines. A drawback of these
approaches is that performance rapidly drops off as the total number of labels  and the number of
labels per document increase. This problem is amplified when the label frequencies exhibit the type of highly skewed distributions that are often observed in real-world datasets.  In this paper we
investigate a class of generative statistical topic models for multi-label documents that associate individual word tokens  with different labels. We investigate the advantages of this approach
relative to discriminative models, particularly with respect to classification problems involving
large numbers of relatively rare labels. We compare the performance of generative and
discriminative approaches on document labeling tasks ranging from datasets with several thousand
labels to datasets with tens of labels. The experimental results indicate that probabilistic generative models can achieve competitive multi-label classification performance compared to discriminative methods, and have advantages for datasets with many labels and skewed label frequencies.
\end{abstract}

\vspace{8pt}

\section{Introduction}

The past decade has seen a wide variety of papers published on multi-label document classification, in which each document can be assigned to one or more classes.
In this introductory section we begin by discussing the limitations of existing multi-label document classification methods when applied to datasets with statistical properties common to real-world datasets, such as the presence of large numbers of labels with power-law-like frequency statistics. We then motivate the use of generative probabilistic models in this context.
In particular, we illustrate how these models can be advantageous in the context of large-scale multi-label corpora, through (1) explicitly assigning individual words to specific labels within each document---rather than assuming that all of the words within a document are relevant to each of its labels, and (2) jointly modeling all labels within a corpus simultaneously, which lends itself well to the task of accounting for the dependencies between these labels.

\subsection{Background and Motivation}

Much of the prior work on multi-label document classification uses data sets
in which there are relatively few labels, and many training instances for each label.  In many
cases, the datasets are constructed such that they contain few, if any, infrequent labels. For
example, in the commonly used RCV1-v2 corpus \citep{Lewis_Etc_2004}, the dataset was carefully
constructed to have approximately 100 labels, with most labels occurring in a relatively large
number of documents.

\begin{figure}[t]
  \centering
  \includegraphics[width=0.75\textwidth]{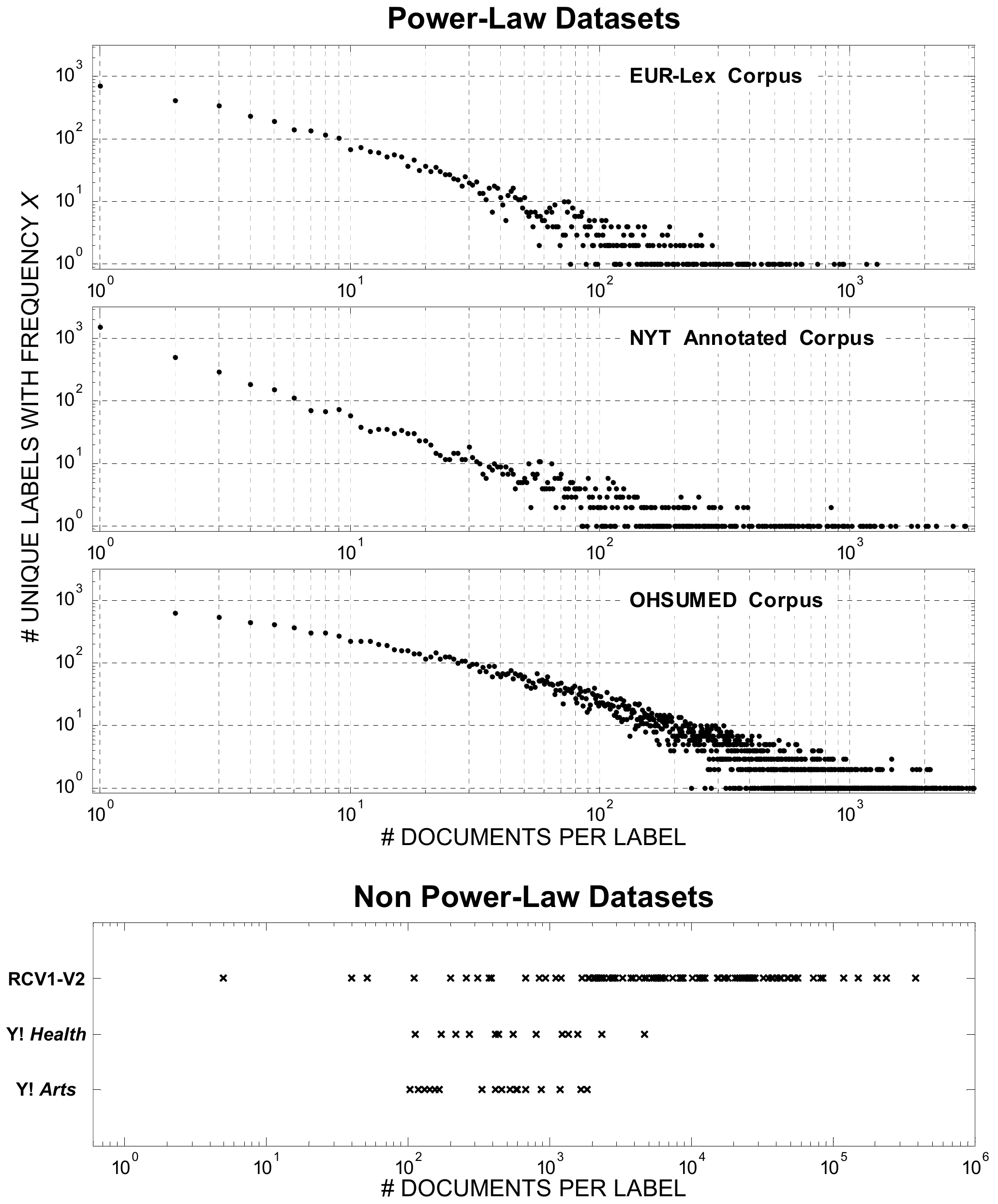}
  \caption{
  {\bf Top}: The number of unique labels (y-axis) that have  $K$ training documents
(x-axis) for three large-scale multi-label datasets. Both axes are shown on a log-scale. The
power-law-like relationship is evident from the near linear trend (in log-space) of this relationship.
{\bf Bottom}: The number of training documents(x-axis) for each unique label in three common (non-power-law) benchmark datasets. Since there are no label-frequencies at which there are more than one unique label in any of the datasets, if these plots were shown using the log-log scale used in the plots above, all points would fall along the y value corresponding to $10^{0}$.  Note that the scaling of the x-axis is not equivalent for the power-law and non power-law plots (this is necessary due to the high upper-bound of label-frequencies on the RCV1-V2 dataset).}
\label{fig:DocsPerLabel}
\end{figure}

In other cases researchers have typically restricted the problem by only considering a subset of
the full dataset. As an example, a popular source of experimental data has been the Yahoo!
directory structure, which utilizes a multi-labeling classification system. The true Yahoo!
directory structure contains thousands of labels and is a very difficult classification problem
that traditional classification methods fail to adequately handle
\citep{LiuYangWanZengChenMa_2005}.   However, the majority of multi-label research conducted using the Yahoo! directory data has been performed on the set of 11 sub-directory datasets constructed by \cite{UedaSaito_2002}. Each of these datasets consists of only the second-level categories from a
single top-level Yahoo! directory, leaving only about 20-30 labels in each of the classification
tasks.  Furthermore, many of the publications \citep[e.g.,][]{UedaSaito_2002,JiTangYuYe_2008} that
use the Yahoo! subdirectory datasets have removed the infrequent labels from the evaluation data,
leaving between 14 and 23 unique labels per dataset.  Similarly, experiments with the
OHSUMED MeSH terms \citep{Hersh_1994_OHSUMED}  are typically performed on a small subdirectory that contains only 119 out of over 22,000 possible labels (for a discussion,
see \citet{Rak_Etc_2005}).

In contrast to the datasets typically utilized in research, multilabel corpora in the real world can contain thousands or tens of thousands of labels, and the label frequencies in
these datasets tend to have highly skewed frequency-distributions with power-law
statistics \citep{Yang_Etc_2003_Scalability,LiuYangWanZengChenMa_2005,Dekel_Shamir_2010_AISTATS}. Figure~\ref{fig:DocsPerLabel} illustrates
this point for three large real-world corpora---each containing thousands of unique labels---by plotting the number of labels within each corpus as a function of label-frequency.  For each corpus, the total number of labels is plotted as a function of label-frequency on a log-log scale (i.e., more precisely, number of unique labels [y-axis] that have been assigned to $k$ documents in the corpus is plotted as a function of $k$ [x-axis]).  Of note is the power-law like distribution of label frequencies for each corpus, in which the vast majority of labels are associated with very few documents, and there are relatively few labels that are assigned to a large number of documents.  For example, roughly one thousand labels are only assigned to a single document in each corpus, and the median label-frequencies are 3, 6, and 12 for the NYT, EUR-Lex, and OHSUMED datasets, respectively.  This stands in stark contrast to the widely-used Yahoo! {\it Arts}, Yahoo! {\it Health}  and RCV1-v2 datasets  (for example), which are shown at the bottom of Figure~\ref{fig:DocsPerLabel}.  In these corpora, there are hardly any labels that occur in fewer than 100 documents, and the median label-frequencies are 530, 500, and 7,410 respectively  (see Section~\ref{sec:datasets} for further details and discussion).  To summarize, these popular benchmark datasets are drastically different from large-scale real-world corpora not only in terms of the number of unique labels they contain, but also with respect to the distribution of label-frequencies, and in particular the number of rare labels.

The mismatch between real-world and experimental datasets has been
discussed previously in the literature, notably by~\cite{LiuYangWanZengChenMa_2005}
who observed that although popular multi-label techniques---such as
``one-vs-all'' binary classification \citep[e.g.][]{Allwein_etc_2001_BinaryTransform,RifkinKlautau_2004}---can perform well on datasets with relatively few labels, performance drops off dramatically on real world datasets that contain many labels and skewed label frequency distributions. In addition,~\cite{Yang_2001} illustrated that discriminative methods which achieve good
performance on standard datasets do relatively poorly on larger datasets such as the full OHSUMED dataset. The obvious reason for this is that discriminative binary classifiers have difficulty
learning models for labels with very few positively labeled documents. As stated by
\cite{LiuYangWanZengChenMa_2005}, in the context of support vector machine (SVM) classifiers:

 \begin{quote}
In terms of effectiveness, neither flat nor hierarchical SVMs can
fulfill the needs of classification of very large-scale
taxonomies. The skewed distribution of the Yahoo! Directory and
other large taxonomies with many extremely rare categories makes
the classification performance of SVMs unacceptable. More
substantial investigation is thus needed to improve SVMs and other
statistical methods for very large-scale applications.  \end{quote}

\begin{figure}[h]
  \centering
  \includegraphics[width=0.75\textwidth]{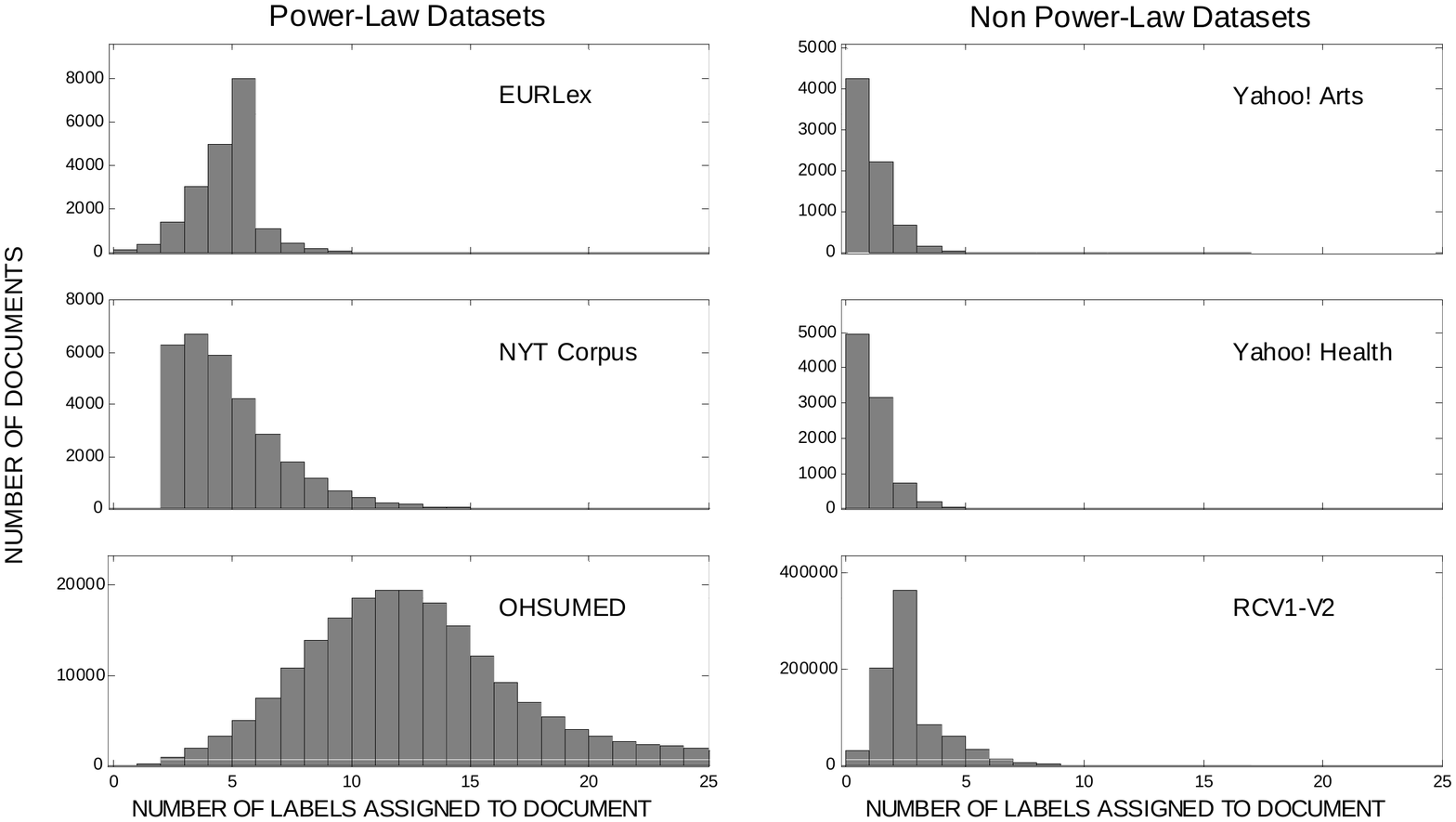}
  \caption{Number of documents (y-axis) that have  $L$ labels (x-axis).
  The version of the NYT Annotated Corpus used in our experiments contains documents
   with 3 or more labels, hence the cutoff at 3.}
  \label{fig:LabelsPerDoc}
\end{figure}

A second critical difference between large scale multi-label corpora and traditional benchmark datasets relates to the number of labels that are assigned to each document.
Figure \ref{fig:LabelsPerDoc} compares the distributions of the number of labels per document for the same corpora shown in Figure \ref{fig:DocsPerLabel}.  The median number of labels per document for the real world, power-law style datasets are 6, 5, and 12 for EUR-Lex, NYT and OHSUMED, respectively. These numbers are  significantly larger than those in the typical datasets used in multi-label classification experiments. For example, among the three benchmark datasets shown, the RCV1-v2 dataset has a median of 3 labels per document, and the Yahoo! {\it Arts} and {\it Health} datasets each have a median of only 1 label per document.  These differences can significantly impact the performance of a classifier.

As the number of labels per document increases, it becomes more difficult for a discriminative
algorithm to distinguish which words are discriminative for a particular label.  This problem is
further compounded when there is little training data per label. For the purposes of illustration, consider the following extreme case: suppose that we are training a binary classifier for a label, $c_1$, that has only been assigned to one document, $d$.  Furthermore, assume that two additional labels, $c_2$ and $c_3$, have been assigned to document $d$, and that these labels occur in a relatively large number of documents. Since document $d$ is the only positive training example for label $c_1$, an independent binary classifier trained on $c_1$ will learn a discriminant function that emphasizes not only words from document $d$ that are relevant to label $c_1$, but also words that are relevant to labels $c_2$ and $c_3$, since the classifier has no way of ``knowing'' which words are relevant to these other labels.  In other words, when training an independent binary classifier for label $c_1$, each additional label that co-occurs with $c_1$ will introduce additional confounding features for the classifier, thereby reducing the quality of the classifier.

Note however that in the above example it should be relatively easy to learn which features are relevant to the labels $c_2$ and $c_3$, since these labels occur in a large number of documents.  Thus, we {\it should} be able to leverage this information to improve our classifier for $c_1$ by removing the features in $d$ which we know to be relevant to these confounding labels.  One possible approach to address this problem is to learn which individual word tokens within a
document are likely to be associated with each label. If we could then use this information to identify which words within $d$ are likely to be related to $c_2$ and $c_3$, we could ``explain away" these words, and then use the remaining words for the purposes of learning a model for $c_1$.  Note that for this purpose it is useful to (1) remove the assumption of label-wise independence,
and (2) learn the models for all of the labels simultaneously, since learning which words within a document are irrelevant to a particular label is a key part of learning which words are relevant to the label.

\subsection{A Generative Modeling Approach}

In a generative approach to document classification, one learns a model for the distribution of
the words given each label, i.e., a model for $P(\uw | c), 1 \le c \le C$, where $\uw$ is the set
of words in a document, and constructs a discriminant
function for the label via Bayes rule. %, i.e., $f_c(\uw) \propto P(\uw|c) P(c)$.
In standard supervised learning, with one label per document,
these $C$ distributions are typically learned independently. With
multi-label data, the distributions should instead be learned
simultaneously since we cannot separate the training data into $C$
groups by label.

A useful approach in this context is a model known as latent Dirichlet allocation (LDA) \citep{BleiNgJordan_2003}, which we will also refer to as topic modeling, which models the words in a document as being generated by a mixture of topics, i.e., $P(w| d) = \sum_c P(w | c) P(c | d)$, where $P(w | d)$ is the marginal probability of word $w$ in document $d$, $P(w | c)$ is the probability of word $w$ being generated given label $c$, and $P( c | d)$ is the relative probability of each of the $c$ labels associated with document $d$. LDA has primarily been viewed as an unsupervised learning algorithm, but can also be used in a supervised context \citep[e.g.,][]{Blei_Mcauliffe_07,Mimno_McCallum_2008,Ramage_Etc_2009}. Using a supervised version of LDA it is possible to learn both the word-label distributions $P(w|c)$ and the document-label weights $P(c | d)$ given a training corpus with multi-label data.

What is particularly relevant is that this approach (1) models the assignment of labels at the
word-level, rather than at the document level as in discriminative models, and (2) learns a model for all labels at the same time, rather than treating each label independently.  In particular, for the
document $d$ in our earlier example that was assigned the set of labels  $\{c_1, c_2, c_3\}$, the model can explain away words
that belong to labels $c_2$ and $c_3$---i.e., words that have high probability $P(w|c)$ under
these labels.  Since $c_2$ and $c_3$ are frequent labels, it will be relatively easy to learn which features are relevant to these labels, since the confounding features introduced by co-occurring labels in a multi-label scheme will tend to cancel out over many documents. The remaining words that cannot be explained well by $c_2$ or $c_3$  will be
assigned to label $c_1$, and the model will learn to associate such words with this label and not
associate with $c_1$ the words that are more likely to belong to labels $c_2$ and $c_3$. This
general intuition is the basis for our approach in this paper. Specifically, we investigate
supervised versions of topic models (LDA) as a general framework for multi-label document
classification. In particular, the topic modeling approach allows for the type of ``explaining away" effect at the word level that we hypothesize should be particularly helpful for the types of rare labels that pose challenges to purely discriminative methods.

\begin{figure}[ht] % float placement: (h)ere, page (t)op, page (b)ottom, other (p)age
  \centering
  \includegraphics[width=0.7\linewidth,keepaspectratio]{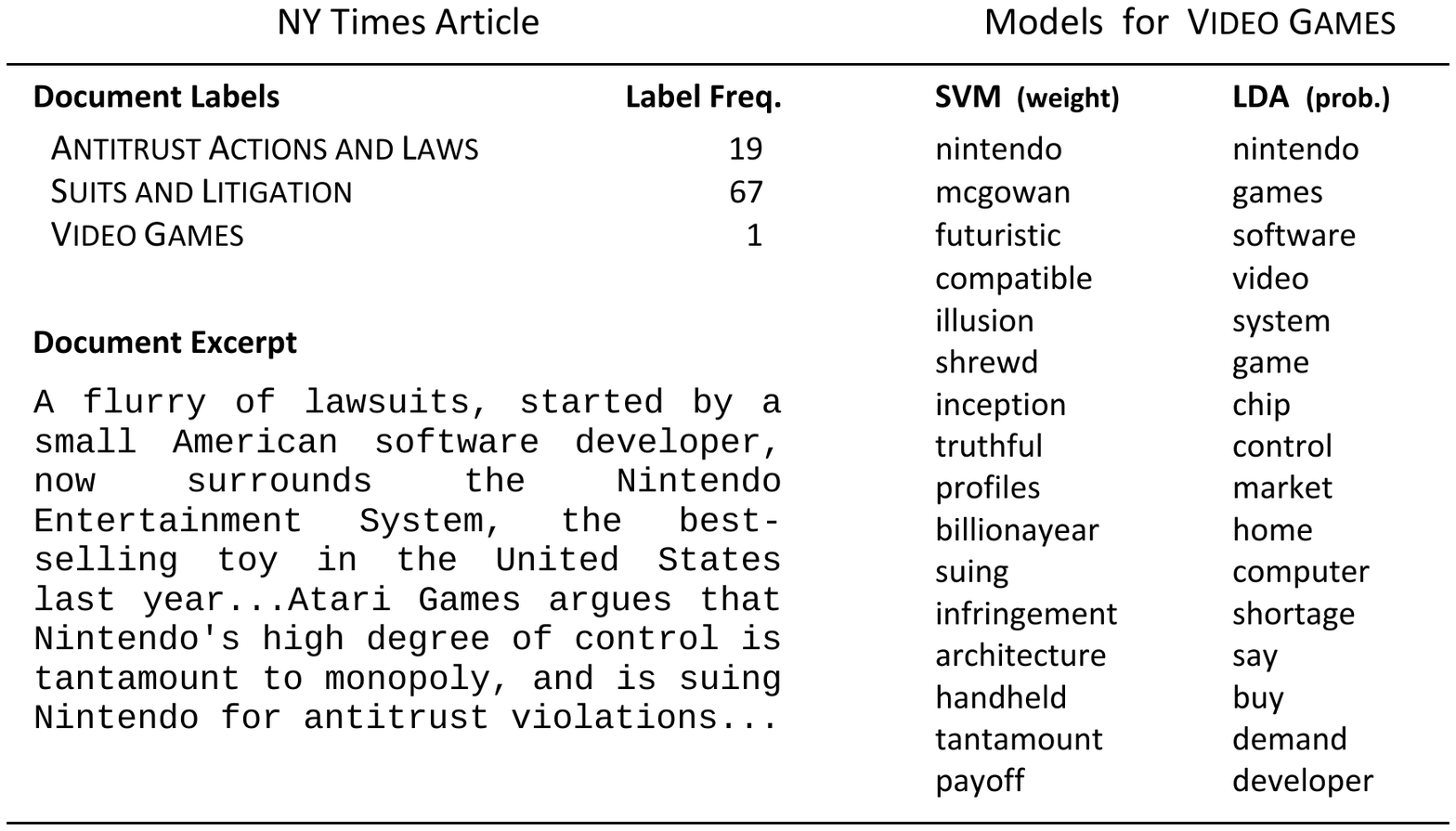}
  \caption{High-weight and high probability words for the label \textsc{video games} learned by an SVM classifier and an LDA model (respectively) from the a set of New York Times articles, in which the label \textsc{video games} only appeared once (text from the article is shown on the left).}
  \label{fig:videogamelabel}
\end{figure}

Figure \ref{fig:videogamelabel} illustrates the advantages an LDA-based approach has in terms of learning rare labels.
On the left is the partial
text of a news article, taken from the New York Times, along with three human-assigned labels:
{\sc antitrust actions and laws} and {\sc suits and litigation} (which both occur in multiple other
documents) and {\sc video games} (for which this document is the only positive example in
the training data). On the right are the words with the highest weights from a binary SVM
classifier trained on the label {\sc video games}. Beside this column are the highest probability
words learned by an LDA-based model (described in more detail later in the paper). The words learned by
the SVM classifier are quite noisy, containing a mixture of words relevant to the other two labels (e.g., \textit{suing}, \textit{infringement}, etc...), as well as rare words that are peculiar to the specific document rather than being relevant features for any of the labels (e.g., \textit{futuristic}, \textit{illusion}, etc...). These words do not match our intuition of words that would be discriminative for the concept {\sc video games}. Furthermore, as we will see later in the experimental results section,
SVM classifiers trained on rare labels in this type of  multi-label problem do not predict well on
new test documents. While the set of words learned by LDA model is still somewhat noisy, it is nonetheless clear the model has done a better job in determining which words are relevant to the label {\sc video games}, and which of the words should be associated with the other two labels (e.g., there are no words with high probability that directly relate to lawsuits).  The model benefits from not assuming independence between the labels, as with binary SVMs, as well as from the ``explaining away'' effect.

Thus far we have focused our discussion on the issue of learning appropriate models for labels during training.  An additional issue that arises as the number of total labels (as well as the number of labels per document) increases, is the importance of accounting for higher-order dependencies between labels at prediction time (i.e., when classifying a new document).
For example, suppose that we are predicting which labels should be assigned to a test-document that contains the word \textit{steroids}.
In a large-scale dataset like the NYT corpus, this word is a high-probability feature among many different labels, such as \textsc{Medicine and health}, \textsc{Baseball}, and  \textsc{Black markets}. The ambiguity in the assignment of this word to a specific label can often be resolved if we account for the other labels within the document; e.g.,  the word \textit{steroids} is likely to be related to the label \textsc{Baseball} given that the label \textsc{Suspensions, dismissals and resignations} is also assigned to the document, whereas it is more likely to be related to \textsc{Medicine and health} given the presence of the label \textsc{Cancer}.

Given this motivation, an additional beneficial feature of the topic model---and probabilistic methods in general--- is that it is relatively straightforward to model the label dependencies that are present in the training data (a feature that we will elaborate on later in the paper).  Modeling label dependencies is widely acknowledged to be important for accurate classification in multi-label problems, yet has been problematic in the past for datasets with large numbers of labels,  as summarized in \citet{Read_Etc_2009}:
\begin{quote}
The consensus view in the literature is that it is crucial to take
into account label correlations during the classification process
..... However  as the size of the multi-label datasets grows,
most methods struggle with the exponential growth in the number of
possible correlations. Consequently these methods are able to be
more accurate on small datasets, but are not as applicable to
larger datasets.
\end{quote}
\indent Thus, the ability of probabilistic models to account for label dependencies is a strong motivation for considering these types of approaches in large-scale multi-label classification settings.

\subsection{Contributions and Outline}

In the context of the discussion above, this paper investigates the application of statistical topic modeling  to the task of multi-label document classification, with an emphasis on corpora with large numbers of labels.  We consider a set of three models based on the LDA framework.  The first model, {\it Flat-LDA}, has been employed previously in various forms.  Additionally, we present two new models: {\it Prior-LDA}, which introduces a novel approach to account for variations in label frequencies, and {\it Dependency-LDA}, which extends this approach to account for the dependencies between the labels.  We compare these three topic models to two variants of a popular discriminative approach (one-vs-all binary SVMs) on five datasets with widely contrasting statistics.

We evaluate the performance of these models on a variety of predictions tasks.  Specifically, we consider (1) {\it document-based} rankings (rank all labels according to their relevance to a test document) and binary predictions (make a strict yes/no classification about each label for a given document), and (2) {\it label-based} rankings (rank all documents according to their relevance to a label) and binary predictions (make a strict yes/no classification about each document for a given label).

The specific contributions of this paper are as follows:
\begin{itemize*}
\item We describe two novel generative models for multi-label document classification, including one (Dependency-LDA) which significantly improves performance over simpler models by accounting for label dependencies, and is highly competitive with popular discriminative approaches on large-scale datasets.

\item We report extensive experimental results on two multi-label corpora with large numbers of labels as well as three smaller benchmark datasets, comparing the proposed generative models with discriminative SVMs. To our knowledge this is the first empirical study comparing generative and discriminative models on large-scale multi-label problems.

\item We demonstrate that LDA-based models---in particular the Dependency-LDA model---can be highly competitive with, or better than, SVMs on large-scale datasets with power-law like statistics.

\item For document-based predictions, we show that Dependency-LDA has a clear advantage over SVMs on large-scale datasets, and is competitive with SVMs on the smaller, benchmark datasets.

\item For label-based predictions, we demonstrate that Dependency-LDA generally outperforms SVMs on large-scale datasets.  We furthermore show that there is a clear performance advantage for the LDA-based methods on rare labels (e.g., labels with fewer than 10 training documents).

\end{itemize*}

\noindent The remainder of the paper is organized as follows. We begin by describing how standard unsupervised LDA can be adapted  to handle multi-labeled text documents, and describe our extensions that incorporate label frequencies and label dependencies.  We then describe how inference is performed with these models, both for learning the model from training data and for making predictions on new test documents.  An extensive set of experimental results are then presented on a wide range of prediction tasks on five multi-label corpora.   We conclude the paper with a discussion of the relative merits of the LDA-based approaches vs. SVM-based approaches, particularly in the context of both the dataset statistics and prediction tasks being considered.

\section{Related Work}

A number of approaches have been proposed for adapting the unsupervised LDA model to the case of supervised learning---such as the Supervised Topic Model
\citep{Blei_Mcauliffe_07}, Semi-LDA \citep{Wang_etc_2007_SemiLDA}, DiscLDA \citep{Julien_Etc_2008}, and MedLDA \citep{Zhu_etc_2009_MedLDA} ---however, these adaptations are designed for single label classification or regression, and are not directly applicable to multilabel classification.

A more recent approach
proposed by \citet{Ramage_Etc_2009}---Labeled-LDA (L-LDA)---was designed specifically for multi-label settings.
In L-LDA, the training of the LDA model is adapted to account for multi-labeled corpora
by putting ``topics'' in 1-1 correspondence with labels and then restricting the sampling
of topics for each document to the set of labels that were assigned to the document,
in a manner similar to the Author-Model described by \citet{Zvi_Etc_2004}
(where the set of authors for each document in the Author Model is now replaced by the set of labels in L-LDA). The primary focus of  \cite{Ramage_Etc_2009} was to illustrate
that L-LDA has certain qualitative advantages over discriminative
methods (e.g., the ability to label  individual words, as well as
providing interpretable snippets  for document summarization).
Their classification results indicate that under certain
conditions LDA-based models may be able to achieve competitive
performance with discriminative approaches such as SVMs.

Our work differs from that of \cite{Ramage_Etc_2009} in two significant aspects. Firstly, we
propose a more flexible set of LDA models for multi-label classification---including one model that takes into account prior label frequencies, and one that can additionally account for label dependencies---which lead to significant improvements in classification performance.  The L-LDA model can be viewed as a special case of these models.
Secondly, we conduct a much larger range and more systematic set of experiments,
including in particular datasets with large numbers of labels with skewed frequency-distributions, and show that generative
models do particularly well in this regime compared to discriminative methods.
In contrast, \cite{Ramage_Etc_2009} compared their L-LDA approach with
discriminative models only on relatively small datasets (primarily on the Yahoo! sub-directory datasets discussed in the introduction).

Our work (as well as the Author Model and L-LDA model) can be seen as building on earlier ideas from the literature in probabilistic modeling for multilabel classification. \citet{McCallum_1999}
and \citet{UedaSaito_2002} investigated mixture models similar to L-LDA, where
each document is composed of a number of word distributions associated with
document labels. These papers can be viewed as early forerunners of the more general LDA frameworks we propose in this paper.

More recently \cite{GhamrawiMcCallum_2005} demonstrated that the probabilistic framework of
conditional random fields showed promise for multilabel classification, compared to discriminative
classifiers, as the number of labels within test documents increased. In follow-up work on these
models, \cite{Druck_Etc_2007} illustrated that this approach has the further benefit of being able
to naturally incorporate unlabeled data for semi-supervised learning.  A drawback of the CRF
approach is scalability, particularly when accounting for label dependencies. Exact inference  ``is
tractable only for about 3-12 [labels]'' \citep{GhamrawiMcCallum_2005}.  Alternatives to exact
inference considered in \citet{GhamrawiMcCallum_2005} include a ``supported  inference" method
which learns only to classify the label combinations that occur in the training set, and a
binary-pruning method that employs an intelligent pruning method which ignores dependencies between
all but the most commonly observed pairs of labels. Although this method may improve upon
approaches that ignore dependencies when restricted to datasets with few labels and many examples
(such as traditional benchmark datasets), it seems unlikely that any such methods will be able to
properly account for dependencies in datasets with power-law frequency statistics (since nearly
all dependencies in these datasets are between labels which have very sparse training data).

\citet{ZhangZhang10} present a hybrid generative-discriminative approach to multi-label classification.
 They first learn a Bayesian network structure that represents the independencies between labels.
 They then learn a discriminative classifier for each label in the order specified by the Bayesian
 network where the classifier for label $c$ takes as features not only the words in the document
 but also the output of the classifiers for each of the labels in the parent set of $c$ (i.e. the
 parent set specified by the Bayesian network). However, they apply their model to only small-scale
 datasets (the largest having $158$ labels).

In terms of discriminative approaches to multi-label classification, there is a large body of prior work, which has been well-summarized elsewhere in the literature
\citep[e.g., see][]{TsoumakasKatakis_2007,Tsoumakas_Etc_2009}.
Most discriminative approaches to multi-label classification have employed some variant of the ``binary problem-transformation" technique, in which the multi-label classification problem is transformed into a {\it set} of binary-classification problems, each of which can then be solved using a suitable binary classifier \citep{RifkinKlautau_2004,TsoumakasKatakis_2007,Tsoumakas_Etc_2009,Read_Etc_2009}.  The
most commonly employed method in the literature is the ``one-vs-all" transformation, in which $C$
independent binary classifiers are trained---one classifier for each label. These binary
classification tasks are then handled using discriminative classifiers, most notably SVMs, but also via other methods such as perceptrons, naive Bayes, and kNN classifiers.  As
our baseline discriminative method in this paper, we  use the ``one-vs-all" approach with SVMs as
the binary classifier, since this is the most commonly used discriminative approach in the current
multi-label classification literature, and has been defended in the literature in the face of an increasing number of proposed alternative methods \citep[e.g., see][]{RifkinKlautau_2004}.  We note also that there is a prior thread of work on discriminative approaches that can handle label-dependencies.   For example, another problem-transformation technique known as the ``Label Powerset''  method \citep{Tsoumakas_Etc_2009,Read_Etc_2009} builds a binary classifier for {\it each} distinct subset of label-combinations that exist in the training data---however, these approaches tend not to scale well with large label sets due to combinatorial effects \citep{Read_Etc_2009}.

\begin{figure}[t] % float placement: (h)ere, page (t)op, page (b)ottom, other (p)age
  \centering
 \includegraphics[width=1.0\textwidth, keepaspectratio]{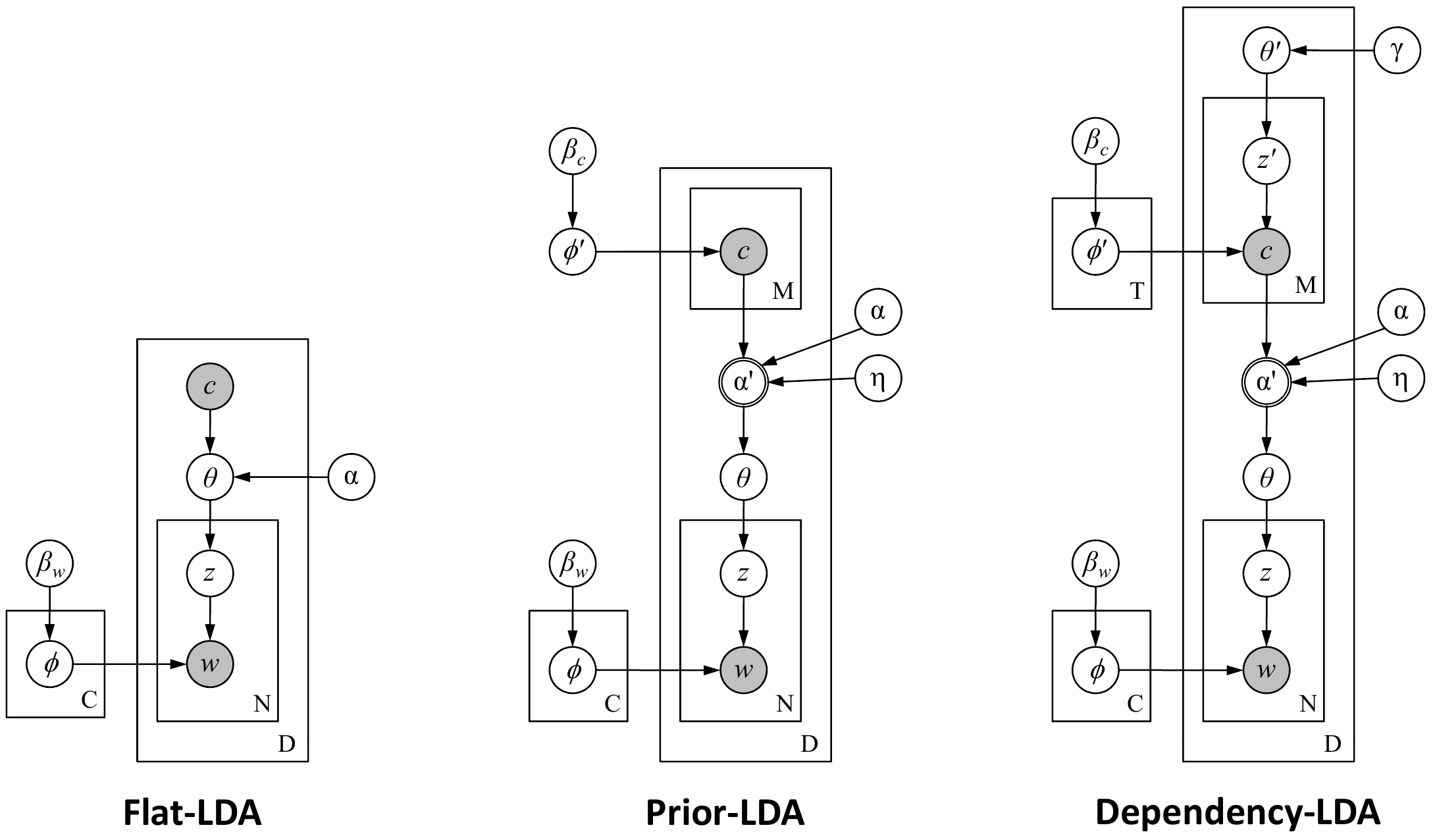}%{GraphicalModels_New}
  \caption{
  {\footnotesize  Graphical models for multi-label documents.
  The observed data for each document $d$ are a set of words $\bf{w}^{(d)}$  and labels $\bf{c}^{(d)}$.
   \textbf{Left}: In Flat-LDA, no generative assumptions are made regarding how labels are generated; labels for each document are assumed to be given.
   % ASSUMED TO BE OBSERVED
   \textbf{Center}: The Prior-LDA model assumes that the label-tokens $c^{(d)}$ for each document are generated by sampling from a corpus-wide multinomial distribution over label-types $\phi'$, which captures the relative frequencies of different label-types across the corpus.
   \textbf{Right}: The Dependency-LDA model assumes that the label-tokens for each document are sampled from a set of $T$ corpus-wide \textit{topics}---where each ``topic'' $t$ corresponds to a multinomial distribution over label-types $\phi'_t$---according to a document-specific distribution $\theta'_d$ over these topics.
  }
  }
  \label{fig:graphicalmodels}
\end{figure}

\section{Topic Models for Multilabel Documents}

In this section, we describe three models (depicted in Figure~\ref{fig:graphicalmodels}
using graphical model notation)  that extend the techniques of topic modeling to multi-label document classification.  Before providing the details for each model, we first briefly introduce the notation that will be used to describe these topic models within the multi-label inference setting, as well as provide a high-level description of the relationships between the three models.

The general setup of the inference task for the multi-label topic models we describe is as follows:  the observed data for each document $d \in \{1,\ldots,D\}$ are a set of words $\bf{w}^{(d)}$ and labels $\bf{c}^{(d)}$.  For all models, each label-type $c \in \{1,\ldots,C\}$ is modeled as a multinomial distribution $\phi_c$ over words.
Each document $d$ is modeled as a multinomial distribution $\theta_d$ over the document's observed label-types.  Words for document $d$ are generated by first sampling a label-type $z$ from $\theta_d$, and then sampling a word-token $w$ from $\phi_z$.  The three models that we present differ with respect to how they model the generative process for labels.

The first model we describe is a straightforward extension of LDA to labeled documents, which we will refer to as Flat-LDA,  where the labels are treated as given; this model makes no generative assumptions regarding how labels $\bf{c}^{(d)}$ are generated for a document.  
We then describe an extension to the Flat-LDA model---Prior-LDA---that incorporates a generative process for the labels themselves via a single corpus-wide multinomial distribution over all the label-types in the corpus.
This assumption of Prior-LDA is very useful for making predictions when the label-frequencies are highly non-uniform.
Lastly, we describe Dependency-LDA, which is a hierarchical extension to the previous two models that captures the dependencies between the labels by modeling the generative process for labels via a topic model; in Dependency-LDA, label-tokens for each document $d$ are sampled from a set of $T$ corpus-wide topics, according to a document-specific distribution ${\theta'}_d$ over the topics. We note that the Flat-LDA and Prior-LDA models can be viewed as special cases of the Dependency-LDA model.  In particular, the Prior-LDA model is equivalent Dependency-LDA if we set the number of topics $T=1$.

\subsection{Flat-LDA}

The latent Dirichlet allocation (LDA) model, also referred to as the topic model, is an unsupervised learning technique for extracting thematic information, called topics, from a corpus. LDA represents topics as multinomial distributions over the $W$ unique word-types in the corpus and represents documents as a mixture of topics. Flat-LDA is a straightforward extension of the LDA model to labeled documents. The set of LDA {\it topics} is substituted with the set of unique {\it labels} observed in the corpus. Additionally, each document's distribution over topics is restricted to the set of observed labels for that document.

More formally, let $C$ be the number of unique labels in the corpus. Each label $c$ is represented by a $W$-dimensional multinomial distribution $\phi_c$ over the vocabulary.  For document $d$, we observe both the words in the document $\bf{w}^{(d)}$ as well as the document labels $\bf{c}^{(d)}$. The generative process for Flat-LDA is shown below. Each document is associated with a multinomial distribution $\theta_d$ over its set of labels. The random vector $\theta_d$ is sampled from a symmetric Dirichlet distribution with hyper-parameter $\alpha$ and dimension equal to the number of labels $|\bf{c}^{(d)}|$. Given the distribution over topics $\theta_d$, generating the words in the document follows the same process as LDA:

{\small \begin{enumerate*}  %\addtolength{\itemsep}{-0.5\baselineskip}
    \item For each label $c \in \{1,\ldots,C\}$, sample a distribution over word-types $\phi_c \sim \mathrm{Dirichlet}(\cdot | \beta)$\vspace{2mm}
    \item For each document $d \in \{1,\ldots,D\}$

    \begin{enumerate*}  %\addtolength{\itemsep}{-0.5\baselineskip}
    \item Sample a distribution over its observed labels $\theta_d \sim \mathrm{Dirichlet}(\cdot | \alpha)$\vspace{2mm}
        \item For each word $i \in \{1, \ldots, N^W_d\}$
        \begin{enumerate*}
        \item Sample a label $z^{(d)}_i \sim \mathrm{Multinomial}\left(\theta_d\right)$
        \item Sample a word $w^{(d)}_i \sim \mathrm{Multinomial}\left(\phi_{c}\right)$ from the label $c=z^{(d)}_i$
    \end{enumerate*}
    \end{enumerate*}
\end{enumerate*}}

\noindent Note that this model assigns each word token within a document to just a single label---specifically to one of the labels that was assigned to the document.  The model is depicted using graphical model notation in the left panel of Figure~\ref{fig:graphicalmodels}.

Due to the similarity between the Flat-LDA model presented here, and both the Author-Model from \citet{Zvi_Etc_2004} and the L-LDA model from \citet{Ramage_Etc_2009}, it is important to note precisely the relationships between these models.
The Author-Model is conditioned on the set of {\it authors} in a document (and a ``topic'' is learned for each author in the corpus), whereas L-LDA and Flat-LDA are conditioned on the set of {\it labels} assigned to a document (and a ``topic'' is learned for each label in the corpus). L-LDA and Flat-LDA are {\it in practice} equivalent models, but employ different generative descriptions. Specifically, L-LDA models the generative process for each label in a document as a Bernoulli variable (where the parameter of the Bernoulli distribution is label-dependent).  However, during training, estimating the Bernoulli parameters is independent from learning the assignment of words to labels (i.e. the $z$ variables). Thus, during training both L-LDA and Flat-LDA reduce to standard LDA with an additional restriction that words can only be assigned to the observed labels in the document. Similarly, when performing inference for unlabeled documents (i.e. at test time), \citet{Ramage_Etc_2009} assume that L-LDA reduces to standard LDA.  In this way, both Flat-LDA and L-LDA are {\em in practice} equivalent despite L-LDA including a generative process for labels\footnote{Due to equivalence of Flat-LDA and L-LDA in practice, the experimental results we present for Flat-LDA are equivalent to what would be expected for L-LDA}. Due to the mismatch between the generative description of L-LDA and how it is employed in practice, we find it pedagogically useful to distinguish between the models presented here and L-LDA

\subsection{Prior-LDA}

An obvious issue with Flat-LDA is that it does not account for differences in
the relative frequencies of the labels within a corpus.  This is not a problem during training, because
all labels are observed for training documents. However, for the purpose of prediction (labeling new documents at test-time), accounting for the prior probabilities of each label becomes
important, particularly when there are dramatic differences in the frequencies of labels
in a corpus (as is the case with power-law datasets, as well as with
many traditional datasets, such as RCV1-V2).  In this section we present Prior-LDA,
which extends Flat-LDA by incorporating a generative
process for labels that accounts for differences in the observed frequencies of different label types. This is achieved using a two-stage generative process for each document,
in which we first sample a set of observed labels from a corpus-wide multinomial distribution,
and then given these labels, generate the words in the document.

Let $\phi'$ be a corpus-wide multinomial distribution over labels (reflecting, for example, a
power-law distribution of label frequencies). For document $d$, we draw $M_d$ samples from $\phi'$. Each sample can be thought of as a single vote for a particular label. We replace $\alpha^{(d)}$, the symmetric Dirichlet prior with hyperparameter $\alpha$, with a $C$-dimensional vector ${\alpha'}^{(d)}$ where the $i$th component is proportional to the total number of times label $i$ was sampled from $\phi'$. Formally, the vector ${\alpha'}^{(d)}$ is defined to be:
\begin{equation}
\label{eq:alphavec_priorlda}
{\alpha'}^{(d)} = \left[ \eta*\frac{N_{d,1}}{M_d}+{\alpha} \; , \; \eta*\frac{N_{d,2}}{M_d}+{\alpha}  \; , \;  \hdots  \; , \;  \eta*\frac{N_{d,C}}{M_d}+{\alpha} \; \right]
\end{equation}
where $N_{d,i}$ is the number of times label $i$ was sampled from $\phi'$. In other words, ${\alpha}'^{(d)}$ is a scaled, smoothed, normalized vector of label counts\footnote{In the training data, we set $M_d$ equal to the number of observed labels in document $d$ and $N_{d,i}$ equal to $0$ or $1$ depending upon whether the label is present in the document.}. The hyper-parameter ${\eta}$ specifies the total weight contributed by the observed labels $\bf{c}^{(d)}$ and the hyper-parameter ${\alpha}$ is an additional smoothing parameter that contributes a flat pseudocount to each label. We define the document's label set $\bf{c}^{(d)}$ to be the set of labels with a non-zero component in ${\alpha'}^{(d)}$. To make this model fully generative, we place a symmetric Dirichlet prior on $\phi'$.

Consider, for example, three labels $\{c_1, c_2, c_3 \}$ with frequencies $\phi' = \{ 0.5, 0.3, 0.2 \}$ in the corpus. For document $d$, we draw $M_d$ samples from $\phi'$. Assume $M_d = 5$ and the set $\{c_1, c_2, c_1, c_1, c_1\}$ was sampled. Then the hyper-parameter ${\alpha'}^{(d)}$ would be:
\begin{equation*}
{\alpha'}^{(d)} = \left[ \eta*\frac{4}{5}+{\alpha} \; , \; \eta*\frac{1}{5}+{\alpha}  \; , \;  \eta*\frac{0}{5}+{\alpha} \; \right]
\end{equation*}
\noindent If hyperparameter $\alpha = 0$, then ${\alpha'}^{(d)}$ has only two non-zero components (because the last component equals zero) and $\bf{c}^{(d)} = \{c_1, c_2\}$. In this case, the multinomial vector $\theta_d$ drawn from Dirichlet$\bigl({\alpha'}^{(d)}\bigr)$ will always have zero count for the third label (i.e. label $c_3$ will have probability zero in the document). If $\alpha > 0$, then $\bf{c}^{(d)} = \{c_1, c_2, c_3\}$ and label $c_3$ will have non-zero probability in the document. As $M_d$ goes to infinity, ${\alpha'}^{(d)}$ approaches the vector $\eta \; \phi' + \alpha$.

The multinomial distribution may seem like an unnatural choice
for a label-generating distribution since the observed labels
in a document are most naturally represented using binary variables
rather than counts.  We experimented with alternative parameterizations such as a multivariate Bernoulli distribution.
However, this introduced problems during both training and testing.
As noted by \citet{Schneider_2004_NBModels} in relation to modeling
document {\it words} (rather than labels), the multivariate Bernoulli
distribution tends to overweight negative evidence (i.e. the absence
of a word in a document) during training, due to the sparsity of the
word-document matrix. This problem is compounded when modeling
document {\it labels} because there are considerably fewer labels
in a document than words. Furthermore, at test time when the document labels are unobserved, a Bernoulli model will converge more slowly since the probability of turning on a label in a document is higher than the probability of turning off a label in a document (this is due to the fact that a label can only be turned off after all words assigned to that label have been assigned elsewhere)\footnote{A related issue was the reason given by \citet{Ramage_Etc_2009} for resorting in practice to a Flat-LDA scheme during inference.}.

The generative process for the Prior-LDA model is:

{\small \begin{enumerate*}  %\addtolength{\itemsep}{-0.5\baselineskip}
    \item Sample a multinomial distribution over labels ${\phi'} \sim \mathrm{Dirichlet}(\cdot | \beta_{\mathcal{C}})$\vspace{2mm}
    \item For each label $c \in \{1,\ldots,C\}$, sample a distribution over word-types $\phi_c \sim \mathrm{Dirichlet}(\cdot | \beta_{\mathcal{W}})$\vspace{2mm}
    \item For each document $d \in \{1,\ldots,D\}$:\vspace{1mm}
        \begin{enumerate*}
        \item Sample $M_d$ label tokens ${c^{(d)}_j} \sim \mathrm{Multinomial}\left({\phi'}\right)$,  $\; 1 \leq j \leq M_d$
        \item Compute the Dirichlet prior ${\alpha'}^{(d)}$ for document $d$ according to Equation~\ref{eq:alphavec_priorlda}\vspace{1.5mm}
        \item Sample a distribution over labels $\theta_d \sim \mathrm{Dirichlet}\left(\cdot | {\alpha'}^{(d)}\right)$\vspace{1mm}
        \item For each word $i \in \{1, \ldots, N^W_d\}$
            \begin{enumerate*}  %\addtolength{\itemsep}{-0.5\baselineskip}
                \item Sample a label $z^{(d)}_i \sim \mathrm{Multinomial}\left(\theta_d\right)$
                \item Sample a word ${w^{(d)}_i} \sim \mathrm{Multinomial}\left(\phi_{c}\right)$ from the label $c=z^{(d)}_i$
            \end{enumerate*}
        \end{enumerate*}
\end{enumerate*}}

\noindent This model is depicted using graphical model notation in the center  panel of
Figure~\ref{fig:graphicalmodels}.

\subsection{Dependency-LDA}

Prior-LDA accounts for the prior label frequencies observed in the training set,  but it does not
account for the dependencies between the labels, which is crucial when making predictions for
new documents.
In this section, we present Dependency-LDA, which extends Prior-LDA by incorporating another topic model to capture the dependencies between labels. The labels are generated via a topic model where each ``topic" is a distribution over labels.
Dependency-LDA is an extension of Prior-LDA in which there are $T$ corpus-wide
probability distributions over labels, which capture the dependencies between the labels, rather than a single corpus-wide distribution that merely reflects relative label frequencies.  We note that several models that represent or induce topic dependencies have been investigated in the past for unsupervised topic modeling (e.g., \citet{BleiLafferty_2005,Teh_etc_2006_hierarchical,Mimno_Etc_2007,Blei_etc_2010_Nested}).  Although these models are related to varying degrees to the Dependency-LDA model, as unsupervised models they are not directly applicable to document classification.

Formally, let $T$ be the total number of topics where each topic $t$ is a multinomial distribution over labels denoted $\phi'_t$. Generating a set of labels for a document is analogous to generating a set of words in LDA. We first sample a distribution over topics $\theta'_d$. To generate a single label we sample a topic $z'$ from $\theta'_d$ and then sample a label from the topic $\phi'_{z'}$. We repeat this process $M_d$ times. As in Prior-LDA, we compute the hyper-parameter vector ${\alpha'}^{(d)}$ according to Equation~\ref{eq:alphavec_priorlda} and define the label set $\bf{c}^{(d)}$ as the set of labels with a non-zero component. Given the set of labels $\bf{c}^{(d)}$, generating the words in the document follows the same process as Prior-LDA.

{\small
    \begin{enumerate*}
    \item For each topic $t \in \{1,\ldots,T\}$, sample a distribution over labels, ${\phi'}_t \sim \mathrm{Dirichlet}(\beta_{\mathcal{C}})$\vspace{1mm}
    \item For each label $c \in \{1,\ldots,C\}$, sample a distribution over words, ${\phi}_c \sim \mathrm{Dirichlet}( \beta_{\mathcal{W}})$\vspace{1mm}
    \item For each document $d \in \{1,\ldots,D\}$:\vspace{1mm}
    \begin{enumerate*}
           \item Sample a distribution over topics  ${\theta'}_d \sim \mathrm{Dirichlet}(\gamma)$\vspace{1mm}
           \item For each label $j \in \{1, \ldots, M_d\}$
            \begin{enumerate*}
                    \item Sample a topic ${z'}^{(d)}_j \sim \mathrm{Multinomial}\left({\theta'}_d\right)$
                    \item Sample a label ${ c^{(d)}_j} \sim \mathrm{Multinomial}\left(\phi'_{t}\right)$ from the topic $t={z'}^{(d)}_j$ %
            \end{enumerate*}
        \item Compute the Dirichlet prior ${\alpha'}^{(d)}$ for document $d$ according to Equation~\ref{eq:alphavec_priorlda}\vspace{1mm}
        \item Sample a distribution over labels $\theta_d \sim \mathrm{Dirichlet}\left(\cdot | {\alpha'}^{(d)}\right)$\vspace{1mm}
    \item For each word $i \in \{1, \ldots, N^W_d\}$
            \begin{enumerate*}
                \item Sample a label $z^{(d)}_i \sim \mathrm{Multinomial}\left(\theta_{d}\right)$
                \item Sample a word ${ w^{(d)}_i} \sim \mathrm{Multinomial}\left(\phi_{c}\right)$ from the label $c=z^{(d)}_i$
        \end{enumerate*}
    \end{enumerate*}
\end{enumerate*}}

\noindent The Dependency-LDA model is depicted using graphical model notation in the right panel of Figure~\ref{fig:graphicalmodels}.

\subsection{Topic Model Inference Methods --- Model Training}\label{sec:Inference_Train}
This section gives an overview of the inference methods used with the three LDA-based models (Flat-LDA, Prior-LDA, and Dependency-LDA).  We first describe how to perform inference and estimate the model parameters during training (i.e., when document labels are observed).  We then describe how to perform inference for test documents (i.e., when labels are unobserved).

Training all three LDA-based models requires estimating the $C$ multinomial distributions $\phi_c$ of labels over word-types.  Additionally, Prior-LDA and Dependency-LDA require estimation of the $T$ multinomial distributions $\phi'_t$ of topics over label types, where $T=1$ for Prior-LDA and $T>1$ for Dependency-LDA.  Additionally, training (and testing) for all models requires setting several hyperparameter values.

Note that we set the hyperparameter $\alpha=0$ in Prior-LDA and Dependency-LDA during training---but not during testing/prediction---which restricts the assignments of words to the set of observed labels for each document (see Equation~\ref{eq:alphavec_priorlda}).  This is consistent with the assumptions of these models, because in the training corpus all labels are observed, and the models assume that words are generated by one of the true labels.  This also greatly simplifies training, because it serves to decouple the upper and lower parts of the models (namely, with $\alpha=0$, the topic-label distributions ${\phi'}_t$ and the label-word distributions $\phi_c$ are conditionally independent from each other, given that we have observed all labels).

Furthermore, estimation of the $\phi_c$ distributions is in fact {\it equivalent} for all three models when $\alpha=0$ for Prior-LDA and Dependency-LDA (and, for consistency, we used the same set of parameter estimates for $\phi_c$ when evaluating all models).  A benefit---in terms of model evaluation---of using the same estimates for $\phi_c$ across all models is that it controls for one possible source of performance variability; i.e., it ensures that observed performance differences are due to factors other than estimation of $\phi_c$.  Specifically, differences in model performance can be directly attributed to qualitative differences between the models in terms of how they parameterize the Dirichlet prior ${\alpha'}^{(d)}$ for each test document.

In addition to the smoothing parameter $\alpha$, there are several other hyperparameters in the models that must be chosen by the experimenter.
For all experiments, hyperparameters were chosen heuristically, and were not optimized with respect to any of our evaluation metrics. Thus, we would expect that at least a modest improvement in performance over the results presented in this paper could be obtained via hyperparameter optimization.  For details regarding the hyperparameter values we used for all experiments in this paper, and a discussion regarding our choices for these values, see Appendix~\ref{sec:All_HyperParameter_Settings}.

\subsubsection{Learning the Label-Word Distributions:  ${\Phi}$}

To learn the $C$ multinomial distributions $\phi_c$ over words, we use a modified form of the collapsed Gibbs sampler described by~\citet{GriffithsSteyvers_2004} for unsupervised LDA.  In collapsed Gibbs sampling, we learn the distributions $\phi_c$ over words, and the $D$ distributions $\theta_d$ over labels,
by sequentially updating the latent indicator $z^{(d)}_i$ variables for all word tokens in the training corpus (where the $\phi_c$ and $\theta_d$  multinomial distributions are integrated--i.e., ``collapsed''--out of the update equations).

For Flat-LDA, the assignment of words in document $d$ is restricted to the set of observed labels $\bf{c}^{(d)}$.  For Prior-LDA and Dependency-LDA a word can be assigned to any label as long as the smoothing parameter $\alpha$ is non-zero.  The Gibbs sampling equation used to update the assignment of each word token $z^{(d)}_i$ to a label $c$ is:

{\footnotesize
\begin{equation}\label{z_wordtokens}
P(z^{(d)}_{i} = c \; | \; w^{(d)}_{i}=w,  \textbf{w}_{-i}, \ {\bf c}^{(d)} , {\boldsymbol\alpha'}^{(d)}, \textbf{z}_{-i} , \ \beta_{\mathcal{W}})
\propto
\frac{N^{WC}_{wc,-i} + \beta_{\mathcal{W}}}{\sum_{w'=1}^{W} \left( {N^{WC}_{w'c,-i}} + \beta_{\mathcal{W}}\right )}
*
\left({N^{CD}_{cd,-i}} + {\alpha'}^{(d)}_{c}\right)
\end{equation}
}

\noindent
where $N^{WC}_{wc}$ is the number of times the word $w$ has been assigned to the label $c$ (across the entire training set), and $N^{CD}_{cd}$ is the number of times the label $c$ has been assigned to a word in document $d$. We use a subscript $-i$ to denote that the current token, $z_i$, has been removed from these counts. The first term in Equation~\ref{z_wordtokens} is the probability of word $w$ in label $c$ computed by integrating over the $\phi_c$ distribution. The second term is proportional to the probability of label $c$ in document $d$, computed by integrating over the $\theta_d$ distribution.

For all results presented in this paper, during training we set $\alpha=0$ and $\eta$ equal to $50$. Early experimentation indicated that the exact value of $\eta$ was generally unimportant as long as $\eta \gg 1$. We ran multiple independent MCMC chains, and took a single sample at the end of each chain, where each sample consists of the current vector of $\bf{z}$ assignments (See Appendix~\ref{sec:All_HyperParameter_Settings} for additional details).  We use the $\bf{z}$ assignments to compute a point estimate of the distributions over words:

\begin{equation} \label{postpredphi}
\hat{{\phi}}_{w,c}=
\frac{N^{WC}_{wc} + \beta_{\mathcal{W}}}{\sum_{w'=1}^{W} \left( {N^{WC}_{w'c}} + \beta_{\mathcal{W}}\right )}
\end{equation}
\noindent where $\hat{{\phi}}_{w,c}$ is the estimated probability of word $w$ given label $c$. The parameter estimates $\hat{{\phi}}_{w,c}$ were then averaged over the samples from all chains. Several examples of label-word distributions, learned from a corpus of NYT documents, are presented in Table~\ref{table:labelworddistributions}.

Similarly, a point estimate of the posterior distribution over labels $\theta_d$ for each document is computed by:
\begin{equation}\label{postpredtheta_labels}
\hat{{\theta}}_{c,d}=
\frac{{N^{CD}_{cd}} + {\alpha'}^{(d)}_c }{\sum_{c'=1}^{C} \left( {N^{CD}_{c'd}} + {\alpha'}^{(d)}_{c'}\right )}
\end{equation}
\noindent where $\hat{\theta}_{c,d}$ is the estimated probability of label $c$ given document $d$.

\begin{table*}
\begin{centering}
{\scriptsize \begin{tabularx}{1\linewidth}{
>{\setlength\hsize{.33\hsize}\raggedright}X >{\setlength\hsize{.07\hsize}\centering}X ||
>{\setlength\hsize{.33\hsize}\raggedright}X >{\setlength\hsize{.07\hsize}\centering}X ||
>{\setlength\hsize{.33\hsize}\raggedright}X >{\setlength\hsize{.07\hsize}\centering}X ||
>{\setlength\hsize{.33\hsize}\raggedright}X >{\setlength\hsize{.07\hsize}\centering}X ||
>{\setlength\hsize{.33\hsize}\raggedright}X >{\setlength\hsize{.07\hsize}\centering}X }
%\textsc{united states armament and defense} & 285 &  \textsc{arms sales abroad} & 176 &  \textsc{abortion} & 24 &  \textsc{acid rain} & 11 &  \textsc{agni missile} & 1 \tabularnewline
\textsc{Politics And Government} & 285 &  \textsc{Arms Sales Abroad} & 176 &  \textsc{Abortion} & 24 &  \textsc{Acid Rain} & 11 &  \textsc{Agni Missile} & 1 \tabularnewline
& & & & & & & & & \\ [-5pt]
\hline
& & & & & & & & & \\ [-4pt]
party   &   .014    &   iran    &   .021    &   abortion    &   .098    &   acid    &   .070    &   missile &   .032    \tabularnewline
government  &   .014    &   arms    &   .019    &   court   &   .033    &   rain    &   .067    &   india    &  .031    \tabularnewline
political   &   .011    &   reagan  &   .014    &   abortions   &   .028    &   lakes   &   .028    &   technology  &   .016    \tabularnewline
leader  &   .006    &   house   &   .014    &   women   &   .017    &   environmental   &   .026    &   missiles    &   .016    \tabularnewline
president   &   .005    &   president   &   .014    &   decision    &   .016    &   sulfur  &   .024    &   western &   .015    \tabularnewline
officials   &   .005    &   north   &   .012    &   supreme &   .016    &   study   &   .023    &   miles   &   .014    \tabularnewline
power   &   .005    &   report  &   .011    &   rights  &   .015    &   emissions   &   .021    &   nuclear &   .013    \tabularnewline
leaders &   .005    &   white   &   .011    &   judge   &   .015    &   plants  &   .021    &   indian  &   .013    \tabularnewline
%new &   .005    &   contras &   .010    &   woman   &   .015    &   pollution   &   .017    &   program &   .013    \tabularnewline
%country &   .004    &   intelligence    &   .010    &   law &   .013    &   reilly  &   .015    &   range   &   .010    \tabularnewline
%general &   .004    &   colonel &   .010    &   issue   &   .011    &   coal    &   .013    &   military    &   .010    \tabularnewline
%members &   .004    &   affair  &   .010    &   legal   &   .011    &   power   &   .012    &   ballistic   &   .010    \tabularnewline
\end{tabularx}}
\caption{The eight most likely words for five labels in the NYT Dataset, along with
the word probabilities. The number to the right of the labels
indicates the number of training documents assigned the label. }
\label{table:labelworddistributions}
\end{centering}
\end{table*}

\subsubsection{Learning the Topic-Label Distributions: ${\Phi'}$}

Note that this section only applies to the Prior-LDA and Dependency-LDA models since the Flat-LDA model does not employ a generative process for labels \footnote{Additionally, since there is only one ``topic'' to learn for the Prior-LDA model, the estimation problem for this model simplifies to computing a single maximum-a-posteriori estimate of the dirichlet-multinomial distribution ${\phi'}$}. Learning the $T$ multinomial distributions $\phi'_t$ over labels is equivalent to applying a standard LDA model to the label tokens. In our experiments, we employed a collapsed Gibbs sampler \citep{GriffithsSteyvers_2004} for unsupervised LDA, where the update equation for the latent topic indicators ${z_{i}'^{(d)}}$ is given by:

{\footnotesize
\begin{equation}\label{z_labeltokens}
P({z'}^{(d)}_{i} = t \; | \  c^{(d)}_{i}=c, \ \textbf{c}_{-i},  \textbf{z'}_{-i} , \ {\gamma}, \ \beta_{\mathcal{C}})
\propto
\frac{{N^{CT}_{ct,-i}} + \beta_{\mathcal{C}}}{\sum_{c'=1}^{C}  \left( {N^{CT}_{c't,-i}} + \beta_{\mathcal{C}}\right )}
*
\left({N^{DT}_{dt,-i}} + {\gamma} \right)
\end{equation}
}

\noindent
where ${N^{CT}_{ct}}$ is the number of times label $c$ has been assigned to topic $t$ (across the entire training set), and ${N^{DT}_{dt}}$ is the number of times topic $t$ has been assigned to a label in document $d$. The subscript $-i$ denotes that the current label-token $z'_i$ has been removed from these counts.
The first term in Equation~\ref{z_labeltokens} is the probability of label $c$ in topic $t$ computed by integrating over the $\phi'_t$ distribution. The second term is proportional to the probability of topic $t$ in document $d$, computed by integrating over the $\theta'_d$ distribution.

For training, we experimented with different values of $T \leq C$ (for Dependency-LDA).  We set $\gamma \ll 1$, and adjusted $\beta_{\mathcal{C}}$ in proportion to the ratio of the number of topics $T$ to the total number of observed labels in each training corpus (See Appendix~\ref{sec:All_HyperParameter_Settings} for additional details).

For each MCMC chain, we ran the Gibbs sampler for a burn-in of 500 iterations, and then took a single sample of the vector of $\bf{z}'$ assignments. Given this vector, we compute a posterior estimate for the $\phi'_t$ distributions:

\begin{equation}\label{postpredtopics}
\hat{{\phi}'}_{c,t} = \frac{{N^{CT}_{ct}} + \beta_{\mathcal{C}}}{\sum_{c'=1}^{C} \left( {N^{CT}_{c't}} + \beta_{\mathcal{C}} \right) }
\end{equation}

\noindent where $\hat{{\phi'}}_{c,t}$ is the estimated probability of label $c$ given topic $t$.  For each training corpus, we ran ten MCMC chains (giving us ten distinct sets of topics)\footnote{We can not average our estimates of ${\phi'}_t$ over multiple chains as we did when estimating $\phi_c$.  This because the topics are being learned in an unsupervised manner, and do not have a fixed meaning between chains.  Thus, each chain provides a distinct estimate of the set of $T$ \ ${\phi'}_t$ distributions.  For test documents, we average our predictions over the set of $10$ chains.  See Appendix~\ref{sec:All_HyperParameter_Settings} for additional details.}.  Several examples of topics, learned from a corpus of NYT documents, are presented in Table~\ref{table:topicsofconcepts}.

\begin{table*}[t]
\begin{centering}
%{\scriptsize \begin{tabular*}{1\linewidth}{@{\extracolsep{\fill}}  l c || l c || l c }
{\tiny \begin{tabular*}{1\textwidth}{@{\extracolsep{\fill}}  l c || l c || l c }
\textbf{``Consumer Safety''} & \textbf{.017} & \textbf{``Warfare And Disputes''}&\textbf{.024}& \textbf{``Cheating and Athletics''} &\textbf{.016}\tabularnewline
                                        &      &                                                &      &                                                &      \\ [-4pt]
\hline
                                        &      &                                                &      &                                                &      \\ [-4pt]
\textsc{cancer}                         & .078 & \textsc{armament, defense and military...}     & .162 & \textsc{olympic games (1988) }                 & .052 \tabularnewline
\textsc{hazardous and toxic substances} & .039 & \textsc{international relations}               & .133 & \textsc{suspensions, dismissals and resig...}  & .038 \tabularnewline
\textsc{pesticides and pests}           & .021 & \textsc{united states international rela...}   & .132 & \textsc{baseball}                              & .033 \tabularnewline
\textsc{research}                       & .021 & \textsc{civil war and guerrilla warfare}       & .098 & \textsc{summer games (olympics) }              & .031 \tabularnewline
\textsc{surgery and surgeons}           & .021 & \textsc{military action}                       & .053 & \textsc{football}                              & .029 \tabularnewline
\textsc{tests and testing}              & .021 & \textsc{chemical warfare}                      & .029 & \textsc{athletics and sports}                  & .026 \tabularnewline
\textsc{food}                           & .018 & \textsc{refugees and expatriates}              & .019 & \textsc{college athletics}                     & .019 \tabularnewline
\textsc{recalls and bans of products}   & .018 & \textsc{independence movements}                & .013 & \textsc{steroids}                              & .019 \tabularnewline
\textsc{consumer protection}            & .016 & \textsc{boundaries and territorial issues}     & .011 & \textsc{gambling}                              & .017 \tabularnewline
\textsc{health, personal}               & .016 & \textsc{kurds}                                 & .010 & \textsc{winter games (olympics)}               & .017\tabularnewline
\end{tabular*}}
\caption{ The ten most likely labels within three of the topics learned by the Dependency LDA model on the NYT dataset. Topic labels (in quotes) are subjective interpretations provided by the authors.}
\label{table:topicsofconcepts}
\end{centering}
\end{table*}

%%%%%%%%%%%%
%EQ(7)
Similarly, a point estimate of the posterior distribution over topics ${\theta'}_d$ for each document is computed by:
\begin{equation}\label{postpredtheta_topics}
\hat{\theta'}_{d,t} = \frac{ {N^{DT}_{dt}} + {\gamma} }{\sum_{t'=1}^{T} \left( {N^{DT}_{d't}} + {\gamma} \right )}
\end{equation}

\noindent where $\hat{\theta'}_{d,t}$ is the estimated probability of topic $t$ given document $d$.

\subsection{Topic Model Inference Methods --- Test Documents} \label{sec:Inference_Test}

In this section, we first describe a proper inference method for sampling the three LDA-based models during test time, when the document labels are unobserved.  In the following section, we describe an approximation to the proper inference method which is computationally much faster, and achieved performance that was as accurate as the true sampling methods.  We note again that the hyperparameter settings used for all experiments are provided in Appendix~\ref{sec:All_HyperParameter_Settings}.

At test time, we fix the label-word distributions $\hat{\phi}_c$, and topic-label distributions $\hat{\phi'}_t$, that were estimated during training. Inference for a test document $d$ involves estimating its distribution over label types $\theta_d$ and a set of label-tokens $\bf{c}^{(d)}$, given the observed word tokens ${\bf w^{(d)}}$.  Additionally, inference for Dependency-LDA involves estimating a document's distribution over topics, ${\theta'}_d$.  We first describe inference at the word-label level (which is equivalent for all three LDA models given the Dirichlet prior ${\alpha}'^{(d)}$), and then describe the additional inference steps involved in Dependency-LDA.  Note that for all models, inference for each test document is independent.

The $\theta_d$ parameter is estimated by sequentially updating the $z^{(d)}_i$ assignments of word tokens to label types.  The Gibbs update equation is modified from Equation~\eqref{z_wordtokens} to account for the fact that we are now using fixed values for the $\phi_c$ distributions, which were learned during training, rather than an estimate computed from the current values of $z$ assignments via ${N^{WC}_{wc}}$:

{\small
\begin{equation}\label{z_testwordtokens}
P\left(z^{(d)}_{i} = c \; \ | \; w^{(d)}_{i}=w, \ \textbf{w}^{(d)}_{-i}, \ {\alpha'}^{(d)}, \ z^{(d)}_{-i} , \ \hat{\phi}_{w,c}\right)
\; \propto \;
{\hat{\phi}_{w,c}}
*
\left({N^{CD}_{cd,-i}} + {\alpha'}^{(d)}_{c}\right)
%\cdot
\end{equation}
}

\noindent where $\hat{\phi}_{w,c}$ was estimated during training using Equation~\eqref{postpredphi},
$N^{CD}_{cd}$ is the number of times the label $c$ has been assigned to a word in document $d$,
and where ${\alpha'}^{(d)}_{c}$ is the value of the document-specific Dirichlet prior
on label-type $c$ for document $d$, as defined in Equation~\eqref{eq:alphavec_priorlda}.

The only difference that arises between the three LDA models when sampling the $\bf{z}$ variables is in the document-specific prior ${\alpha'}^{(d)}$.
To simplify the following discussion, we describe inference in terms of Dependency-LDA.  We note again that Prior-LDA is a special case of Dependency-LDA in which $T=1$, and therefore the descriptions of inference for Dependency-LDA are fully applicable to Prior-LDA.\footnote{In Flat-LDA, there is no document-specific Dirichlet prior.  Instead, the prior for each document is simply a symmetric Dirichlet with hyperparameter $\alpha$, i.e.  \ ${\alpha'}^{(d)}_{c}=\alpha, c\in{1 \ldots C}$.  Since this does not depend on any additional parameters, the remaining steps provided in this section are irrelevant to Flat-LDA.}

Since the label tokens are unobserved for test documents, exact inference requires that we sample the label tokens $\bf{c}^{(d)}$ for the document.  The label tokens $\bf{c}^{(d)}$ are dependent on the assignment $\bf{z}'$ of label-tokens to topics in addition to the vector of word-assignments $\bf{z}$.  We therefore must also sample the variables $z'^{(d)}$. The Gibbs sampling equation for $c^{(d)}_i$, given the trained model, and a document's vector of $z$ and $z'$ assignments, is:

{\small
\begin{equation}\label{c_testlabelc}
p \left( c^{(d)}_i = c \; | \; {z'}^{(d)}_{i}=t , \; z'^{(d)}_{-i}, \; c^{(d)}_{-i}, \;  z^{(d)}, \ \hat{\phi'}_{t,c} \; \right)
\propto
\frac{ \prod_{c'=1}^C\Gamma\left(\alpha'^{(d)}_{c'}+ N^{CD}_{c',d}\right)} {\prod_{c=1}^C\Gamma\left( \alpha'^{(d)}_{c'}\right)}
\; \cdot \;
\hat{\phi'}_{t,c}
\end{equation}
}

\noindent where the first term on the right-hand side of the equation is the likelihood of the current vector of word assignments to labels $\bf{z^{(d)}}$ given the proposed set of label-tokens $\bf{c}^{(d)}$ (i.e., updated with value $c^{(d)}_i = c$), and ${N^{CD}_{cd}}$ is the total number of words in document $d$ that have been assigned to label $c$.  The second term $\hat{\phi'}_{c,t}$ was estimated during training using Equation~\eqref{postpredtopics}. Since the update equation for $c^{(d)}_i$ is not transparent from the model itself, and has not been presented elsewhere in the literature, we provide a derivation of Equation~\eqref{c_testlabelc} in Appendix~\ref{sec:Derivation_CSampler}.

Given the current values of the label tokens $\bf{c}^{(d)}$, the topic assignment variables ${z'}^{(d)}$ are conditionally independent of the label assignment variables $z^{(d)}$.  The update equations for the ${z'}^{(d)}$ variables are therefore equivalent to Equation~\eqref{z_testwordtokens}, except that we are now updating the assignment of labels to topics rather than words to labels:

{\small
\begin{equation}\label{z_testlabeltokens}
P\left({z'}^{(d)}_{i} = t \; \ | \; \ c^{(d)}_{i}=c, \ \gamma, \ {\bf z'}^{(d)}_{-i} , \ \hat{\phi'}_{t,c}\right)
\; \propto \;
\hat{\phi'}_{c,t}
*
\left({N^{DT}_{dt,-i}} + {\gamma}\right)
%\cdot
\end{equation}
}

\noindent where ${N^{DT}_{dt,-i}}$ is the number of times topic $t$ has been assigned to a label in document $d$, and the document-specific distribution over topics ${\theta'}_d$ has been integrated out.

For each test document $d$, we sequentially update each of the values in the vectors $\bf{z}^{(d)}$, $\bf{c}^{(d)}$, and $\bf{z'}^{(d)}$.  Since the $\bf{z}^{(d)}$ variables are conditionally independent of the $\bf{z'}^{(d)}$ variables given the $\bf{c}^{(d)}$ variables, the $\bf{c}^{(d)}$ variables are the means by which the word-level information contained in $\bf{z}^{(d)}$ and the topic-level information contained in $\bf{z'}^{(d)}$ can propagate back and forth. Thus, a reasonable update order is as follows:

{\small
\begin{enumerate*}
    \item Update the assignment of the {\it observed} word tokens {\bf $w^{(d)}$} to the labels: $z^{(d)}$ \ \ \ (Eq.~\ref{z_testwordtokens})
    \item Sample a new set of label-tokens: $c^{(d)}$ \ \ \ (Eq.~\ref{c_testlabelc})
    \item Update the assignment of the {\it sampled} label-tokens to one of $T$ topics: ${z'}^{(d)}$ \ \ \ (Eq.~\ref{z_testlabeltokens})
    \item Sample a new set of label-tokens: $c^{(d)}$ \ \ \  (Eq.~\ref{c_testlabelc})
\end{enumerate*}
}

\noindent Each full cycle of these updates provides a single `pass' of information from the words up to the topics and back down again.  Once the sampler has been sufficiently burned in, we can then use the vectors $\bf{z}^{(d)}$, $\bf{c}^{(d)}$ and $\bf{z'}^{(d)}$ to compute a point estimate of a test document's distribution $\hat{{\theta}}_d$ over the label types using Equation~\ref{postpredtheta_labels} (and the prior as defined in Equation~\ref{eq:alphavec_priorlda}).

Unfortunately, the proper Gibbs sampler runs into problems with computational efficiency.  Intuitively, the source of these problems is that the $c$ variables act as a bottleneck during inference since they are the only means by which information is propagated between the $z$ and $z'$ variables.  To limit the extent of this bottleneck, we can increase the number of label tokens $M_d$ that we sample.  However, this is computationally expensive because sampling each $c$ value requires substantially more computation than sampling the $z$ and $z'$ assignments,  since computing each proposal value requires taking a product of $C$ gamma values.\footnote{There are  methods to optimize the sampler for $c^{(d)}$, which reduces the amount of computation required by several orders of magnitude (using simplification of the expression in Eq.~\ref{c_testlabelc} and careful storage and updating of the vector of gamma values).  However, this method was still slower by an order of magnitude per iteration than the `fast inference' method presented in the following section, and required a much longer burn-in (while giving similar, or worse, prediction performance).}

\subsubsection{Fast Inference for Dependency-LDA}

\begin{table}[t]
\begin{centering}
\begin{tabular}{rrll||rrll}
\multicolumn{ 3}{c}{{\bf Training}} &     {\bf } &            &   \multicolumn{ 3}{c}{{\bf Testing}} \\[.5pt]
\hline
           &            &            &            &            &     &            &  \\[-6.5pt]
Training $\Phi$  &            & $O(N_W(N_C/D))$  &            &            &   Flat-LDA &       & $O(N_W C)$ \\
Training $\Phi'$ &            & $O(N_C T)$  &            &            &  Prior-LDA &            & $O(N_W C)$ \\
           &            &            &            &            &    Dep-LDA &            & $O(N_W(C+T))$  \\[1.5pt]
\hline
\end{tabular}
\vspace{-4pt}
\caption{Computational Complexity (per iteration) for the three LDA-based methods.  $N_W$: Number of word-tokens in the dataset; $N_C$: Number of observed label tokens in the (training) set; $D$: Number of documents in the training set; $C$: Number of unique label-types; $T$: Number of topics.}
\vspace{-6pt}
\label{table:ModelComplexity}
\end{centering}
\end{table}

We now describe an efficient alternative to the sampling method described above.  Experimentation with this alternative inference method suggests that, in addition to requiring substantially less time, it in fact achieves similar or better prediction performance compared to proper inference.

The idea behind the fast-inference method is that, rather than explicitly sampling the values of $c$, we directly pass information between the label-level and topic-level parameters (thus avoiding the information bottleneck created by the $c$ tokens, and also avoiding this costly inference step).  This can be achieved by directly passing the $z$ values up to the topic-level, and treating each $z$ value as if it was an observed label token $c$.  In other words, we substitute the vector of sampled label tokens $c^{(d)}$ with the vector of label assignments $z^{(d)}$ for each document; since both $z^{(d)}_i$ and $c^{(d)}_i$ can take on the same set of values (between 1 and $C$), these vectors can be treated equivalently when sampling the topic-assignments ${z'}^{(d)}_{i}$ for them.  Then, after updating the $z'$ values, we can directly compute the posterior predicted distribution over label types, $p(c|d)$, by conditioning on the current $z'$ assignments, and use this to compute $\alpha'^{(d)}$.

To motivate this approach, let $\Phi'$ be the $T$-by-$C$ matrix where row $t$ contains ${\phi'}_t$. Let $\theta'_d$ be the $T$-dimensional multinomial distribution over topics. We can directly compute the posterior predictive distribution over labels given $\Phi'$ and $\theta'_d$, as follows:
\vspace{-3mm}
{\small
\begin{equation}
\begin{split}
p(c^{(d)}_i=c \ | \ {\theta'}_d, {\Phi'}) &\propto \sum_{t=1}^T p(c^{(d)}_i = c \ | \ {z'}^{(d)}_i = t) \cdot p({z'}^{(d)}_i = t  \ | \ d) \\
						       &= \sum_{t=1}^{T} {\Phi'_{t,c}} \cdot {\theta'}_{d,t}
\end{split}
\end{equation}}
\noindent Thus, given the matrix $\Phi'$ (learned during training) and an estimate of the $T$-dimensional vector $\theta'_d$, which we can compute using Equation~\eqref{postpredtheta_topics}, the hyper-parameter vector $\alpha'^{(d)}$ can be directly computed using:
\begin{equation}\label{alphavec_deplda}
\begin{split}
{\alpha'}^{(d)}
\; \; & = \; \;
\eta \ ( \hat{{\theta'}_d} \cdot \Phi' ) + \alpha \\
\end{split}
\end{equation}
\noindent Once we have updated the $z'$ variables, Equation~\eqref{alphavec_deplda} allows us to compute $\alpha'^{(d)}$ directly without explicitly sampling the $c$ variables\footnote{This is in fact the correct posterior-predicted value of ${\alpha'}^{(d)}$ in the generative model, given the variables $\Phi'$ and ${\theta'}_d$.  However, technically this is not correct during inference, because it ignores the values of the $\bf{z}^{(d)}$ variables, which are accounted for in the first term in Equation~\ref{c_testlabelc}.}. An alternative defense of this approach is that as $M_d$ goes to infinity in the generative model for Dependency-LDA, the vector $\alpha'^{(d)}$ approaches the expression given in Equation~\ref{alphavec_deplda}.

The sequence of update steps we use for this approximate inference method is:

{\small
\begin{enumerate*}
    \item Update the assignment of the {\it observed} word tokens {\bf $w^{(d)}$} to one of the $C$ label types: $z^{(d)}$ \ \ (Eq.~\ref{z_testwordtokens})
    \item Set the label-tokens ($c^{(d)}$) equal to the label assignments: $c^{(d)}_i = z^{(d)}_i$
    \item Update the assignment of the label tokens to one of $T$ topics: ${z'}^{(d)}$ \ \ (Eq.~\ref{z_testlabeltokens})
    \item Compute the hyperparameter vector:  ${\alpha'}^{(d)}$   \ \  (Eq.~\ref{alphavec_deplda})
\end{enumerate*}
}

\noindent As before, each full cycle of these updates provides a single `pass' of information from the words up to the topics and back down again.  But rather than sampling the $c^{(d)}$ label-tokens, we directly pass the $z^{(d)}$ variables up to the topic-level sampler, and use these as an approximation of the vector $c^{(d)}$.  Then, given the current estimate of ${\theta'}^{(d)}$ (shown in Equation ~\ref{postpredtheta_topics}), we compute the $\alpha'^{(d)}$ prior directly using Equation~\ref{alphavec_deplda}.\footnote{ Note that the computational steps involved in this method are in fact very close to the proper inference methods.  The first and third steps (updating $z$ and $z'$) are equivalent to the true sampling updates.
The second step actually closely replicates what we would expect if we set $M_d=N^W_d$ and then sampled each $c^{(d)}_i$ explicitly, except that we are now ignoring the topic-level information when we actually construct the vector $c^{(d)}$ (although this information has a strong influence on the $z$ assignments, so it is not unaccounted for in the $c^{(d)}$ vector).}

Once the sampler has been sufficiently burned in, we can then use the assignments $\bf{z}^{(d)}$, and $\bf{z'}^{(d)}$ to compute a point estimate of a test document's distribution $\hat{{\theta}}_d$ over the label types using Equation~\ref{postpredtheta_labels} (and the prior as defined in Equation~\ref{alphavec_deplda}).

We compared performance
between this method and the proper inference method (with $M_d=1000$) on a single split of the EURLex corpus.  In addition to providing significantly better predictions on the test dataset, the fast inference method was more efficient.  Even after optimizing the $c^{(d)}_i$ sampling, the fast inference method was well over an order of magnitude faster (per iteration) than proper inference, and also converged in fewer iterations.  Due to its computational benefits, we employed the fast inference method for all experimental results presented in this paper.

The computational complexity for training and testing the three LDA-based algorithms is presented in Table~\ref{table:ModelComplexity}.\footnote{Complexity for Dependency-LDA during testing is given for the fast-inference method.}  Note that the complexity of Dependency-LDA does not involve a term corresponding to the square of the number of unique labels ($C$), which is often the case for algorithms that  incorporate label dependencies \citep[a discussion of this issue can be found in, e.g.,][]{Read_Etc_2009}.

\subsection{Illustrative Comparison of Predictions across Different Models}

\begin{figure*}[t]
  \centering
  \includegraphics[width=1\linewidth]{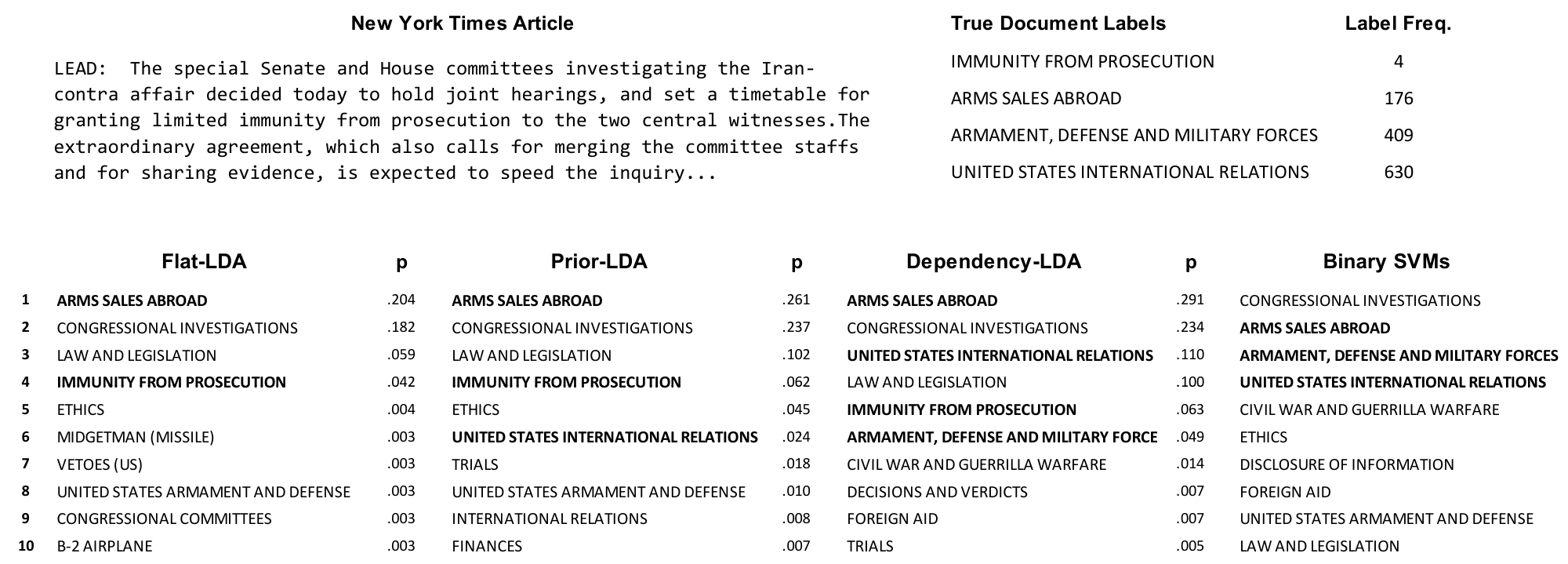}
   \caption{Illustrative comparison of a set of prediction results for a single NYT test document.}
  \label{fig:DocPredictionExample_NYT}
\end{figure*}

To illustrate the differences between the three models, consider a word $w$ that  has equal
probability under two labels $c_1$ and $c_2$ (i.e., $\phi_{1,w} = \phi_{2,w}$). In Flat-LDA, the
Dirichlet prior on $\theta_d$ is uninformative, so the only difference between the probabilities
that $z$ will take on value $c_1$ versus $c_2$ are due to the differences in the number of
current assignments ($N^{CD}$ for $c_1$ and $c_2$) of word tokens in document $d$. In Prior-LDA,
the Dirichlet prior reflects the relative {\it a-priori} label-probabilities (from the single
corpus-wide topic), and therefore the $z$ assignment probabilities will reflect the baseline
frequencies of the two labels in addition to the current $z$ counts for this document. In
Dependency-LDA, the Dirichlet prior reflects a prior distribution over labels given an (inferred)
document-specific mixture of the $T$ topics, and therefore the assignment probabilities reflect the relationships between the (inferred) document's labels and all other labels, in addition to the
current counts of $z$.

Figure~\ref{fig:DocPredictionExample_NYT} shows an illustrative example of the predictions different models made for a single document in the NYT collection. An excerpt from this document is shown alongside the four true labels that were manually assigned by the NYT editors. The top ten label predictions (with the true labels in bold) illustrate how Dependency-LDA leverages both baseline frequencies and correlations to improve predictions over the simpler Prior-LDA and Flat-LDA models.  Additionally, this illustration indicates how Dependency-LDA can achieve better performance than SVMs by improving performance on rare labels.

Given the set of label-word distributions learned during training, Flat-LDA predicts the labels which most directly correspond to the words in the document
(i.e., the labels that are assigned the most words when we do not account for any information beyond the label-word distributions, due to the words having high probabilities $\phi_{c,w}$ under the models for these labels).
As shown in Figure~\ref{fig:DocPredictionExample_NYT}, this Flat-LDA approach ranks two out of four of the true labels among its top ten predictions, including the rare label {\sc Immunity from Prosecution}.    Prior-LDA improves performance over Flat-LDA by excluding infrequent labels, except when the evidence for them overwhelms the small prior. For example,  the rare label {\sc Midgetman (Missile)} which is ranked sixth for Flat-LDA---but has a relatively small probability under the model---is not ranked in the top ten for Prior-LDA, whereas {\sc Immunity from Prosecution}, which is also a rare label but has a much higher probability under the model, stays in the same ranking position under Prior-LDA.  Also, the label {\sc United States International Relations}, which isn't ranked in the top ten under Flat-LDA, is ranked sixth under Prior-LDA due in part to its high prior probability (i.e. its high baseline frequency in the training set).

The Dependency-LDA model improves upon Prior-LDA by additionally including {\sc Armament, Defense and Military Forces} high in its rankings. This improvement is attributed to the semantic relationship between this label and the
labels {\sc Arms Sales Abroad} and {\sc United States International Relations} (e.g., note that the labels {\sc Armament, Defense and Military Forces} and {\sc United States International Relations} are, respectively, the first and third most likely labels under the middle topic shown in Table~\ref{table:topicsofconcepts}). Lastly, note that binary SVMs\footnote{These predictions were generated by the ``Tuned SVM'' implementation, the details of which are provided in Section~\ref{sec:SVM_Implementation}} performed well on the three frequent labels, but missed the rare label {\sc Immunity From Prosecution}. This is because the binary SVMs learned a poor model for the label due to the infrequency of training examples, which---as discussed in the introduction---is one of the key problems with the binary SVM methods.

\section{Experimental datasets} \label{sec:datasets}

The emphasis of the experimental work in this paper is on two multi-label datasets each
containing many labels and skewed label-frequency distributions: the NYT annotated
corpus~\citep{NYTCorpus} and the EUR-Lex text dataset \citep{MenciaFurnkranz_2008_Efficient}. We use a
subset of 30,658 articles from the full NYT annotated corpus of 1.5 million documents, with over
4000 unique labels that were assigned manually by The New York Times Indexing Service. The EUR-Lex
dataset contains 19,800 legal documents with 3,993 unique labels. In addition, for comparison, we
present results from three more commonly used benchmark multi-label datasets: the
RCV1-v2 dataset of \cite{Lewis_Etc_2004} and the {\it Arts} and {\it Health} subdirectories from
the Yahoo! dataset
\citep[][]{UedaSaito_2002, JiTangYuYe_2008}, all of which have significantly fewer labels, and more
examples per label, than the NYT and EUR-Lex datasets.  Complete details on all of the datasets are provided in Appendix~\ref{sec:Appendix_DatasetDetails}.

Aspects of document classification relating to feature-selection and document-representation are active areas of research \citep[e.g.,
see][]{Forman_2003_FeatureSelection,ZhangEtc_2009_FeatureSelection}.  In order to avoid confounding
the influence of feature selection and document representation methods with performance differences between
the models, we employed straightforward methods for both.  Feature selection for all datasets was
carried out by (1) removing stop words and (2) removing highly-infrequent words.  For LDA-based
models, each document was represented using a {\it bag-of-words} representation (i.e, a vector of
word counts).  For the binary SVM classifiers, we normalized the word counts for each document such that each document feature-vector summed to one (i.e., a vector of reals).

\begin{table}[t]
{\small \begin{tabular*}{1\linewidth}{@{\extracolsep{\fill}}  r|rrrrrrrrrr}
\begin{sideways}Dataset\end{sideways} & \begin{sideways}Labels $(C)$\end{sideways}   & \begin{sideways}Documents. $(D)$\end{sideways}  & \begin{sideways}Cardinality\end{sideways}   & \begin{sideways}Density\end{sideways}  & \begin{sideways}Mean Label Freq.\end{sideways} & \begin{sideways}Median Label Freq. \end{sideways}& \begin{sideways}Mode Label Freq. \end{sideways}& \begin{sideways}Distinct Labelsets\end{sideways} & \begin{sideways}Labelset Freq.\end{sideways}  & \begin{sideways}Unique Labelset Prop.\end{sideways}  \\ [.5 ex]
\hline
                &               &            &             &            &            &            &         &            &                         \\ [-1 ex]
Y!  {\it Arts}  &         19    &       7,441 &      1.6    &      .0855 &        636 &        530 &     -- &        527 &       14.1 &      .0406 \\
Y! {\it Health} &         14    &       9,109 &      1.6    &      .1149 &      1,047 &        500 &     -- &        241 &       37.8 &      .0113 \\
RCV1-V2         &        103    &     804,414 &      3.2    &      .0315 &     25,310 &      7,410 &     -- &     13,922 &       57.8 &      .0093 \\ [1.2 ex]
%               &               &             &             &            &            &            &        &            &            %             \\ [-1 ex]
NY Times        &       4,185   &      30,658 &      5.4    &      .0013 &         40 &          3 &      1 &     27,207 &       1.13 &      .8371 \\
EUR-Lex         &       3,993   &      19,800 &      5.3    &      .0013 &         26 &          6 &      1 &     16,871 &       1.17 &      .7548 \\ [.5 ex]
\hline
\end{tabular*}}
\caption{Statistics of the experimental datasets.  Traditional benchmark datasets are presented in the first three rows, and datasets with power-law-like statistics are presented in the last two rows.}
\label{table:datastats}
\end{table}

Table~\ref{table:datastats} presents the statistics for the datasets considered in this paper.  In addition to several statistics that have been previously presented in the multi-label literature, we present additional statistics which we believe help illustrate some of difficulties with classification for large scale power-law datasets. All statistics are explained in detail below:

{\small
\begin{itemize*} %\addtolength{\itemsep}{-0.5\baselineskip}

\item \textsc{Cardinality} : The average number of labels per document\\ [-5pt]

\item \textsc{Density} : The average number of labels per document divided by the number of unique labels (i.e., the cardinality divided by $C$), or equivalently, the average number of documents per label divided by the number of documents (i.e., Mean Label-Frequency divided by $d$)\\ [-5pt]

\item \textsc{Label Frequency (Mean, Median, and Mode)} : The mean, median, and mode of the distribution of the number of documents assigned to each label.  \\ [-5pt]

\item \textsc{Distinct Label Sets}: The number of distinct combinations of labels that occur in documents.\\ [-5pt]

\item \textsc{Label-set Frequency (Mean)} : The average number of documents per distinct combination of labels (i.e.,  $D$ divided by Distinct Label-sets). \\ [-5pt]

\item \textsc{Unique Label-set Proportion} : The proportion of documents containing a unique combination of labels.\\ [-5pt]
    \end{itemize*}
}

The cardinality of a dataset reflects the degree to which a dataset is truly multi-label (a single-label classification corpus will have a cardinality $= 1$).  The density of a dataset is a measure of how frequently a label occurs on average. The mean, median, and mode for label frequency reflects how many training examples exist for each label (see also Figure 1). All of these statistics reflect the sparsity of labels, and are clearly quite different among the two groups of datasets.

The last three measures in the table relate to the notion of label combinations.  For example, the label-set proportion tells us the average number of documents that have a unique combination of labels, and the label-set frequency tells us on average how many examples we have for each of these unique combinations.  These types of measures are particularly relevant to the issue of dealing with label dependencies.  For example, one approach to handling label-dependencies is to build a binary classifier for each unique {\it set} of labels \citep[e.g., this approach is described as the ``Label Powerset'' method in][]{Tsoumakas_Etc_2009}.  For the three smaller datasets, there is a relatively low proportion of documents with unique combinations of labels, and in general numerous examples of each unique combination.  Thus, building a binary classifier for each combination labels of could be a reasonable approach for these datasets.  On the other hand, for the NYT and EUR-Lex datasets these values are both close to $1$, meaning that nearly all documents have a unique set of labels, and thus there would not be nearly enough examples to to build effective classifiers for label-combinations on these datasets.

\section{Experiments}

In this section we introduce the prediction tasks and evaluation metrics used to evaluate model performance for the three LDA-based models and two SVM methods.
The results of all evaluations described in this section--which are performed on the five datasets shown in Table~\ref{table:datastats}--will be presented in the following section.
The objectives of these experiments were (1) to compare the Dependency-LDA model to the simpler LDA-based models (Prior-LDA and Flat-LDA), (2) to compare the performance of the LDA-based models with SVM-based models, and (3) to explore the conditions under which LDA-based models may have advantages over more traditional discriminative methods, with respect to both prediction tasks and to the dataset statistics.

Before delving into the details of our experiments, we first describe the binary SVM classifiers we implemented for comparisons with our LDA-based models.

\subsection{Implementation of Binary SVM Classifiers} \label{sec:SVM_Implementation}

In both of our SVM approaches we used a ``one-vs-all'' (sometimes referred to as ``one-vs-rest'') scheme, in which a binary Support Vector Machine (SVM) classifier was independently trained for each of the $C$ labels.  Documents were represented as a normalized vector of word counts, and SVM training was implemented using the LibLinear version 1.33 software package  \citep{REF08a}.

For ``Tuned-SVMs'', we followed the approach of \cite{Lewis_Etc_2004} for training $C$ binary support vector machines (SVMs). All parameters except the weight parameter for positive instances were left at the default value. In particular, we used an L2-loss SVM with a regularization parameter of $1$. The weight parameter for negative instances was kept at the default value of $1$. The weight parameter for positive instances ($\mathrm{w1}$) was determined using a hold-out set. The weight parameters alter the penalty of a misclassification for a certain class. This is especially useful for labels with small support where it is often desirable to penalize misclassifying a positive instance more heavily than misclassifying a negative instance \citep{JapkowiczStephen_2002}. The parameter $\mathrm{w1}$ was selected from the following values: \begin{equation*}
\{1, 2, 5, 10, 25, 50, 100, 250, 500, 1000, w_c\}
\end{equation*}
The last value, $w_c$, is a ratio of the number of negative instances to the number of positive instances in the training set for label $c$.  If there are an equal number of negative and positive instances then $w_c = 1$.

The hold-out set consisted of $10\%$ of the positive instances and $10\%$ of the negative instances from the training set. If a label had only one positive instance it was included in both the training set and the hold-out set. The weight value that had the highest accuracy on the hold-out set was selected. If there was a tie, the weight value closest to $1$ was chosen. Once the best value of $\mathrm{w1}$ was determined, the final SVM was re-trained on the entire training set.

We additionally provide results for ``Vanilla SVMs'', which were generated using LibLinear with default parameter settings (the default parameter value for $\mathrm{w1}$ was $1$) for all labels.

\subsection{Multi-Label Prediction Tasks}
Numerous prediction tasks and evaluation metrics have been adopted in the multi-label literature \citep{Sebastiani_2002, Tsoumakas_Etc_2009, CarvalhoFreitas_2009}.
There are two broad perspectives on how to approach multi-label datasets: (1) \textit{document-pivoted} (also known as \textit{instance-based} or \textit{example-based}), in which the focus is on
generating predictions for each test-document, and (2) \textit{label-pivoted} (also known as \textit{label-based}), in which the focus is on
generating predictions for each label.   Within each of these classes, there are two types of predictions that we can consider: (1) \textit{binary} predictions, where the goal is to make a strict yes/no classification about each test item, and (2) \textit{ranking} predictions, in which the goal is to rank relevant cases above irrelevant cases.  Taken together, these choices comprise four different prediction tasks that can be used to evaluate a model, providing an extensive basis for comparing LDA and SVM-based models.

Figure~\ref{fig:DataPivoting} illustrates the relationship between both the \textit{label-pivoted} vs. \textit{document-pivoted} and the \textit{binary} vs. \textit{ranking} tasks.  In order to produce as informative and fair a comparison of the LDA-based and SVM-based models as possible, we considered both ranking-predictions and binary-predictions for both the document-pivoted and label-pivoted prediction tasks.

Traditionally, multi-label classification has emphasized the label-pivoted binary classification task, but increasingly there has been growing interest in performance on document-pivoted
ranking \citep[e.g., see][]{Har-Peled_Etc_2002_constraintclass,CrammerSinger_2003,MenciaFurnkranz_2008,MenciaFurnkranz_2008_Efficient} and binary predictions \citep[e.g., see][]{Furnkranz_Etc_2008}. To calibrate our results with respect to this literature,
we adopt many of the ranking-based evaluation metrics used in this literature in addition to the more traditional metrics based on ROC-analysis.
We also provide results which can be compared with values that have been
published in the literature (although this is often difficult, due to the dearth of published results for large multi-label datasets and the variability of different versions of benchmark datasets, as well as the lack of consensus over evaluation metrics and prediction tasks).
Appendix~\ref{sec:Comparisons_PublishedResults} contains a detailed discussion of how our results compare to earlier results reported in the literature.

\begin{table}[t]
{\scriptsize \begin{tabular}{cccccccccccccc}
&                                                  \multicolumn{ 6}{c}{{\bf Binary}} &            &            &                        \multicolumn{ 5}{c}{{\bf Ranking-Based}} \\ [.5 ex]
           &            &                             \multicolumn{ 5}{c}{Label-Pivoted} &            &            & \multicolumn{ 2}{c}{Document-Pivoted} &            & \multicolumn{ 2}{c}{Label-Pivoted} \\ [.5 ex]
           &            &   {\bf c1} &   {\bf c2} &   {\bf c3} &   {\bf c4} &   {\bf c5} &            &            &   {\bf d1:} & $\{c_1,c_2,c_3 | c_4,c_5\}$ &            &   {\bf c1:} & $\{d_1,d_2 | d_3\}$ \\
\multicolumn{ 1}{c}{}  &   {\bf d1} &          + &          + &          + &          - &          - &            &            &  {\bf d2:} & $\{c_1,c_3,c_4 | c_2,c_5 \}$ &            &   {\bf c2:} & $\{d_1,d_3 | d_2\}$ \\
\multicolumn{ 1}{c}{Document-Pivoted}&   {\bf d2} &          + &          - &          + &          + &          - &            &            &   {\bf d3:} & $\{c_2,c_5 | c_1,c_3,c_4\}$ &            &   {\bf c3:} & $\{d_1,d_2 | d_3\}$ \\
\multicolumn{ 1}{c}{} &   {\bf d3} &          - &          + &          - &          - &          + &            &            &            &            &            &   {\bf c4:} & $\{d_2 | d_1,d_3\}$ \\
           &            &            &            &            &            &            &            &            &            &            &            &   {\bf c5:} & $\{d_3 | d_1,d_2\}$ \\
\end{tabular}}
\caption{Illustration of the relationship between the two prediction tasks (binary predictions vs. rankings), for both the label-pivoted and document-pivoted perspectives on multi-label datasets.  The table on the left shows the ground-truth for a toy dataset with three documents and five labels. For binary predictions, the goal is to reproduce this table by making hard classifications for each label or each document (for example, a perfect document-pivoted binary prediction for document $d1$ assigns a positive prediction `{\bf+}' to labels $c_1$, $c_2$ and $c_3$, and a negative prediction `{\bf-}' to labels $c_4$ and $c_5$).  For ranking-based predictions, one ranks all items for each test-instance and the goal is to rank relevant items above irrelevant items (for example, a perfect document-pivoted ranking for document $d1$ is any predicted ordering in which labels $c_1$, $c_2$ and $c_3$ are all ranked above $c_4$ and $c_5$).  In the notation used for this illustration, the vertical bar `$|$' indicates the ranking which partitions positive and negative items; thus, any permutation on the order of the items between a vertical-bar `$|$' and a bracket is equivalent from an accuracy viewpoint (since there is no ground truth about the relative values within the set of true labels or within the set of false labels)}
\label{fig:DataPivoting}
\end{table}

\subsection{Rank-based Evaluation Metrics}

On the label-ranking task, for each test document we predict a ranking of all $C$ possible labels, where the broad goal is to rank the relevant labels (i.e., the labels that were assigned to the document) higher than the irrelevant labels (the labels that were not assigned to the document)\footnote{For simplicity, we describe the rank-based evaluation metrics in terms of the document-pivoted rankings.  However, we also use these metrics for evaluating label-pivoted rankings (where the goal is to predict a ranking of all $D$ documents, for each label).}.  We consider several evaluation metrics that are rooted in ROC-analysis, as well as measures that have been used more recently in the label-ranking literature.
We provide a general description of these measures below \citep[more formal definitions of these measures can be found in, e.g.,][]{CrammerSinger_2003}\footnote{In order to provide results consistent with published scores on the EURLex dataset we use the same $[0,100]$ scaling used by ~\citet{MenciaFurnkranz_2008} of the last four measures}.
For each measure, the range of possible values is given in brackets, and the best possible score is in bold:

{\small
\begin{itemize*} %\addtolength{\itemsep}{-0.5\baselineskip}

\item \textsc{$\mathrm{AUC_{ROC}}$ [ $0-\textbf{1}$ ]} : The area under the ROC-curve.  The ROC-curve plots the false-alarm rate versus the true-positive rate for each document as the number of positive predictions changes from $0-C$.  To combine scores across documents we compute a macro-average (i.e. the $\mathrm{AUC_{ROC}}$  is first computed for each document and is then averaged across documents).\\ [-5pt]

\item \textsc{$\mathrm{AUC_{PR}}$ [ $0-\textbf{1}$ ]} : The area under the precision-recall curve\footnote{Although the area under the ROC curve is more traditionally used in ROC-analysis, \citet{David_Goadrich_2006_AUCPR} demonstrated that the area under the Precision-Recall curve is actually a more informative measure for imbalanced datasets}.  This is computed for each document using the method described in \citet{David_Goadrich_2006_AUCPR}, and scores are combined using a macro-average. \\ [-5pt]

\item \textsc{Average Precision [ $0-\textbf{1}$ ]} : For each relevant label $x$, the fraction of all labels ranked higher than $x$ which are correct.  This is first averaged over all relevant labels within a document and then averaged across documents. \\ [-5pt]

\item \textsc{One-Error [ $\textbf{0}-100$} ] : The percentage of all documents for which the highest-ranked label is incorrect.\\ [-5pt]

\item \textsc{Is-Error [ $\textbf{0}-100$} ] : The percentage of documents without a \textit{perfect} ranking (i.e., the percentage of all documents for which all relevant labels are not ranked above all irrelevant labels.\\ [-5pt]

\item \textsc{Margin [ $\textbf{1}-C$} ] : The difference in ranking between the highest-ranked irrelevant label and the lowest ranked relevant label, averaged across documents. \\ [-5pt]

\item \textsc{Ranking Loss [ $\textbf{0}-100$} ]: Of all possible comparisons between the rankings of a single relevant label and single irrelevant label, the percentage of these that are incorrect.  First averaged across all comparisons within a document, then across all documents. \footnote{We note that the \textsc{Ranking Loss} statistic corresponds to the complement of the area under the ROC curve (scaled):  $\textsc{RankLoss}=100\times(1-\mathrm{AUC_{ROC}})$, which, furthermore is equivalent to the Mann-Whitney U statistic.  To simplify comparisons with published results, we present the results in terms of both the Ranking Loss and
    the $\mathrm{AUC_{ROC}}$.}  \\ [-5pt]
\end{itemize*}
}

\subsection{Binary Prediction Measures}

The basis of all binary prediction measures that we consider are macro-averaged and micro-averaged F1 scores (\textit{Macro-F1} and \textit{Micro-F1}) \citep{Yang_1999, Tsoumakas_Etc_2009}. Traditionally, the literature has emphasized the label-pivoted perspective, in which F1 scores are first computed for each label and then averaged across labels.  However, recently there has been an increased interest in binary predictions on a per-document basis ~\citep[e.g., see][who refer to this task as {\it calibrated label-ranking}]{Furnkranz_Etc_2008}.  We consider both the document-pivoted and label-pivoted approaches to the evaluation of binary predictions.

The F1 score for a document $d_{i}$, or a label $c_{i}$, is the harmonic mean of precision and recall of the set of binary predictions for that item.  Given the set of $C$ binary predictions for a document, or the set of $D$ binary predictions for a label, the F1-score is defined as:

\begin{equation}
\label{eq:EvalMetric_F1}
F1(i) =
\frac{2 \times Recall(i) \times Precision(i)}
{Recall(i) + Precision(i)}
\end{equation}

\noindent After computing the F1 scores for all items, the performance can be summarized using either {\it micro-averaging} or {\it macro-averaging}.  In macro-averaging, one first computes an F1-score for each of the individual test items using its own confusion matrix, and then takes the average of the F1-scores.  In micro-averaging, a single confusion matrix is computed for all items (by summing across the individual confusion matrices), and then the F1-score is computed for this single confusion matrix.  Thus, the micro-average gives more weight to the items that have more positive test-instances (e.g., the more frequent labels), whereas the macro-average gives equal weight to each item, independent of its frequency.

We note that one must be careful when interpreting F1-scores, since these measures are very sensitive to differences in dataset statistics as well as to differences in model performance.
As the label frequencies become increasingly skewed (as in the power-law datasets like NY Times and EUR-Lex), the potential disparity between the Macro-F1 and Micro-F1 becomes increasingly large; a model that performs well on frequent labels but very poorly on infrequent labels (which are in the vast majority for a power-law dataset) will have a poor Macro-F1 score but can still have a reasonably good Micro-F1 score.

\subsection{Binary Predictions and Thresholding}
As illustrated in Table~\ref{fig:DataPivoting}, a binary-prediction task can be seen as a direct extension of a ranking task.  If we have a classifier that outputs a set of real-valued predictions for each of the test instances, then a predicted ranking can be produced by sorting on the prediction values. We can transform this ranking into a set of binary predictions by either
\begin{inparaenum}[(1)]
\item learning a threshold on the prediction values, above which all instances are assigned a positive prediction (e.g. the `SCut' method \citep{Yang_2001} is one example of this approach), or
\item making a positive prediction for the top $N$ ranked instances for some chosen $N$.
\end{inparaenum}

The issue of choosing a threshold-selection method is non-trivial (particularly for large-scale datasets) and threshold selection comprises a significant research problem in and of itself \citep[e.g., see][]{Yang_2001,FanLin_2007,Ioannou_EtAl_2011_Bipartitions}.
Since threshold-selection is not the emphasis of our own work, and we do not wish to confound differences in the models with the effects of thresholding, we followed a similar approach to that of \citet{Furnkranz_Etc_2008} and considered several rank-based cutoff approaches\footnote{Note that the cutoff-points we use are slightly different from those presented in \citet{Furnkranz_Etc_2008}.  In particular, since our models are not learning a calibrated cutoff during inference, we substituted their {\sc Predicted} method with the more traditional {\sc Break-Even-Point} (BEP) method.  Additionally, our {\sc Proportional} cutoff has been modified from the {\sc Median} approach that they use in order to extend it to the label-pivoted case, since the median value is generally not applicable for label-pivoted predictions.}.
The three rank-cutoff values which we consider are:

{\small
\begin{enumerate*}
    \item {\sc Proportional:} Set $\hat{N_{i}}$ equal the expected number of positive instances for item $i$, based on training-data frequencies: \\ [-5pt]
    \begin{itemize*}
        \item For label $c_{i}$ \ (i.e., label-pivoted predictions):
        {\footnotesize $\hat{N_{i}} = \mathrm{ceil}  \left(\frac{D^{TEST}}{D^{TRAIN}} * N^{TRAIN}_{i}\right)$}
         , where $N^{TRAIN}_{i}$ is the number of training documents assigned label $c_{i}$, and $D^{TRAIN}$ and $D^{TEST}$ are the total number of documents in the training and test sets, respectively.\footnote{For label-pivoted predictions, SVMs do in fact learn a threshold which partitions the data during training, unlike the LDA models.  However, we found that in most cases the performance at these thresholds is much worse than performance using the {\sc Proportional} method (this is particularly true on the power-law datasets, due to the difficulties with learning a proper SVM model on rare labels).  This is consistent with results that have been noted previously in the literature--e.g., see~\citet{Yang_2001}.} \\ [-5pt]
        \item For test document $d_i$ \ (i.e., document-pivoted predictions):
        $\hat{N_{i}} = \mathrm{{median}}(N^{TRAIN}_{d})$ where $N^{TRAIN}_{d}$ is the number of labels for training document $d$.\\ [-5pt]
    \end{itemize*}
    \item {\sc Calibrated:} Set $\hat{N_{i}}$ equal to the true number of positive instances for item $i$. \\ [-5pt]
    \item {\sc Break-Even-Point (BEP):} Set $\hat{N_{i}}$ such that it optimizes the F1-score for that item, given the predicted order.  This method is commonly referred to as the {\it Break-Even Point} (BEP) because it selects the location on the Precision-Recall curve at which $Precision = Recall$. \\ [-5pt]
\end{enumerate*}
}

\noindent  Note that the latter two methods both use information from the test set, and thus do not provide an accurate representation of performance we would expect for the models in a real-world application.  However, in addition to the practical value of these methods for model comparison, they each provide measures of model performance at points of theoretical interest:  The {\sc Calibrated} method gives us a measure of model performance if we assume that there is some external method (or model) which tells us the {\it number} of positive instances, but not {\it which} of these instances are positive.  The {\sc BEP} method (which has been commonly employed in multi-label classification literature) tells us the highest attainable F1-score for each item given the predicted ordering.  Thresholding methods which attempt to maximize the macro-averaged F1 score are in fact searching for a threshold as close to the BEP as possible.  Note that although the BEP provides the highest possible macro-F1 score on a dataset, this does not mean that it will optimize the Micro-F1 score; in fact, since the method optimizes the F1-score for each label independently, it will generate a large number of false-positives when the predicted ordering has assigned the actual positive instances a low rank, which can have large negative impact on Micro-F1 scores.

We additionally point out that whereas the {\sc BEP} method will vary the number of positive predictions to account for a model's specific ranking, the {\sc Proportional} and {\sc Calibrated} methods will produce the same number of positive predictions for all models.  Thus, scores on these predictions reflect model performance at a fixed cutoff point which is independent of the model's ranking.

\section{Experimental Results}

Results below are organized as follows:  (1) document-pivoted results on all datasets for (a) ranking-predictions and (b) binary-predictions, and then (2) label-pivoted results on all datasets for (a) ranking-predictions and (b) binary- predictions.  For completeness, we provide a table for each of the four tasks using all evaluation metrics and datasets.

\subsection{Document-Pivoted Results}

\begin{figure}[t] % float placement: (h)ere, page (t)op, page (b)ottom, other (p)age
  \centering
  \includegraphics[width=1\linewidth]{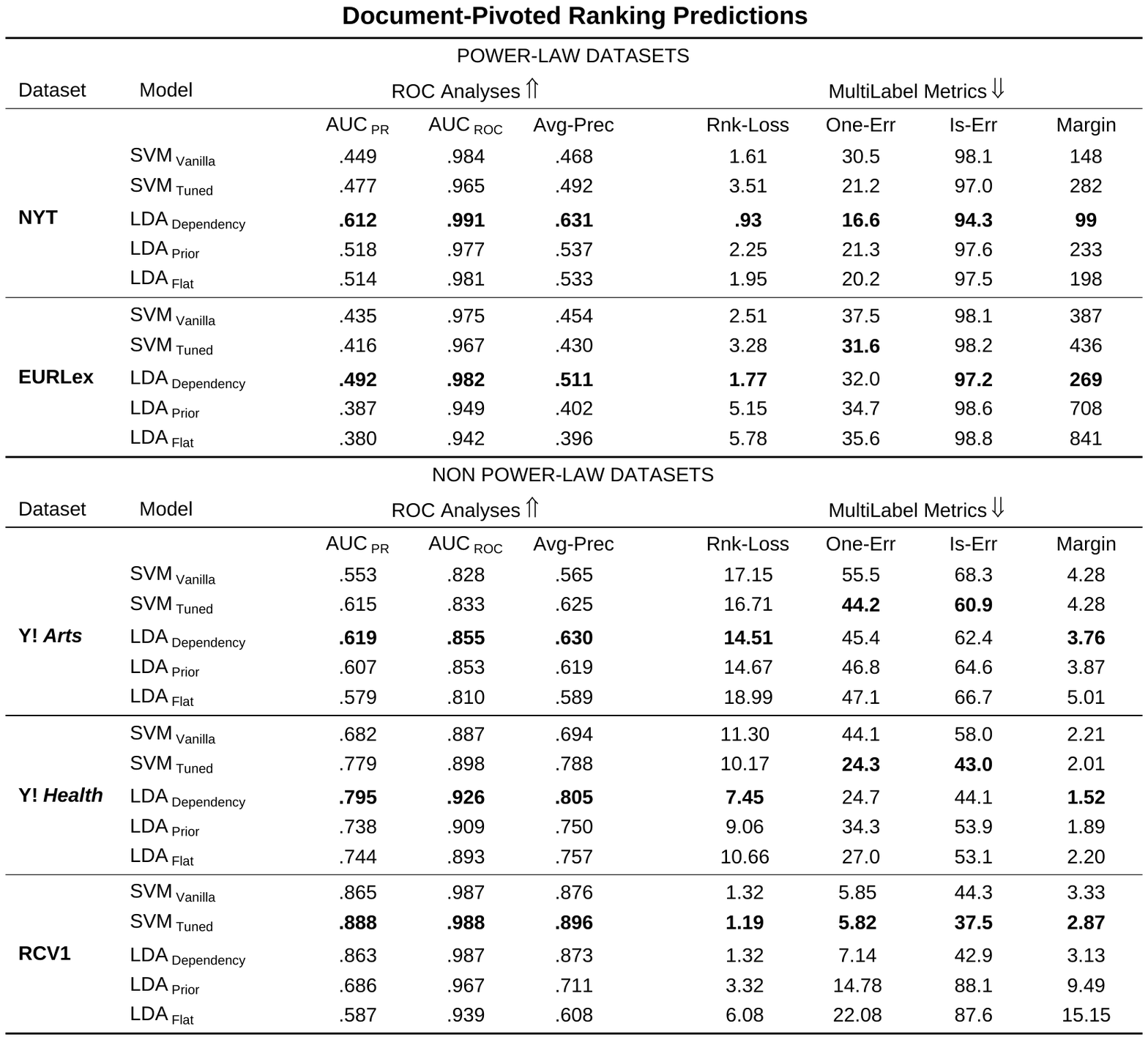}
  \caption{Document-Pivoted-Ranking-Predictions. For each dataset and model, we present scores on all rank-based evaluation metrics.  These have been grouped in accordance with how they are used in the literature (where the first three evaluation metrics are used in ROC-analysis literature, and the remaining four metrics are used in used in the label-ranking literature).  We note again that $\textsc{RankLoss}=100 \times (1-AUC_{ROC})$; we provide results for both metrics for ease of comparison with published results}
  \label{fig:Document-Pivoted-Ranking-Predictions}
\end{figure}

\begin{figure}[t] % float placement: (h)ere, page (t)op, page (b)ottom, other (p)age
  \centering
  \includegraphics[width=1\linewidth]{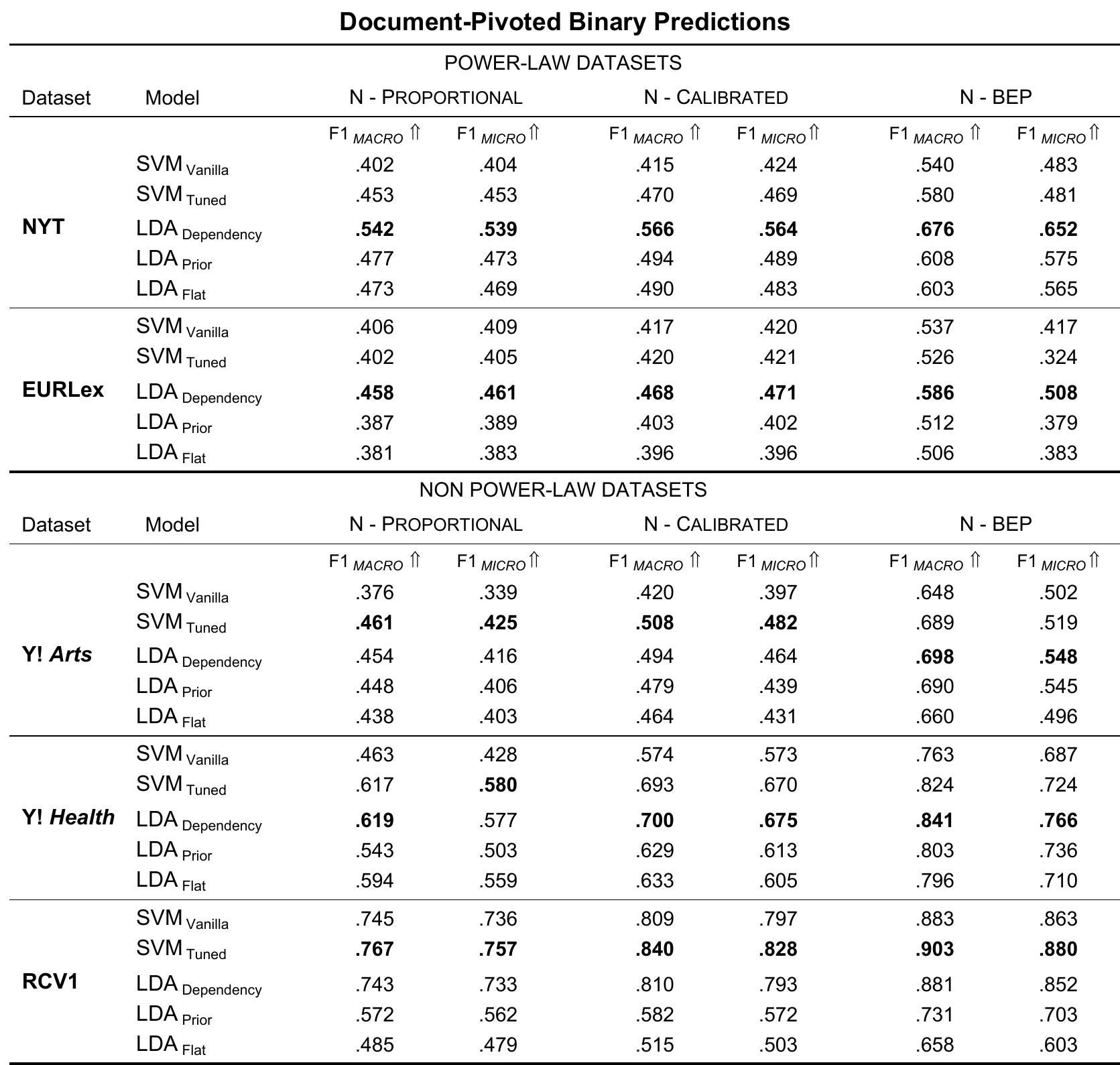}
  \caption{Document-pivoted binary predictions.   For each dataset and model, we present the Micro- and Macro-F1 scores achieved using the three different cutoff-point methods (from left to right: {\sc Proportional}, {\sc Calibrated}, and {\sc BEP}).  Note that the absolute difference between the Micro and Macro scores for a model are generally smaller for the document-pivoted results than for the label-pivoted evaluations; this is due to the relatively low variability in the number of labels per document (as opposed to the generally large variability in the number of documents per label).}
  \label{fig:Document-Pivoted-Binary-Predictions}
\end{figure}

The document-pivoted predictions provide a ranking of all labels in terms of their relevance to each test-document $d$. The seven {\it ranking}-based metrics directly evaluate aspects of each of these rankings. The six {\it binary} metrics evaluate the binary predictions after these rankings have been partitioned into positive and negative labels for each document, using the three aforementioned cutoff-points. Results for the rank-based evaluations are shown in Figure~\ref{fig:Document-Pivoted-Ranking-Predictions}, and results for the binary predictions are shown in Figure~\ref{fig:Document-Pivoted-Binary-Predictions}.

\subsubsection{Comparison within LDA-based and SVM-based models (Doc-Pivoted)}

Among the LDA-based models, Dependency-LDA performs significantly better than both Prior-LDA and Flat-LDA across all datasets on all 13 evaluation metrics across Figures~\ref{fig:Document-Pivoted-Ranking-Predictions} and~\ref{fig:Document-Pivoted-Binary-Predictions}.  For the simpler LDA models, Prior-LDA outperformed Flat-LDA on the EUR-Lex (12/13), Yahoo! {\it Arts} (13/13) and RCV1-v2 (12/13) datasets whereas performance on NYT and Yahoo! {\it Health} was more evenly split. In almost all cases, the absolute differences between the Prior-LDA and Flat-LDA scores is much smaller than the differences between either of them and Dependency-LDA.  The scale of the differences between the three LDA-based models demonstrates that Dependency-LDA is achieving its improved performance by successfully incorporating information beyond simple baseline label-frequencies.

Among the SVM models, the Tuned-SVMs convincingly outperform Vanilla-SVMs on the three non power-law datasets.  On NYT, the Tuned-SVMs generally outperformed Vanilla-SVMs (9/13), whereas they performed worse on the EUR-Lex dataset (3/12).  Generally, in the cases in which there were significant differences between the two SVM approaches on the power-law datasets, the Vanilla-SVMs outperformed Tuned-SVMs on measures that emphasize the full range of ratings (such as the \textsc{Margin} and the  Areas Under Curves), whereas Tuned-SVMs outperformed Vanilla-SVMs on metrics emphasizing the top-ranked predictions (such as the {\sc One-Error} and {\sc IsError} metrics).  This overall pattern indicates that Tuned-SVMs may generally make better predictions among the top-ranked labels but have difficulty calibrating predictions for the lower-ranked labels (which will in general be largely comprised of infrequent labels).  Thus, the observed contrast between overall SVM performance on the EUR-Lex and NYT datasets may reflect the fact that predictions for the NYT dataset were evaluated on only labels that showed up in test documents (thereby excluding many of the infrequent labels from these rankings), whereas predictions for EUR-Lex were evaluated across {\it all} labels. This observation is supported by performance of the SVMs on the benchmark datasets, on which Tuned-SVMs clearly outperform Vanilla-SVMs; on these datasets, there are many fewer total labels to rank, and a much higher percentage of these labels is present in the test-documents, so therefore the scores on these datasets are much less influenced by low-ranked labels.

\subsubsection{Comparison between LDA-based and SVM-based models (Doc-Pivoted)}

Looking across all document-pivoted model results, one can see a clear distinction between the relative performance of LDA and SVMs on the power law datasets vs. the non power-law datasets.  The Dependency-LDA model clearly outperforms SVMs on the power-law datasets (on 13/13 measures for NYT, and on 12/13 measures on EUR-Lex).
Note that on the NYT dataset, which has the most skewed label-frequency distribution and the largest cardinality, both the Prior-LDA and the Flat-LDA methods outperform the Tuned-SVMs as well.

On the non power-law datasets, results are more mixed.  For rank-based metrics on both of the Yahoo! datasets, Dependency-LDA outperforms SVMs on the five measures which emphasize the full range of rankings, but are outperformed by Tuned-SVM's on the measures emphasizing the very top-ranked labels (namely, the One-Error and Is-Error measures).  For binary evaluations, Dependency-LDA generally outperforms Tuned-SVMs on the {\it Health} dataset (5/6) but performs worse on the {\it Arts} dataset. On the RCV1 dataset, Tuned-SVMs have a clear advantage over all of the LDA models (outperforming them across all 13 measures).

\subsubsection{Relationship Between Model Performance and Dataset Statistics (Doc-Pivoted)}

\begin{figure}[t]
\centering
\includegraphics[width=.7\linewidth]{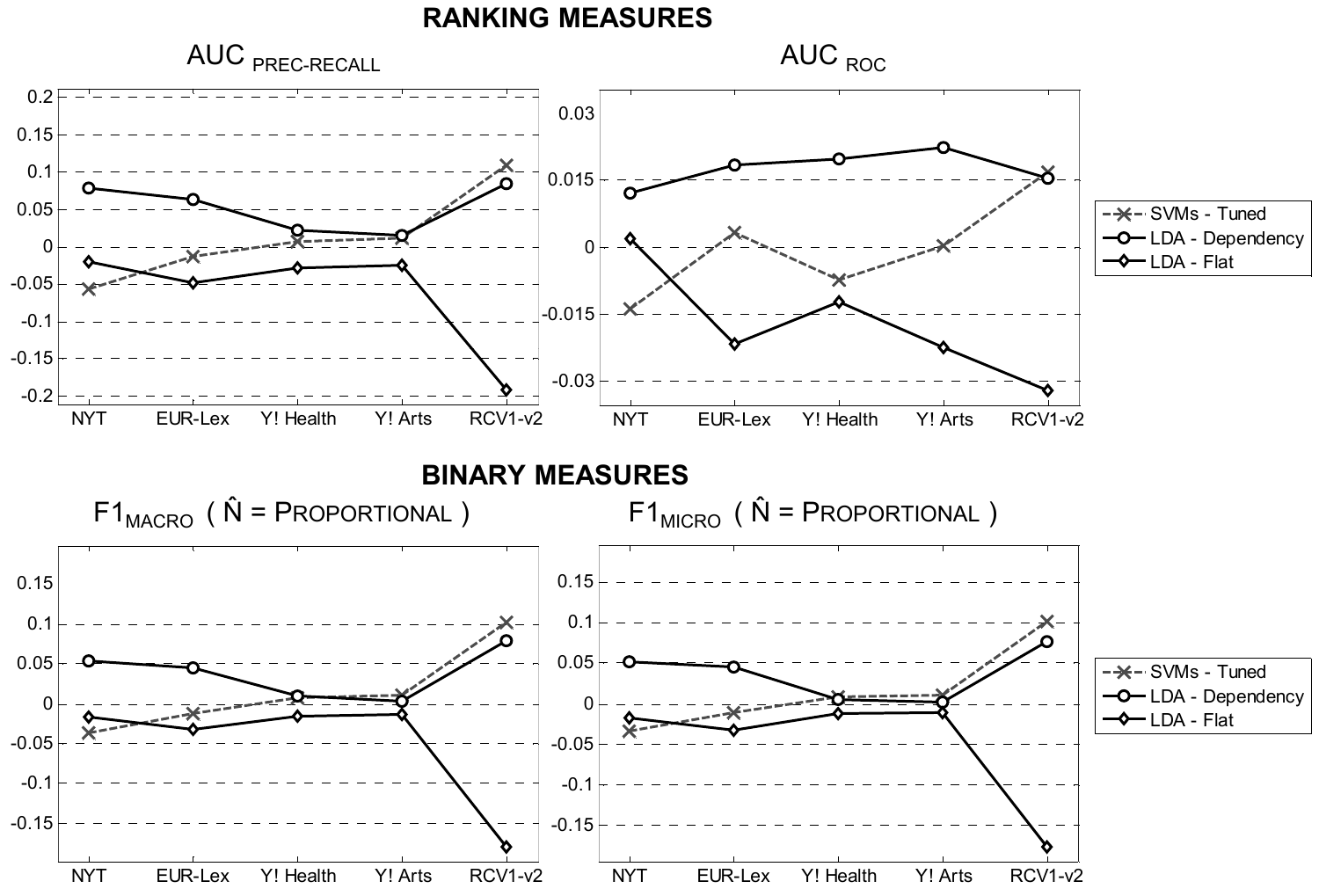}
\caption{{\bf Dataset Label-Frequency and Model Performance}:  Relative performance of Tuned-SVMs, Dependency-LDA, and Flat-LDA on several of the evaluation metrics for document-pivoted predictions.  Datasets are ordered in terms of their median label-frequencies (the median number of documents-per-label increases from left to right). Scores have been centered around zero in order for each dataset to emphasize {\it relative} performance of the models.  As the amount of training data per label decreases, performance for LDA-based models tends to improve relative to SVM performance}
\label{fig:DocPivoted_SortedByDataset_FREQ}
\end{figure}

\begin{figure}[t]
\centering
\includegraphics[width=1\linewidth]{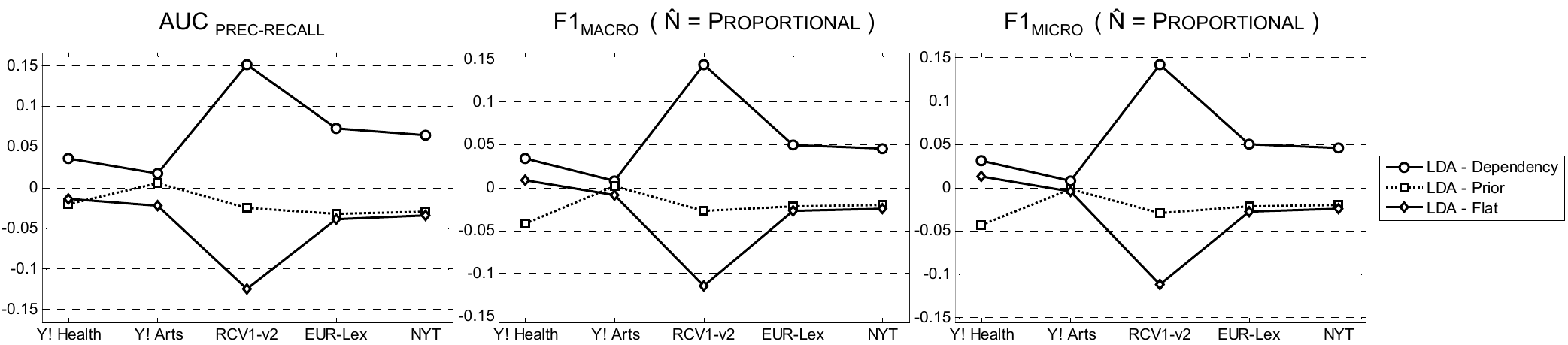}
\caption{{\bf Dataset Cardinality and Model Performance}:  Relative performance of the three LDA-based models on several of the evaluation metrics for document-pivoted predictions.  Datasets are ordered in terms of their cardinality (the average number of labels-per-document increases from left to right). Scores have been centered around zero in order for each dataset to emphasize {\it relative} performance of the models.  As the average number of labels-per document increases, the relative improvement of Dependency-LDA over the simpler models increases.}
\label{fig:DocPivoted_SortedByDataset_CARDINALITY}
\end{figure}

The overall pattern of results indicates that there is a strong interaction between the statistics of the datasets and the performance of LDA-based models relative to SVM-based models.
These effects are illustrated in Figures~\ref{fig:DocPivoted_SortedByDataset_FREQ} and~\ref{fig:DocPivoted_SortedByDataset_CARDINALITY}.  To help illustrate the {\it relative} performance differences between models,  the results within each dataset have been centered around zero in these figures  (without the centering, it is more difficult to see the interaction between the datasets and models, since most of the variance in model performance is accounted for by the main effect of the datasets).

In Figure~\ref{fig:DocPivoted_SortedByDataset_FREQ}, performance on each of the five datasets has been plotted in order of the dataset's median label-frequency (i.e., the median number of documents per label). One can see that as the amount of training data increases, the performance of Dependency-LDA {\it relative} to Tuned-SVMs drops off and eventually becomes worse.  A similar pattern exists for Flat-LDA. Note that although both LDA-based models are worse than Tuned-SVMs on the RCV1-v2 dataset (which has the most training data), Dependency-LDA performance is in the range of Tuned-SVMs, whereas Flat-LDA performs drastically worse.

Figure~\ref{fig:DocPivoted_SortedByDataset_CARDINALITY} plots the same results as a function of dataset Cardinality (i.e., the average number of labels per document).  Here, one can see that the relative performance improvement for Dependency-LDA over Flat-LDA increases as the number of labels per document increases.  Since both Flat-LDA and Dependency-LDA use the same set of label-word distributions learned during training, this performance boost can only be attributed to inference for Dependency-LDA at test time (where unlike Flat-LDA, Dependency-LDA accounts for the dependencies between labels).  These results are consistent with the intuition that it is increasingly important to account for label-dependencies as the number of labels per document increases.

\subsection{Label-Pivoted Results}

\begin{figure}[t] % float placement: (h)ere, page (t)op, page (b)ottom, other (p)age
  \centering
  \includegraphics[width=1\linewidth]{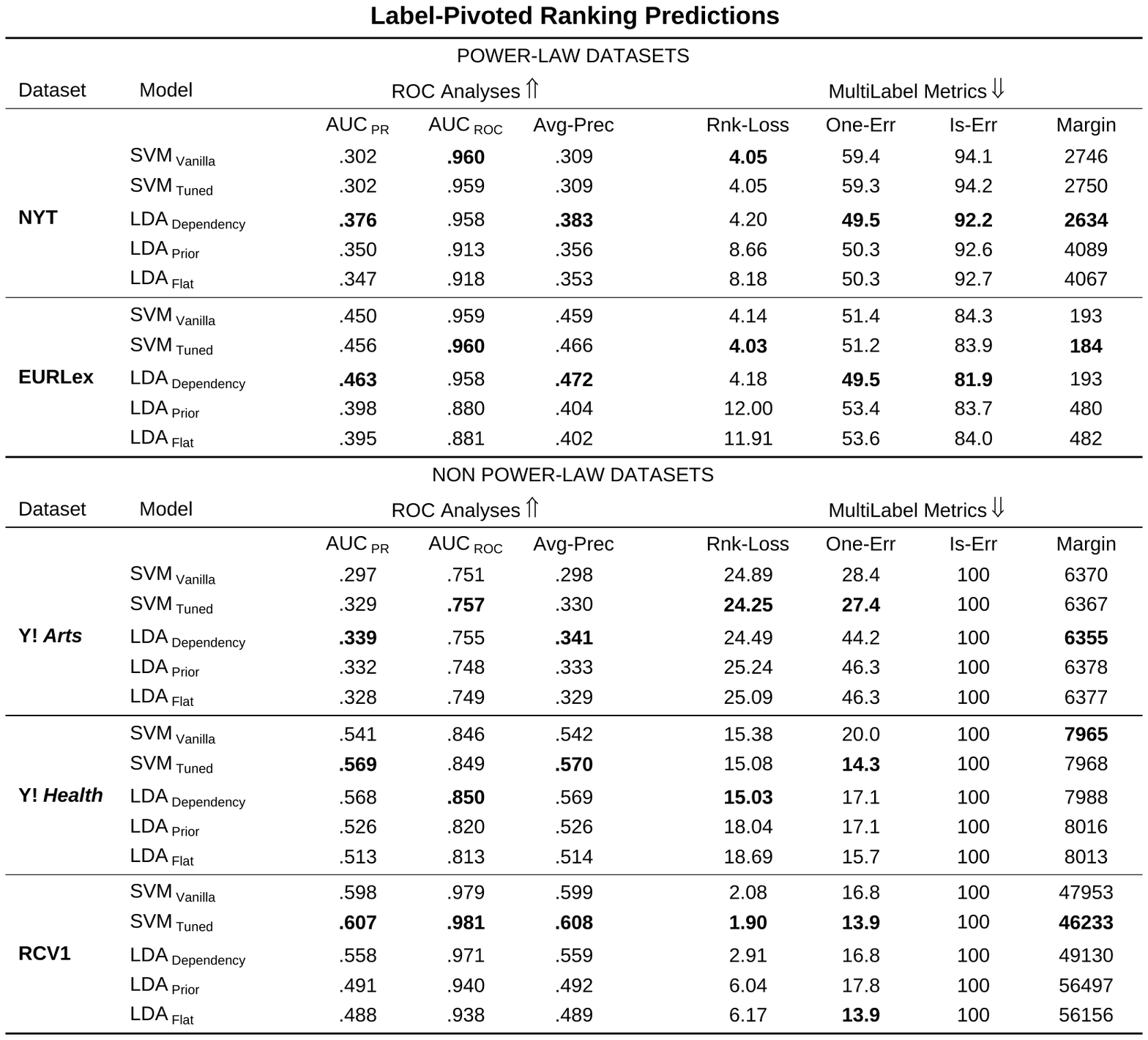}
  \caption{Label-Pivoted-Ranking-Predictions. For all rank-based evaluation metrics, we present results for the label-pivoted model predictions.  Note that the {\sc Is-Error} measure is not well-suited for the label-pivoted results on the non power-law datasets.  Specifically, since all labels have numerous test-instances, and the number of documents is very large, it is extremely difficult to predict a perfect ordering of all documents for {\it any} labels.  In fact, none of the models assigned a perfect ordering for a single label, which is why all scores are 100.  We have nonetheless included these results for completeness.}
  \label{fig:Label-Pivoted-Ranking-Predictions}
\end{figure}

\begin{figure}[t] % float placement: (h)ere, page (t)op, page (b)ottom, other (p)age
  \centering
  \includegraphics[width=1\linewidth]{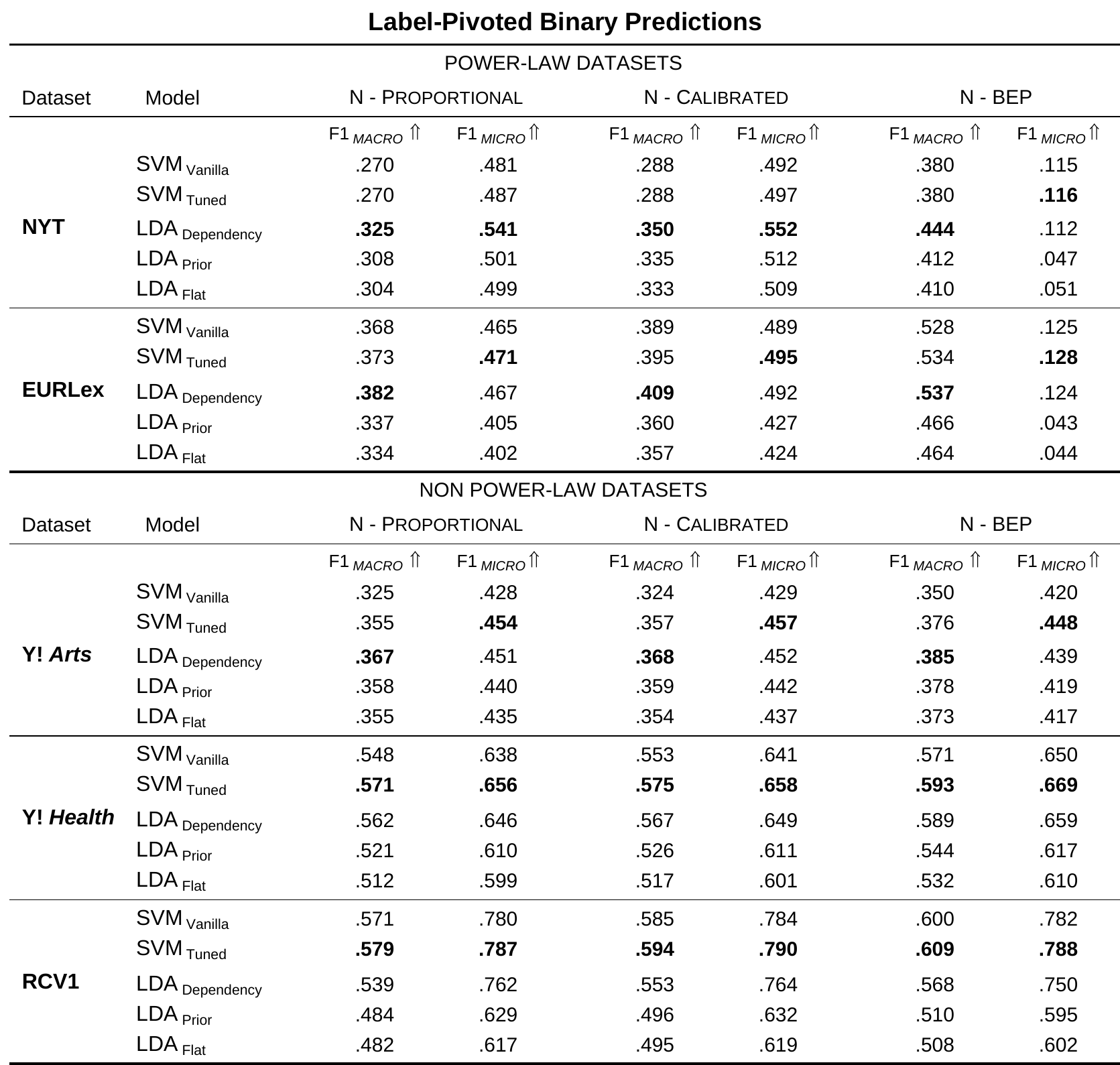}
  \caption{Label-Pivoted-Binary-Predictions.  For each set of results, we present the Micro-F1 and Macro-F1 scores achieved using the three different cutoff-point methods (From left to Right: {\sc Proportional}, {\sc Calibrated}, and {\sc BEP}).  Note that the only results representing {\it true} performance are the {\sc Proportional} results, and thus these are the values which should be used for comparison with benchmarks presented in the literature (although for model comparison, all values are useful, since they easily calculated from model output).}
  \label{fig:Label-Pivoted-Binary-Predictions}
\end{figure}

The label-pivoted predictions provide a ranking of all documents in terms of their relevance to each label $c$. The seven {\it ranking}-based metrics directly evaluate aspects of each of these rankings. The six {\it binary} metrics evaluate the binary predictions after the rankings have been partitioned into positive and negative documents for each label, using the three aforementioned cutoff-points. Results for the rank-based evaluations are shown in Figure~\ref{fig:Label-Pivoted-Ranking-Predictions}, and results for the binary predictions are shown in Figure~\ref{fig:Label-Pivoted-Binary-Predictions}.

\subsubsection{Comparison within LDA-based and SVM-based models (Label-Pivoted)}

The relative performance among the LDA-based models follows a similar pattern to what was observed for the document-pivoted predictions.  Dependency-LDA consistently outperforms both Prior-LDA and Flat-LDA, beating them on nearly every measure on all five datasets.

The improvement achieved by Dependency-LDA seems generally to be related to the number of labels per document; there is a very large performance gap in the power-law datasets (which have about 5.5 labels per document each on average), whereas this gap is relatively smaller on the Yahoo! datasets (which have on average 1.6 labels per document).  The improvement observed for RCV1 is nearly as large or even larger than for the power-law datasets, which may be in part due to the automated, rule-based assignment of labels in the dataset's construction (which introduces very strict dependencies in the true label-assignments).

Tuned-SVMs consistently outperformed Vanilla-SVMs on all datasets except for NYT, where the two methods show nearly equivalent performance overall. This is notable in that it indicates that the NYT dataset poses a problem for binary SVMs which parameter tuning cannot fix; in other words, it suggests that there is some feature of this dataset which binary SVMs have an intrinsic difficulty dealing with.  Since the straightforward answer---given what we have seen, as well as our motivations presented in the introduction---is that this difficulty relates to the power-law statistics of the NYT dataset, it is somewhat surprising that there is not a similar effect for the EUR-Lex dataset (on which the Tuned-SVMs outperform Vanilla-SVMs on all measures).  Why should these two datasets, both of which have fairly similar statistics, show different improvement due to parameter tuning?

We conjecture that the differences in the effect of parameter tuning between the NYT and EUR-Lex datasets are misleading.  First, although Tuned-SVMs achieve better scores on all measures for EUR-Lex, the {\it scale} of these differences is actually quite small.  Secondly, and perhaps more importantly, some of these differences are likely to be due to the relative proportion of training vs. testing data between the two datasets.  For EUR-Lex, only one-tenth of the documents are in each test-set, whereas NYT has a roughly 50-50 train-test split.  As a result, far fewer rare-labels are tested in any given split of EUR-Lex (since a label is only included in the label-wise evaluations if it appears in both the train and test-data).  Thus, the EUR-Lex splits somewhat de-emphasize performance on rare labels.  This assertion is strongly supported by the Document-pivoted results for EUR-Lex (in which {\it all} labels that appeared in the training set must be ranked, and thus, influence the performance scores); Tuned-SVMs perform worse than Vanilla SVMs on 10/13 of the Document-Pivoted evaluation metrics for EUR-Lex.  
Overall, it appears that the intrinsic difficulties that SVMs have on rare labels is a problem with both NYT and EUR-Lex, and that the observed differences between Tuned and Vanilla-SVMs on these two datasets is likely due in part to the differences in the construction of the datasets.

\subsubsection{Comparison between LDA-based and SVM-based models (Label-Pivoted)}
The performance of Dependency-LDA relative to SVMs was highly dependent on the dataset. On the power-law datasets, Dependency-LDA generally outperformed SVMs; Dependency-LDA outperformed Tuned-SVMs on 10/13 measures for the NYT dataset and on 7/13 measures for the EUR-Lex dataset.

Of special interest is the Macro-F1 measures since Macro-averaging gives equal weight to all labels (regardless of their frequency in the test set). Since power-law datasets are dominated by rare labels, the Macro-F1 measures reflect performance on rare labels. On EUR-Lex, Dependency-LDA outperforms the SVMs for all Macro-F1 measures. On NYT, all three LDA models outperform the SVMs for all Macro-F1 measures.  This supports the hypothesis---motivated in our introduction---that LDA is able to handle rare labels better than binary SVMs.

On the non power-law datasets, results were much more mixed, with SVMs generally outperforming Dependency-LDA.  Dependency-LDA was competitive with Tuned-SVMs for the {\it Arts} subset, but generally inferior in performance on the {\it Health} subset. Performance was even worse on the RCV1-v2 dataset where both SVM methods clearly outperformed all LDA-based methods. Some of the variability in performance on the three datasets may be due to the amount of training data per label. RCV1-v2 has the most training data per label (despite containing more labels) and on this dataset the SVM methods dominate the LDA methods. The {\it Arts} subset has the least amount of training data per label and on this dataset the LDA methods fair better.

Again, it is of interest that on the {\it Arts} subset Dependency-LDA dominates the SVM methods on the Macro-F1 measures. In fact, the {\sc Proportional} Macro-F1 scores for this dataset seem to be higher than any of the Macro-F1 scores previously reported in the literature (including the large set of results for discriminative methods published by ~\citet{JiTangYuYe_2008}, which includes a method that accounts for label-dependencies); see Appendix~\ref{sec:Comparisons_PublishedResults} for additional comparisons.

\subsection{Comparing Algorithm Performance across Label Frequencies}

\begin{figure}[t]
%\begin{minipage}[b]{0.5\linewidth}
\begin{center}
\subfigure[NYT]{\label{fig:fscore_nyt_sparsity_sub}\includegraphics[width=.47\linewidth]{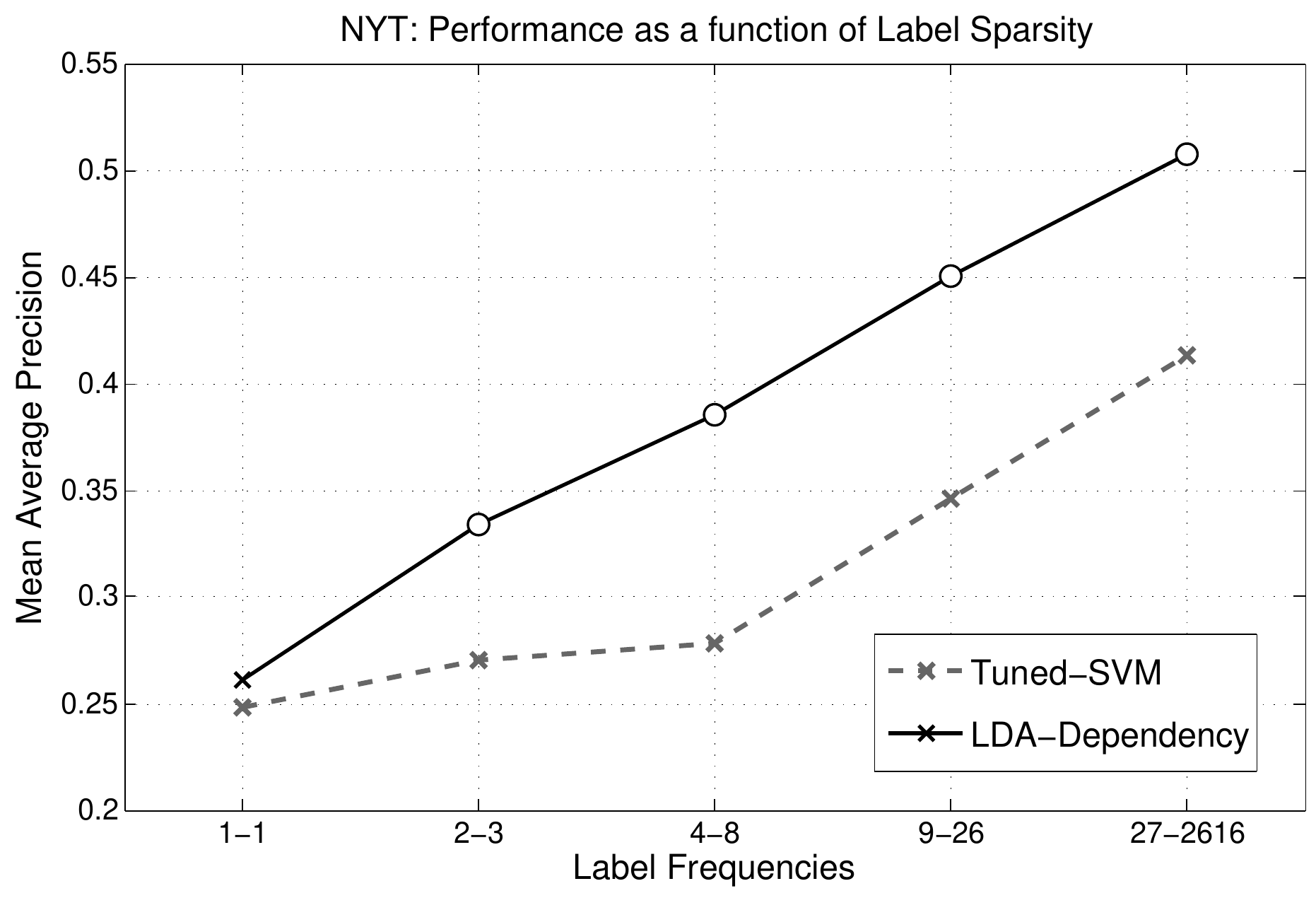}}
\subfigure[EUR-Lex]{\label{fig:fscore_eur_sparsity_sub}\includegraphics[width=.47\linewidth]{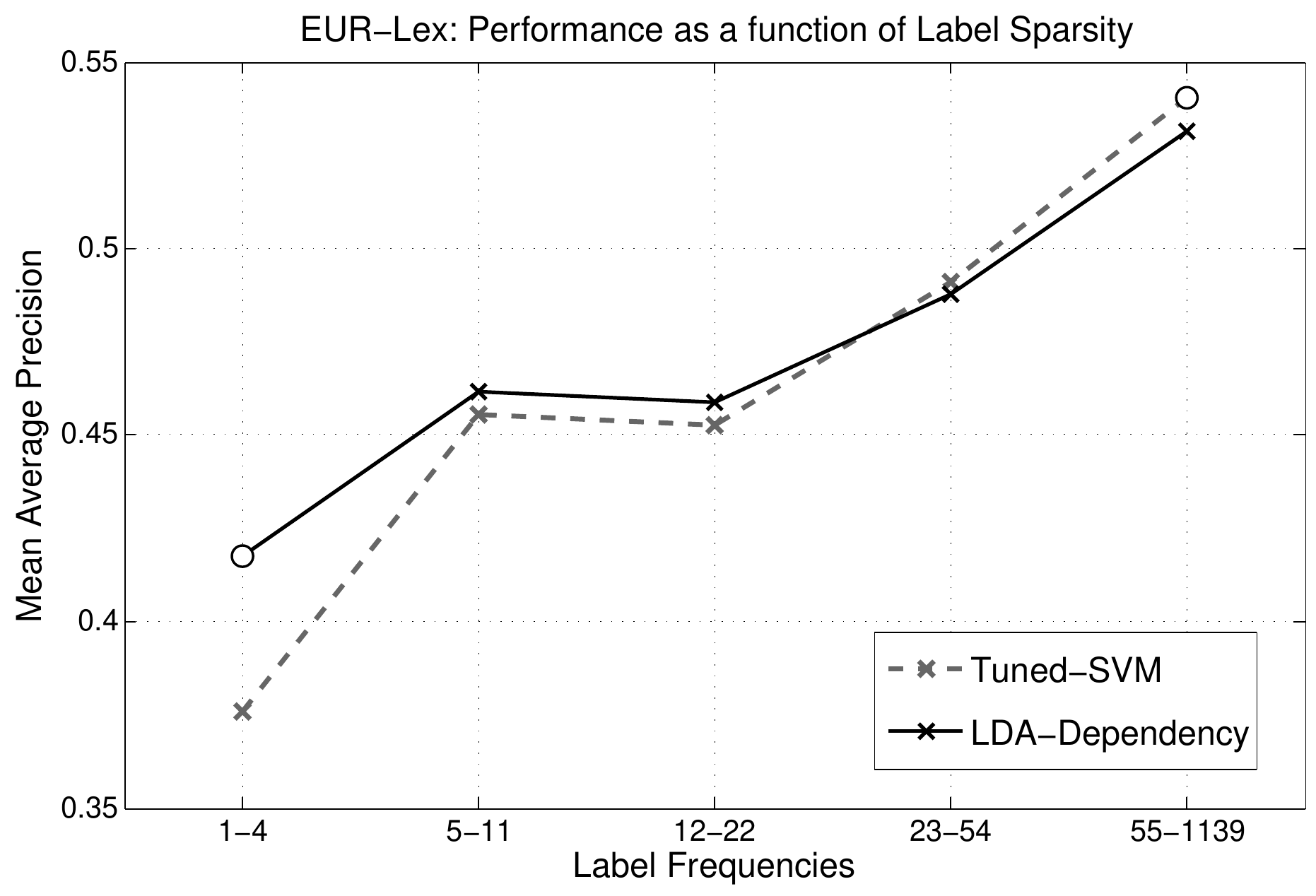}}
\caption{Mean Average Precision scores for the NYT and EUR-Lex datasets as a function of the number of training documents per label.  For each dataset, labels have been binned into quintiles by training frequency.  Performance scores are macro-averaged across all labels within each of the bins. Circle (`$\circ$') markers indicate where the differences were statistically significant at the  $\alpha=.05$ level as determined by pairwise t-tests within each bin.  (In all cases in which the difference was significant: $p<.001$).}
\end{center}
\end{figure}

As discussed in the introduction, there are reasons to believe that LDA-based models should have an advantage over one-vs-all binary SVMs on labels with sparse training data.  To address this question, we can look at the relative performance of the models as a function of the amount of training data.  Figures~\ref{fig:fscore_nyt_sparsity_sub} and ~\ref{fig:fscore_eur_sparsity_sub} compare the average precision scores for Dependency-LDA and Tuned-SVMs across labels with different training frequencies in the NYT and EUR-Lex datasets, respectively.  To compute these scores, labels were first binned according to their quintile in terms of training frequency, and the Macro-average of the average precision scores was computed for each label within each bin.  For each bin, significance was computed via a paired t-test.\footnote{To be precise:  The performance score for SVMs and Dependency LDA on each label with a training frequency in the appropriate range, for each split of the dataset, was treated as single a pair of values for the t-test.}

On both datasets, it is clear that Dependency-LDA has a significant advantage over Tuned-SVMs on the rarest labels.  On the EUR-Lex dataset, Dependency-LDA significantly outperforms SVMs on labels with training frequencies of less than five, and performs better than SVMs (though not significantly at the $\alpha=.05$ level) on the three lower quintiles of label frequencies.  SVM performance catches up to Dependency-LDA on labels somewhere in the upper-middle range of label-frequencies, and surpasses Dependency-LDA (significantly) for the labels in the most frequent quintile.  On the NYT dataset, Dependency-LDA outperforms SVMs across all label frequencies (this difference is significant on all quintiles except the one containing labels with a frequency of one).

\begin{figure}[ht] % float placement: (h)ere, page (t)op, page (b)ottom, other (p)age
  \centering
  \includegraphics[width=1\linewidth]{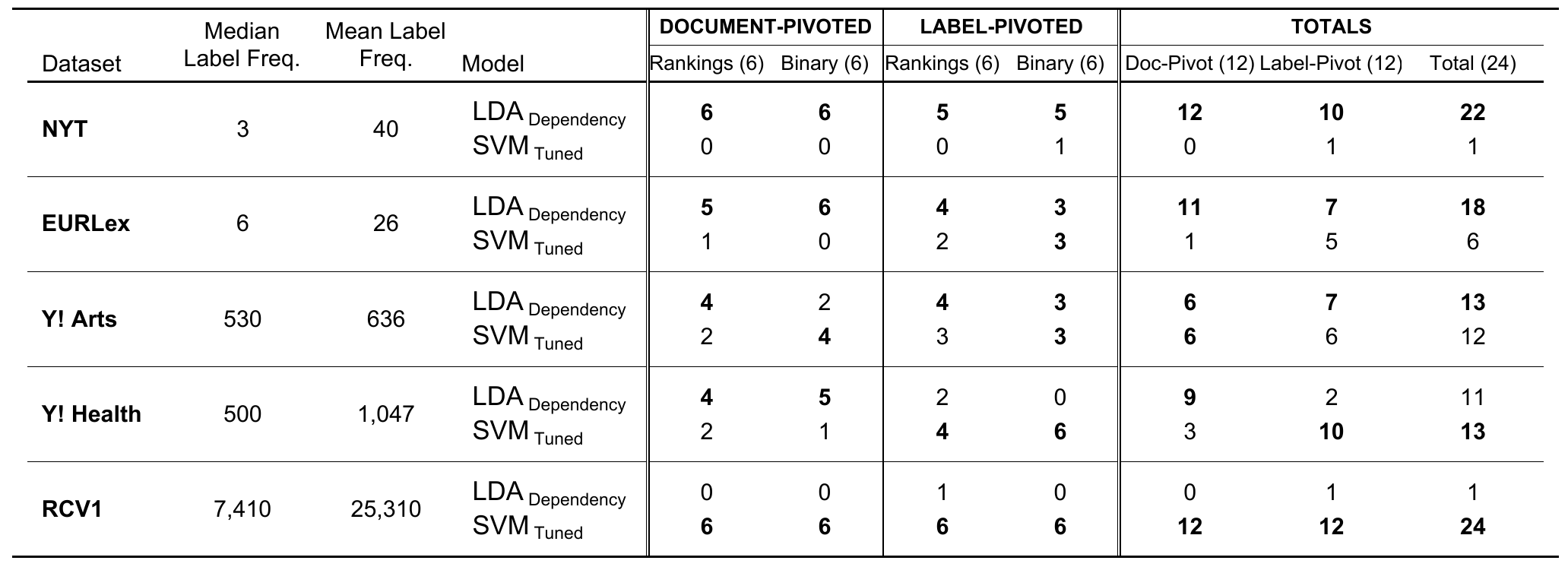}
  \caption{Summary comparison of the performance of Dependency-LDA vs. Tuned SVMs across the five datasets.  For each type of prediction (Document/Label Pivoted), we show the number of evaluation metrics on which each model achieved the best overall score. Performance is first broken down by the type of evaluation metric used (Rank-Based vs. Binary).  Totals are shown in the three right columns.  Note that {\it six} is the maximum achievable value here for both binary and rank-based predictions; although seven rank-based scores were presented in previous tables, the $AUC_{ROC}$ and {\sc Rank-Loss} metrics have been combined here.}
  \label{fig:DepLDA_Vs_TunedSVM_Overall}
\end{figure}

\subsection{Summary: Dependency-LDA vs. Tuned SVMs}

There are several key points which are evident from the experimental results presented above.  First, the Dependency-LDA model significantly outperforms the simpler Prior-LDA and Flat-LDA models, and that the scale of this improvement depends on the statistics of the datasets.  Secondly, under certain conditions, the LDA-based models (and most notably, Dependency-LDA) have a significant advantage over the binary SVM methods, but under other conditions the SVMs have a significant advantage.  We have already discussed some of the specific factors that play a role in these differences.  However, it is useful to take a step back, and consider the key model comparisons across all four of the prediction tasks.  Namely, we wish to more generally explore the conditions in which probabilistic generative models such as topic models may have benefits compared to discriminative approaches such as SVMs, and vice-versa.  To this purpose, we now focus on the {\it overall} performance of our best LDA-based approach (Dependency-LDA) and our best discriminative approach (Tuned-SVMs), rather than focusing on performance with respect to specific evaluation metrics.

In Figure~\ref{fig:DepLDA_Vs_TunedSVM_Overall}, we present a summary of the performance for Dependency-LDA and Tuned-SVMs across all four prediction tasks and all five datasets.  For each dataset and prediction task, we present the total number of evaluation metrics for which each model achieved the {\it best} score out of all five of our models (in the case of ties, both models are awarded credit).\footnote{Note that although we presented seven rank-based evaluation metrics in the previous tables, the maximum score for each element of the table is {\it six}, because we collapse the performance for the $AUC_{ROC}$ and {\sc Rank-Loss} metrics, due to their equivalence.}  The results have been ordered from top-to-bottom by the relative amount of training data there is in each dataset.  Note that these datasets fall into three qualitative categories: (1) the power-law datasets (NYT and EUR-Lex), (2) the Yahoo! datasets (which are not highly multi-label, and do not have large amounts of training data per label), and (3) the RCV1-V2 dataset, which has a large amount of training data for each label, and is more highly multi-label than the Yahoo! datasets but less than the power-law datasets (and, additionally, unlike the other datasets, had many algorithmically-assigned labels).

Looking at the full totals in the rightmost column of Figure~\ref{fig:DepLDA_Vs_TunedSVM_Overall}, one can see that for the power-law datasets, Dependency-LDA has a significant overall advantage over SVMs. For the two Yahoo! datasets, the overall performance of the two models is quite comparable.  Finally, for the RCV1-V2 dataset, Tuned SVMs clearly outperform Dependency-LDA.  This general interaction between the amount of training data and the relative performance of these two models has been discussed earlier in the paper, but is perhaps most clearly illustrated in this simple figure.

A second feature that is evident in Figure~\ref{fig:DepLDA_Vs_TunedSVM_Overall} is that, all else being equal, the Dependency-LDA model seems better suited for Document-Pivoted predictions and SVMs seem better suited for Label-Based predictions.  For example, although Dependency-LDA greatly outperforms SVMs overall on EUR-Lex, the performance for Label-Pivoted predictions on this dataset are in fact quite close.  And although overall performance is quite similar for the Yahoo! {\it Health} dataset, Dependency-LDA dominates SVMs for Document-pivoted predictions, and the reverse is true for Label-pivoted predictions.
A likely explanation for this difference is the fundamentally different way that each model handles multi-label data.
In Dependency-LDA (and all of the LDA-based models), although we learn a model for each label during training, at test time it is the {\it documents} that are being modeled.  Thus the ``natural direction'' for LDA-based models to make predictions is {\it within} each document, and {\it across} the labels.  The SVM approach, in contrast, builds a binary classifier for each label, and thus the ``natural direction'' for Binary SVMs to make predictions is {\it within} each label, and {\it across} documents.  Thus, if one is to consider which type of classifier would be preferable for a given application, it seems important to consider whether label-pivoted or document-pivoted predictions are more suited to the task, in addition to what the statistics of the corpus look like.

\section{Conclusions}

In conclusion, in terms of the three LDA-based models considered in this paper, our experiments indicate that (1) Prior-LDA improves performance over the Flat-LDA model by accounting for baseline label-frequencies, (2) Dependency-LDA significantly improves performance relative to both Flat-LDA and Prior-LDA by accounting for label dependencies, and (3) The relative performance improvement that is gained by accounting for label dependencies is much larger in datasets with large numbers of labels per document.

In addition, the results of comparing LDA-based models with SVM models indicate that on large-scale datasets with power-law like statistics, the Dependency-LDA model generally outperforms binary SVMs.  This effect is more pronounced for document-pivoted predictions, but is also generally the case for label-pivoted predictions.  The results of label-pivoted predictions across different label-frequencies indicate that the performance benefit observed for Dependency-LDA is in part due to improved performance on rare labels.

Our results with SVMs are consistent with those obtained elsewhere in the literature; namely, binary SVM performance degrades rapidly as the amount of training data decreases, resulting in relatively poor performance on large scale datasets with many labels.  Our results for the LDA-based methods, most notably for the Dependency-LDA model, indicate that probabilistic models are generally more robust under these conditions.   In particular, the comparison of Dependency-LDA and SVMs on labels at different training frequencies demonstrates that Dependency-LDA clearly outperformed SVMs on the rare labels on our large scale datasets.  Additionally, Dependency-LDA was competitive with, or better than, SVMs on labels across all training frequencies on these datasets (except on the most frequent quintile of labels in the EUR-Lex dataset).  Furthermore, Dependency-LDA clearly outperformed SVMs on the document-pivoted predictions on both large scale datasets.

Robustness in the face of large numbers of labels and small numbers of training documents has not been extensively commented on in the literature  on multi-label text classification, since the majority of studies have focused on corpora with relatively few labels, and many examples of each label. Given that human labeling is an expensive activity, and that many annotation applications consist of a large number of labels with a long tail of relatively rare labels,  prediction with large numbers of labels is likely to be an increasingly important problem in multi-label text classification and one that deserves further attention.

A potentially useful direction for future research is to combine discriminative learning with the types of generative models proposed here, possibly using extensions of existing discriminative adaptations of the LDA model \citep[e.g.,][]{Blei_Mcauliffe_07,Julien_Etc_2008,Mimno_McCallum_2008}.  A hybrid approach could combine the benefits of  generative LDA models---such as explaining away, natural calibration for sparse data, semi-supervised learning \citep[e.g., ][]{Druck_Etc_2007}, and interpretability \citep[e.g., ][]{Ramage_Etc_2009}---with the advantages of discriminative models such as task-specific optimization and good performance under conditions with many training examples.  The approach we propose can also be applied to domains outside of text classification; for example, it can be applied to multi-label images in computer vision \citep{CaoFeiFei_2007}.

\subsection*{Acknowledgements}

The authors would like to thank the anonymous reviewers of this paper, as well as the guest editors of this issue of the Machine Learning Journal, for their helpful comments and suggestions for improving this paper.
This material is based in part upon work supported by the National Science Foundation under grant number IIS-0083489, by the Intelligence Advanced Research Projects Activity (IARPA) via Air Force Research Laboratory contract number FA8650-10-C-7060, by the Office of Naval Research under MURI grant N00014-08-1-1015, by a Microsoft Scholarship (AC), by a Google Faculty Research award (PS).  The U.S. Government is authorized to reproduce and distribute reprints for Government purposes notwithstanding any copyright annotation thereon.  Disclaimer: the views and conclusions contained herein are those of the authors and should not be interpreted as necessarily representing the official policies or endorsements, either expressed or implied, of NSF, IARPA, AFRL, ONR, or the U.S. Government

\appendix

\vfill\eject

\section{Details of Experimental datasets} \label{sec:Appendix_DatasetDetails}

\subsection{The New York Times (NYT) Annotated Corpus}

The New York Times Annotated Corpus  (available from the Linguistic Data Consortium) contains nearly every article published by The New York Times
over the span of 10 years from January 1st, 1987 to June 19th, 2007 \citep{NYTCorpus}.  Over 1.5 million of these articles have ``descriptor'' tags that were manually assigned by human labelers via the New York Times Indexing Service, and correspond to the subjects mentioned within each article\footnote{Note that additional types of meta-data are available for many of these documents.  This includes additional labeling schemes, such as the ``general online descriptors'' ---which are algorithmically assigned---and that for the purposes of this paper we specifically used the hand-assigned ``descriptor'' tags.  We refer to these ``descriptor'' tags as ``labels'' for consistency throughout the paper.}
(see Tables~\ref{table:labelworddistributions} and~\ref{table:topicsofconcepts} for numerous examples of these descriptors).

To construct an experimental corpus of NYT documents, we selected all documents that had both text in their body and at least three descriptor labels from the ``News$\backslash$U.S'' taxonomic directory. After removing common stopwords, we randomly selected 40\% of the articles for training and reserved the remaining articles for testing. Any test article containing label(s) that did not occur in the training set was then re-assigned to the training set so that all labels had at least one positive training instance. This procedure resulted in a training corpus containing 14,669 documents and 4,185 unique labels, and a test corpus with 15,989 documents and 2,400 unique labels.

For feature selection in all models, we removed words that appeared fewer than 20 times within the training data, which left us with a vocabulary of 24,670 unique words. For this dataset, evaluation on test documents was restricted to the subset of 2,400 labels that occurred at least once in both the training and test sets (this approach for handling missing labels is consistent with common practice in the literature).

\subsection{The EUR-Lex Text Dataset}
The EUR-Lex text collection \citep{MenciaFurnkranz_2008_Efficient} contains documents related to European Union law (e.g. treaties, legislation),  downloaded from the publicly available EUR-Lex online repository \citep{EURLex_Repository}. The dataset we downloaded contained 19,940 documents documents and 3,993 EUROVOC descriptors.  Note that there are additional types of meta-data available in the dataset, but we restricted our analysis to the EUROVOC descriptors, which we will refer to as labels.

The dataset provides $10$ cross-validation splits of the documents into training and testing sets equivalent to those used in \cite{MenciaFurnkranz_2008_Efficient}. We downloaded the stemmed and tokenized forms of the documents and performed our own additional pre-processing of the data splits.  For each split, we first removed all empty documents (documents with no words)\footnote{We note that after removing empty documents, we were left with 19,348 documents. The dataset statistics in terms of the EUROVOC descriptors (shown in Table~\ref{table:datastats})  however are based on the 19,800 documents for which there was at least one descriptor assigned to the document.}, and then removed all words appearing fewer than 20 times within the training set.  This was done independently for each split so that no information about test documents was used during training.

%\subsection{RCV1-v2 and Yahoo Subdirectory datasets}
\subsection{The RCV1-v2 Dataset}
The RCV1-v2 dataset \citep{Lewis_Etc_2004}---an updated version of the Reuters RCV1 corpus---is one of the more commonly used benchmark datasets used in multi-label document classification research \citep[e.g., see][]{FanLin_2007,Furnkranz_Etc_2008,Tsoumakas_Etc_2009}. The dataset consists of over 800,000 newswire stories that have been assigned one or more of the 103 available labels (categories).  We used the original training set from the {\it LYRL2004} split given by \cite{Lewis_Etc_2004}.  Only 101 of the 103 labels are present in the 23,149 document training-set, and we employ the commonly-used convention of restricting our evaluation to these 101 labels.  We randomly selected 75,000 of the documents from the {\it LYRL2004} test split for our test set\footnote{Early experiments that we performed found that results on this subset were nearly identical to those for the full {\it LYRL2004} test set.  The only score that is significantly different is the {\sc Margin} for the label-pivoted results (because this metric is closely tied to the total number of documents in the test set).}.

One problematic feature of the RCV1-v2 dataset is that many of the labels were not manually assigned by editors but were instead automatically assigned via automated expansion of the topic hierarchy \citep{Lewis_Etc_2004}.  Although it is possible to avoid evaluating predictions on these automatically-assigned labels---by only considering the subset of labels which are leaves within the topic hierarchy \citep{Lewis_Etc_2004}---these automatically-assigned labels still play a major role during training.  Furthermore, this type of automated hierarchy expansion (although sensible) leads to some unnatural and perhaps misleading statistical features in the dataset.  For example, although the average number of labels per document in the dataset is relatively large (which indicates that this is a highly multi-label dataset), the number of unique sets of labels is actually quite small relative to the number of documents, likely due to the fact that many of the documents were originally single-label and then automatically expanded such that they seemed multi-label.  Although there is nothing inherently wrong with this approach, it (1) may lead to misleadingly positive results for models that are able to pick up on the automatically assigned labels, rather than the human-assigned labels, (2) leads to statistics which significantly deviate from the types of power-law distributions observed in many real-world situations, and (3) can lead one to assume that the dataset contains a much more complex space of label-combinations than is actually contained in the dataset. Note that, as illustrated in Table~\ref{table:datastats}, the RCV1-V2 dataset is in most respects much more similar to the small Yahoo! subdirectory datasets than to the real-world power-law datasets.

\subsection{The Yahoo! Subdirectory Datasets}
The Yahoo! datasets that we use consist of the \textit{Arts} and the \textit{Health} subdirectories from the collection used by \cite{UedaSaito_2002}.  We use the same training and test splits as presented in recent work by \cite{JiTangYuYe_2008} (where each training split consists of $1000$ documents, and all remaining documents are used for testing). These datasets contain 19 and 14 unique labels respectively. The number of labels per document in each dataset is quite small;
about 55-60\% of training documents are assigned a single label, and about 85-90\% are assigned either one or two labels.  This was in large part due to the methods used to collect and pre-process the data, wherein only the second-level categories of the Yahoo! directory structure which had at least 100 examples were kept \citep{JiTangYuYe_2008}.  We evaluated models across all of five of the available train/test splits for both the \textit{Arts} and the \textit{Health} sub-directories.

\vfill\eject

\section{Hyperparameter and Sampling Parameter Settings for Topic Model Inference} \label{sec:All_HyperParameter_Settings}

\begin{table}[t] % float placement: (h)ere, page (t)op, page (b)ottom, other (p)age
  \centering
 \includegraphics[width=.85\textwidth, keepaspectratio]{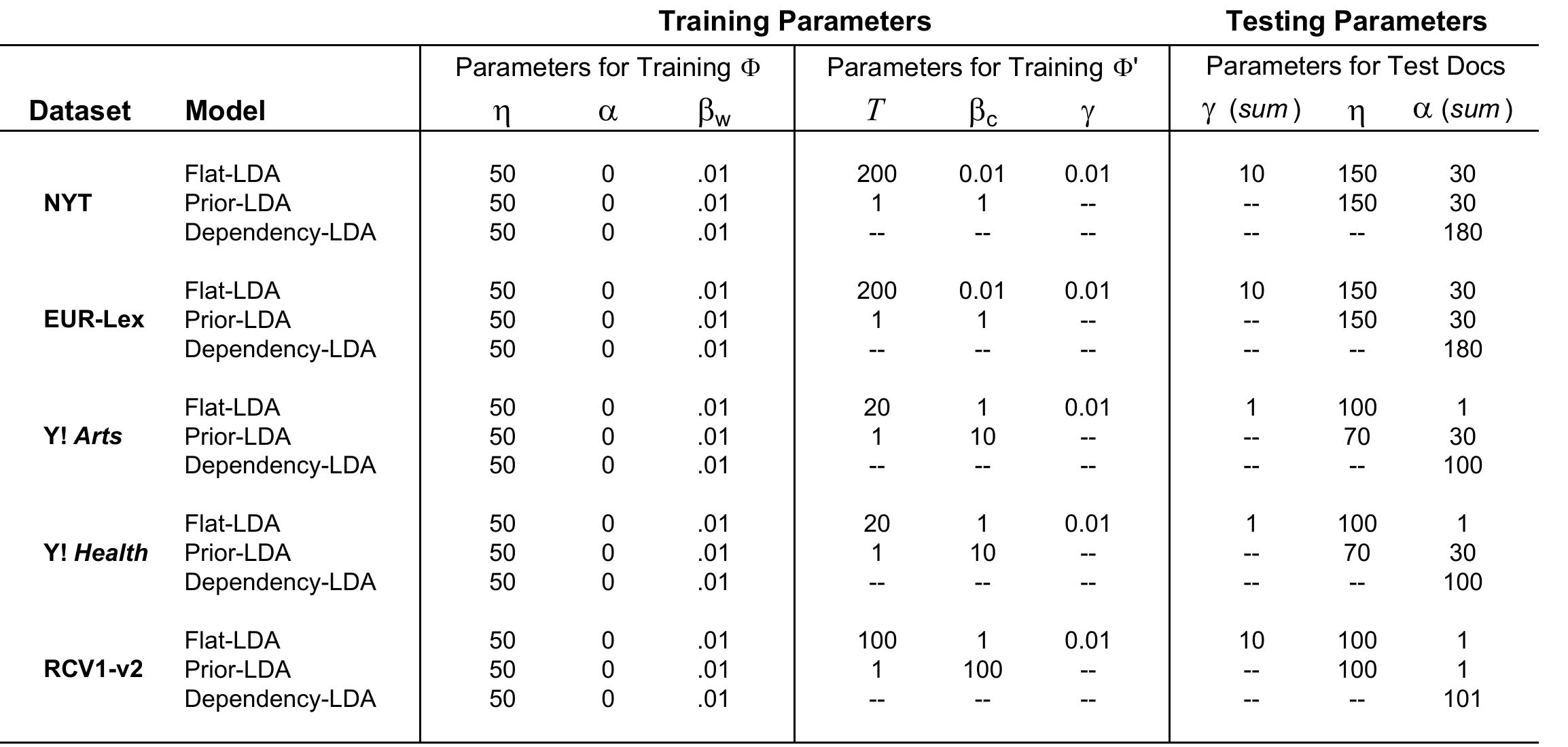}
   \vspace{-6pt}
   \caption{
   {\small
   Hyperparameter values used for training and testing Dependency-LDA, Prior-LDA, and Flat-LDA, on all datasets.  Note that the test-document parameters values for $\gamma$ and $\alpha$ are given in terms of their {\it sums}; the actual pseudocount added to each element of $\theta'_d$ is $\gamma / T$ and the flat pseudocount added to each element of $\theta_d$ is $\alpha / C$.}
   }
  \label{table:Hyperparameter_Settings}
   \vspace{-4pt}
\end{table}

In this section, we present the complete set of parameter settings used for training and testing all LDA-based models, and motivate our particular choices for these settings.
Note that all parameter settings were chosen heuristically were not optimized with respect to any of the evaluation metrics.
It would be reasonable to expect some improvement in performance over the results presented in this paper by optimizing the hyperparameters via cross-validation on the training sets (as we did with Binary SVMs).

\subsection{Hyperparameter Settings}

Table~\ref{table:Hyperparameter_Settings} shows the hyperparameter values that were used for training and testing the three LDA-based models on the five experimental datasets.  Note that not all parameters are applicable for all models; for example, since Flat-LDA does not incorporate any  $\phi'$ distributions of topics over labels, parameters such as $\beta_{\mathrm{C}}$  and $\gamma$ do not exist in this model.

For all models, we used the same set of parameters to train the $\phi$ distributions of labels over words; $\eta=50$, and $\beta_{\mathrm{W}}=.01$.  Early experimentation indicated that the exact values of $\eta$ and $\beta_{\mathrm{W}}$ were generally unimportant as long as $\eta \gg 1$ and $\beta_{\mathrm{W}} \ll 1$.  The total strength of the Dirichlet prior on $\theta$, which is dictated by $\eta$, is significantly larger than what is typically used in topic modeling.  This makes sense in terms of the model; unlike in unsupervised LDA, we know a-priori which labels are present in the training documents, and setting a large value for $\eta$ reflects this knowledge.

Parameters used to train the $\phi'_t$ distributions of topics over labels were chosen heuristically as follows.  In Dependency-LDA, we first chose the number of topics ($T$).  For the smaller datasets, we set the number of topics ($T$) approximately equal to the number of unique labels ($C$).  For the two datasets with power-law like statistics, NYT and EUR-Lex, we set $T=200$, which is significantly smaller than the number of unique labels.  In addition to controlling for model complexity, some early experimentation indicated that setting $T \ll C$ improved the interpretability of the learned topic-label distributions in these datasets
\footnote{Specifically, early in experimentation for NYT and EUR-Lex we trained a set of topics with $T=50, 100, 200, 400$, and $1000$. Visual inspection of the resulting topic-label distributions indicated that setting $T$ to be too small (e.g., $T \leq 100$) over-penalized infrequent labels; labels that had fewer than approximately $25$ training documents rarely had high probabilities in the model, even when the labels were clearly relevant to a topic.  Setting $T$ to be too large (e.g., $T=1000$) led to both redundancy among the topics and to topics which appeared  to be over-specialized (i.e., some of the topics had only a few documents with labels assigned to them).}.

Given the value of $T$ for each dataset, we set $\beta_{\mathrm{C}}$ such that the {\it total} number of pseudocounts that were added to all topics was approximately equal to one-tenth of the total number of counts contributed by the observed labels.  For example, each split of EUR-Lex contains approximately $90,000$ label tokens in total.  Given our choice of $T=200$ topics, and the approximately $4,000$ unique label types, by setting $\beta_{\mathrm{C}}=.01$, the total number of pseudocounts that are added to all topics is $200 \times 4000 \times .01 = 8000$, (which is approximately one-tenth the total number of observed labels).  For Prior-LDA, since there is only one topic ($T=1),$ we increased the value of $\beta_{\mathrm{C}}$ in order to be consistent with this general principle.

For setting the parameters for test documents, we kept the {\it} total number of pseudocounts that were added to the test documents consistent across all models.  To help illustrate this, the hyperparameter settings for test document parameters $\alpha$ and $\gamma$ are shown in terms of their {\it sums} in Table~\ref{table:Hyperparameter_Settings}, rather than in terms of their element-wise values.
For the two power-law datasets, the total weight of the prior on $\theta$ was equal to 180, and for the three benchmark datasets the total weight of the prior on $\theta$ was equal to 100.  We used smaller priors for the benchmark datasets because these documents were shorter on average, and we wished to keep the pseudocount totals roughly proportional to document lengths.

\subsection{Details about Sampling and Predictions}

Here we provide details regarding the number of chains and samples taken at each stage of inference (e.g., the total number of samples that were taken for each test document).  These settings were equivalent for all three of the LDA-based models and for all datasets.

To train the $C$ label-word distributions $\phi_c$, we ran $48$ independent MCMC chains (each initialized using a different random seed)\footnote{The exact number of chains is unimportant. However, it is well known that averaging multiple samples from an MCMC chain systematically improves parameter estimates.  The particular number of chains that we ran ($48$) is circumstantial; we had $8$ processors available and ran $6$ chains on each.}.  After a burn-in of 100 iterations we took a single sample at the end of each chain, where a sample consists of all $z_i$ assignments for the training documents.  These samples were then averaged to compute a single estimate for all $\phi_c$ distributions (as mentioned elsewhere in the paper, the same estimates of $\phi_c$ were used across all three LDA-based models).

To train the $T$ topic-label distributions $\phi'_t$ for Dependency-LDA, we ran $10$ MCMC chains, taking a single sample from each after a burn-in of $500$ iterations.  One can not average the estimates of  $\phi'_t$ over multiple chains as we did when estimating $\phi$, because the topics
are being learned in an unsupervised manner and do not have a fixed interpretation between chains.  Thus, each chain provides a unique set of $T$ topic distributions.  These 10 estimates are then stored for test time (at which point we can eventually average over them).

At test time, we took $900$ total samples of the estimated parameters for each test document ($\theta_d$ for all models, plus $\theta'_d$ for Dependency-LDA)\footnote{As noted previously, we used the ``fast inference'' method, in which we do not actually sample the $c$ parameters}.  For each model, we ran $60$ independent MCMC chains, and took $15$ samples from each chain using an initial burn-in of $50$ iterations and a $5$ iteration lag between samples (to reduce autocorrelation).  For Dependency-LDA, in order to incorporate the ten $10$ separate estimates of $\phi'_t$, we distributed the $60$ MCMC chains across the different sets of topics;  specifically, $6$ chains were run using each of the $10$ sets of topics (giving us $60$ in total).

In order to average estimates across the chains, we used our $900$ samples to compute the posterior estimates of $\theta_d$ and $\alpha'^{(d)}$ (where $\alpha'^{(d)}$ only changes across samples for Dependency-LDA; for Prior-LDA, this estimate is fixed, and it is not applicable to Flat-LDA).  The final (averaged) estimate of the prior $\alpha'^{(d)}$ is added to the final estimate of $\theta_d$ to generate a single posterior predictive distribution for $\theta_d$ (due to the conjugacy of the Dirichlet and multinomial distributions). We note that at this step we used one last heuristic; when combining the estimates of the $\alpha^{(d)}$ and $\theta_d$ for each document, we set the total weight of the Dirichlet prior $\alpha'^{(d)}$ equal to the total number of words in the document (i.e., we set $\sum_{c}{\alpha'^{(d)}} = \sum_{c}{\theta_d}$).  We chose to do this because, whereas the total weight of $\alpha'^{(d)}$ used during sampling was fixed across all documents, the documents themselves had different numbers of words.  Therefore, for very long documents, the final predictions would otherwise be mostly influenced by the word-assignments, and for very short documents the prior would overwhelm word-assignments
\footnote{Early experimentation with a smaller version of the NYT dataset indicated that this method leads to modest improvements in performance.}.
The final posterior estimate of $\theta_d$ computed from the $900$ samples was used to generate all predictions.

\vfill\eject

\vfill\eject

\section{Derivation of Sampling Equation for Label-Token Variables ($C$)} \label{sec:Derivation_CSampler}

\noindent
In this appendix, we provide a derivation of Equation~\eqref{c_testlabelc}, for sampling a document's label-tokens $\bf{c}^{(d)}$.

The variable {\footnotesize$c^{(d)}_{i}$} can take on values {\footnotesize$\{1, 2, \hdots, C\}$}. We need to compute the probability of {\footnotesize$c^{(d)}_{i} = c$} (for $c \leq C$) conditioned on the label assignments {\footnotesize$z^{(d)}$}, the topic assignments {\footnotesize${z'}^{(d)}$}, and the remaining variables {\footnotesize$c^{(d)}_{-i}$}.
{\footnotesize
\begin{equation}\label{eq:samplec}
\begin{split}
p( c^{(d)}_{i} = c \; | \; z^{(d)}, {z'}^{(d)}, c^{(d)}_{-i})
                                        &= \frac{p(z^{(d)}, c^{(d)} \; | \;  {z'}^{(d)})}{p(z^{(d)} \; | \;  {z'}^{(d)})}\\
									    &\propto p(z^{(d)}, c^{(d)} \; | \;  {z'}^{(d)})\\
									    &= p(z^{(d)} | c^{(d)}) \cdot p( c^{(d)} | {z'}^{(d)}_{i})\\
									   &\propto p(z^{(d)} | c^{(d)}) \cdot p( c^{(d)}_i | {z'}^{(d)}_{i}, c^{(d)}_{-i})
\end{split}
\end{equation}}
%\vspace{-1mm}

\noindent Thus, the conditional probability of {\footnotesize$c^{(d)}_{i} = c$} is a product of two factors. The first factor in Equation~\eqref{eq:samplec} is the likelihood of the label assignments {\footnotesize$z^{(d)}$} given the labels {\footnotesize$c^{(d)}$}. It can be computed by marginalizing over the document's distribution over labels {\footnotesize$\theta^{(d)}$ \ }:
{\footnotesize
\begin{equation}
\begin{split}
p(z^{(d)} | c^{(d)}) &= \int_{\theta^{(d)}}  p(z^{(d)} | \theta^{(d)}) \cdot p( \theta^{(d)} | c^{(d)} )  \; d\theta^{(d)}\\
	     &= \int\limits_{\theta^{(d)}} \Biggl( \prod_{i=1}^{N} \theta^{(d)}_{z^{(d)}_{i}} \Biggr) \Biggl( \frac{1}{\mathcal{B}( {{\alpha'}^{(d)}})} \prod_{j=1}^{C} \left(\theta^{(d)}_{j}\right) ^ {{\alpha'}^{(d)}_{j}- 1} \Biggr)   \; d\theta^{(d)}\\
	     &= \frac{1}{\mathcal{B}({{\alpha'}^{(d)}})} \int\limits_{\theta^{(d)}} {\left(\theta^{(d)}_{\cdot}\right)}^{{N^{CD}_{\cdot,d}}} \prod_{j=1}^{C} \left(\theta^{(d)}_{j}\right) ^{ {\alpha'}^{(d)}_{j}- 1} \; d\theta^{(d)}\\
	     &= \frac{1}{\mathcal{B}({{\alpha'}^{(d)}})} \int\limits_{\theta^{(d)}}  \prod_{j=1}^{C} \left(\theta^{(d)}_{j}\right)^{ {\alpha'}^{(d)}_{j} + N^{CD}_{j,d} - 1} \; d\theta^{(d)}\\
         &= \frac{ \mathcal{B}( {\alpha'}^{(d)}_{j}+ N^{CD}_{\cdot,d})}{ \mathcal{B}({{\alpha'}^{(d)}})}
\end{split}
\end{equation}}
%\vspace{-1mm}

\noindent Here {\footnotesize$N^{CD}_{j,d}$} represents the number of words in document $d$ assigned the label {\footnotesize$j \in \{1, 2,\hdots,C\}$} and {\footnotesize$\mathcal{B}(\alpha)$} represents the multinomial Beta function whose argument is a real vector $\alpha$.
The numerator on the last line is an abuse of notation that denotes the Beta function whose argument is the vector sum {\footnotesize$\left([{\alpha'}^{(d)}_{1}\hdots {\alpha'}^{(d)}_{C}]~+~[N^{CD}_{1,d}\hdots N^{CD}_{C,d}]\right)$}.
The Beta function can be expressed in terms of the Gamma function:
{\footnotesize
\begin{equation}
\begin{split}
p(z^{(d)} | c^{(d)})  &= \frac{ \mathcal{B}( {{\alpha'}^{(d)}}+ N^{CD}_{\cdot,d})}{ \mathcal{B}({{\alpha'}^{(d)}})}\\
			     &= \frac{ \prod_{j=1}^C\Gamma({\alpha'}^{(d)}_{j} + N^{CD}_{j,d})} { \prod_{j=1}^C\Gamma( {\alpha'}^{(d)}_{j} ) } \; * \; \frac{ \Gamma( \sum_{j=1}^C{\alpha'}^{(d)}_{j})}{ \Gamma( \sum_{j=1}^C {\alpha'}^{(d)}_{j} + N^{CD}_{j,d})}\\
			     &\propto \frac{ \prod_{j=1}^C\Gamma({\alpha'}^{(d)}_{j} + N^{CD}_{j,d})} { \prod_{j=1}^C\Gamma( {\alpha'}^{(d)}_{j} ) }
\end{split}
\end{equation}
}

\noindent Here the Gamma function takes as argument a real-valued number. As the value of {\footnotesize$c_{i}^{(d)}$} iterates over the range {\footnotesize$\{1,2,\hdots, C\}$}, the prior vector {\footnotesize${\alpha'}^{(d)}$} changes but the summation of its entries {\footnotesize$\sum_{j=1}^{C}{\alpha'}^{(d)}_{j}$} and the data counts {\footnotesize$N^{CD}_{j,d}$} do not change.

The second term in Equation~\eqref{eq:samplec}, {\footnotesize \  $p( c^{(d)}_i | {z'}^{(d)}_{i}, c^{(d)}_{-i})$},
is the probability of the label {\footnotesize$c_{i}^{(d)}$} given its topic assignment {\footnotesize${z'}^{(d)}_{i}$} and the remaining labels {\footnotesize$c^{(d)}_{-i}$}. This is analogous to the probability of a word given a topic in standard unsupervised LDA (where the {\footnotesize$c_{i}^{(d)}$} variable is analogous to a ``word", and the {\footnotesize${z'}^{(d)}_{i}$} variable is analogous to the ``topic-assignment" for the word). This probability---denoted as {\footnotesize${\phi'}^{(t)}_{c}$}---is estimated
during training time.  Thus, the final form of
Equation~\eqref{eq:samplec} is given by:
{\footnotesize
\begin{equation}
\begin{split}
p( c^{(d)}_{i} = c \; | \; z^{(d)}, {z'}^{(d)}, c^{(d)}_{-i}) \propto \frac{ \prod_{j=1}^C\Gamma({\alpha'}^{(d)}_{j}+ N^{CD}_{j,d})} { \prod_{j=1}^C\Gamma( {\alpha'}^{(d)}_{j})} \; \cdot \; {\phi'}^{(t)}_{c}
\end{split}
\end{equation}
}
\vfill\eject

\section{Comparisons With Published Results} \label{sec:Comparisons_PublishedResults}

The one-vs-all SVM approach we employed for comparison with our LDA-based methods is a highly popular benchmark in the multi-label classification literature.  However, there are a large number of alternative methods (both probabilistic and discriminative) which have been proposed, and this is an active area of research.  In order to put our results in the larger context of the current state of multi-label classification, we compare below our results with published results for alternative classification methods.  Because of the variability of published results---due to the
lack of consensus in the literature in terms of the prediction-tasks,
evaluation metrics, and versions of datasets that have been used for model evaluation---there are relatively few results that we can compare to.  Nonetheless, for all but one of our datasets (the NYT dataset, which we constructed ourselves), we were able to find published values for at least some of the evaluation metrics we utilized in this paper.

In this Appendix we present a comparison of our own scores (for the two SVM and three LDA-based approaches) with published scores on equivalent training-test splits of equivalent datasets.  The goals of this Appendix are (1) to put our own results in the context of the larger state of the area of multi-label classification, (2) to demonstrate that our Tuned-SVM approach is competitive with other similar Tuned-SVM benchmarks that have been used elsewhere, and (3) to demonstrate that on power-law datasets, our Dependency-LDA model achieves scores that are competitive with state-of-the art discriminative approaches.

\subsection*{Comparison With Published scores on the EUR-Lex Dataset}

\begin{figure}[ht] % float placement: (h)ere, page (t)op, page (b)ottom, other (p)age
  \centering
  \includegraphics[width=.8\linewidth]{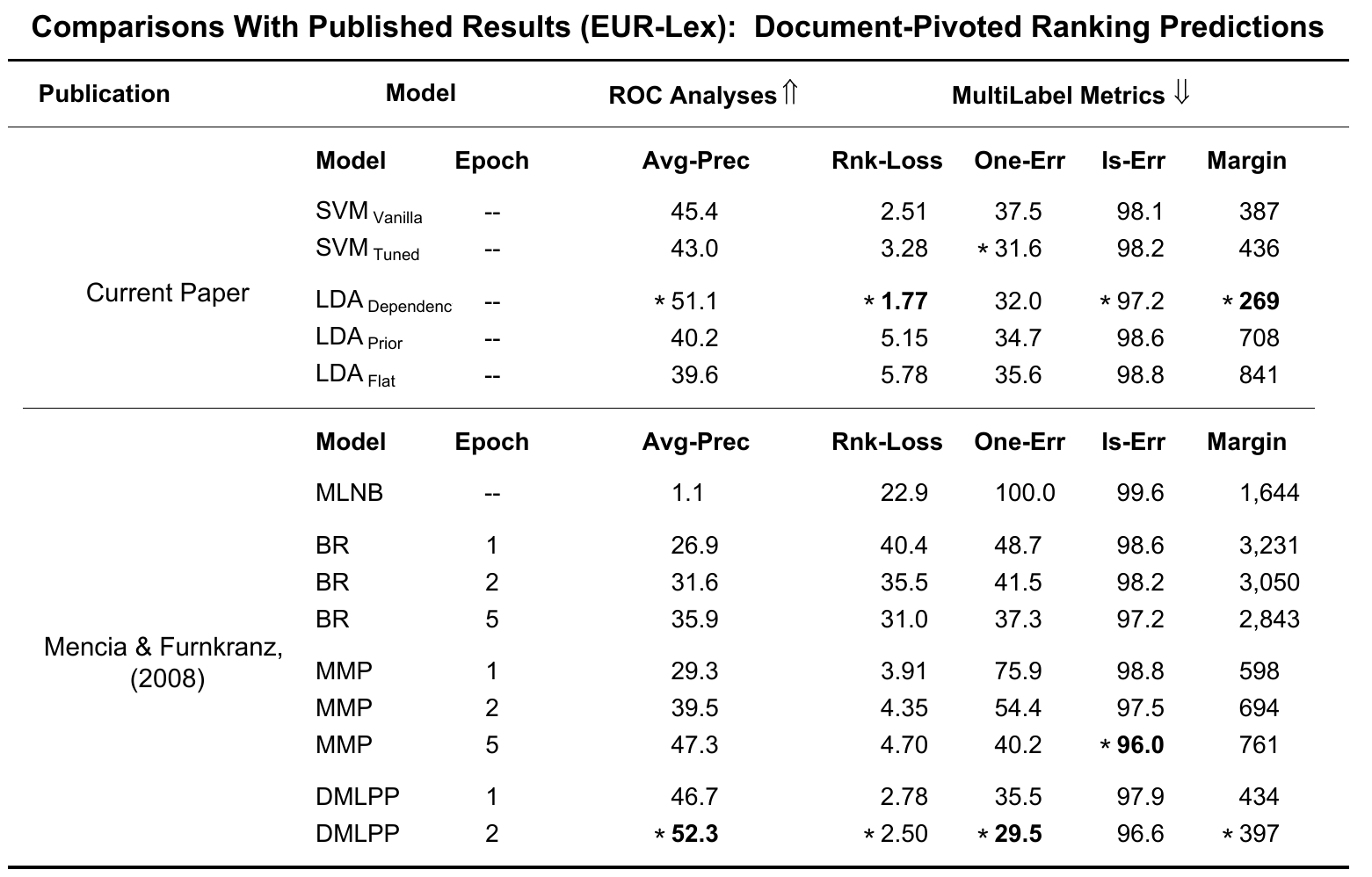}
  \caption{Comparison of results from the current paper with result from ~\citet{MenciaFurnkranz_2008}, on document-pivoted ranking evaluations.}
  \label{fig:ComparisonWithPublished_Ranking}
\end{figure}

To the best of our knowledge, only one research group has published results using the EUR-Lex dataset ~\citep[][]{MenciaFurnkranz_2008,MenciaFurnkranz_2008_Efficient}.
Figure~\ref{fig:ComparisonWithPublished_Ranking} compares our results with all results presented in ~\citet{MenciaFurnkranz_2008}\footnote{Note that we did not use an equivalent feature selection method as in their paper; due to memory constraints of their algorithms, ~\citet{MenciaFurnkranz_2008} reduced the number of features to $5,000$ for each split of the dataset, whereas our feature selection method (where we removed words occurring fewer than $20$ times in the training set) left us with approximately 20,000 features for each split.} for the EUR-Lex Eurovoc descriptors.  The best two algorithms from  ~\citet{MenciaFurnkranz_2008}---MMP ({\it Multilabel Multiclass Perceptron}) and DMLPP ({\it Dual Multilabel Pairwise Perceptrons})---are discriminative, perceptron-based algorithms.  Both algorithms account for label-dependencies, and are designed specifically for the task of document-pivoted label-ranking (thus, no results are presented for label-pivoted predictions).  Training of these algorithms was performed to optimize rankings with respect to the Is-Error loss function.

Dependency-LDA outperforms all algorithms (on all five measures) from ~\citet{MenciaFurnkranz_2008} except MMP (at 5 epochs) and DMLPP (at 2 epochs)\footnote{In the perceptron-based algorithms from ~\citet{MenciaFurnkranz_2008}, the number of Epochs corresponds to the number passes over the training corpus during which the model weights are tuned.  See reference for further details.}.  Dependency-LDA outperforms MMP(5) on all metrics but Is-Error (which was the metric the algorithm was tuned to optimize).   Dependency-LDA beats DMLPP at 1 epoch on all metrics, but at 2 epochs (which gave their best overall set of results) performance between the two algorithms is quite close overall; Dependency-LDA outperforms DMLPP(2) on 2/5 measures, and performs worse on 3/5 measures (although, the relative improvement of DMLPP over Dependency-LDA on Average-Precision is fairly small relative to differences on other scores).
In terms of overall performance, it is not clear that either Dependency-LDA or DMLPP(2) is a clear winner.  However, it seems fairly clear that the Dependency-LDA outperforms MMP overall, and at the very least is reasonably competitive with DMLPP(2).  This is particularly surprising given that both the MMP and DMLPP algorithms are designed specifically for the task of label-ranking, and were optimized specifically for one of the measures considered (whereas Dependency-LDA was not optimized with respect to any specific measure, or even with the specific task of label-ranking in mind).

\subsection*{Comparison With Published scores on Yahoo! Datasets}

\begin{figure}[ht] % float placement: (h)ere, page (t)op, page (b)ottom, other (p)age
  \centering
  \includegraphics[width=.85\linewidth]{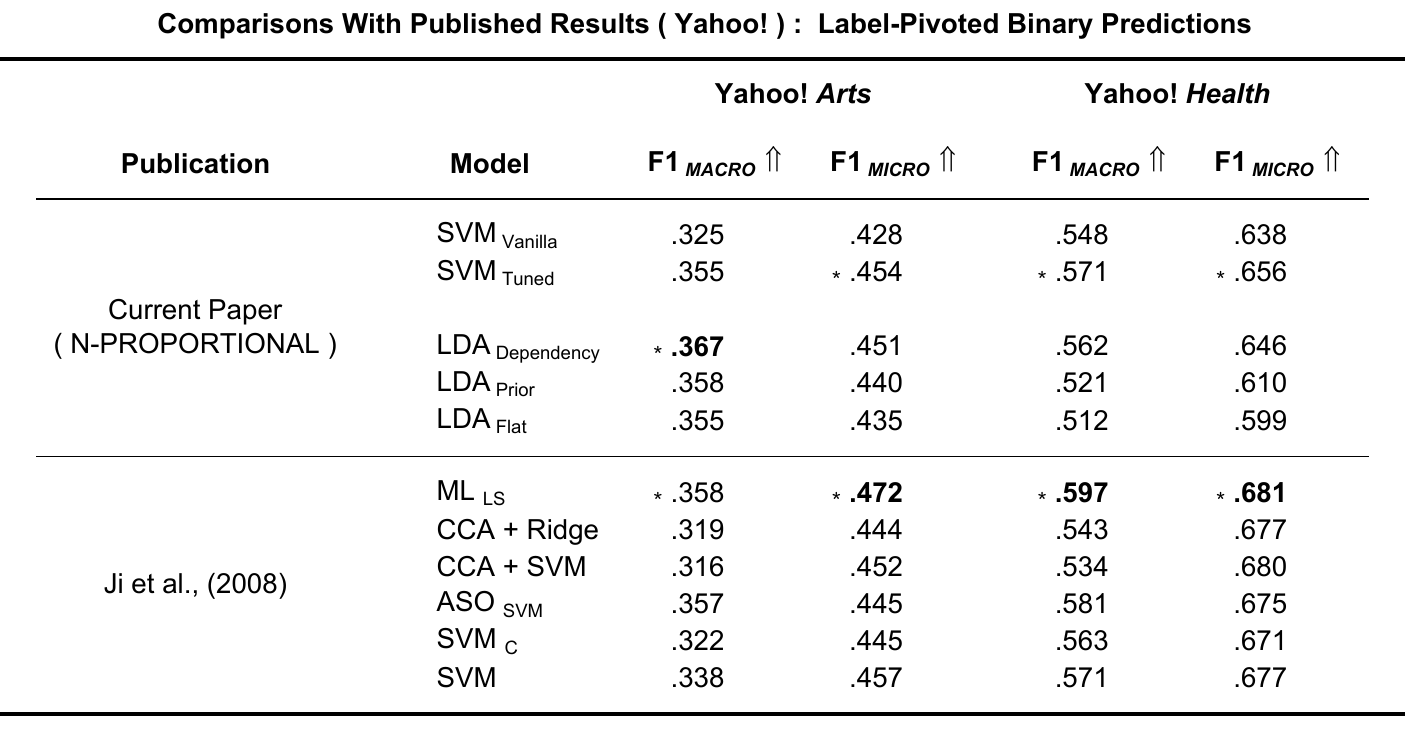}
  \caption{Comparison of Macro-F1 and Micro-F1 scores for the models utilized in the current paper with previously published results from ~\citet{JiTangYuYe_2008}.}
  \label{fig:ComparisonWithPublished_Binary_Yahoo}
\end{figure}

To the best of our knowledge, the only paper which has been published using an equivalent version of the Yahoo! {\it Arts} and {\it Health} datasets is ~\citet{JiTangYuYe_2008}.  Note that numerous additional papers have been published using this dataset, but most of these have used different sets of train-test splits, or used a different number of labels ~\citep[e.g.,][]{UedaSaito_2002,FanLin_2007}\footnote{The version we used had some of the infrequent labels removed from the dataset, and had exactly $1,000$ training documents in each of the five train-test splits.}.  In Figure~\ref{fig:ComparisonWithPublished_Binary_Yahoo} we compare our results on the Yahoo! subdirectory datasets with the numerous discriminative methods presented in ~\citet{JiTangYuYe_2008}.  For complete details on all the algorithms from ~\citet{JiTangYuYe_2008}, we refer the reader to their paper.  However, we note that our $\mathrm{SVM_{VANILLA}}$ and $\mathrm{SVM_{TUNED}}$ methods are essentially equivalent to their $\mathrm{SVM_C}$ and $\mathrm{SVM}$ methods, respectively.  Additionally, the Multi-Label Least Squares ($\mathrm{ML_{LS}}$) method introduced in their paper, uses a discriminative approach for accounting for label-dependencies.

First, we note that the results from our own SVM scores are quite similar to the SVM scores from ~\citet{JiTangYuYe_2008}, which serves to demonstrate that the discriminative classification method we have used throughout the paper for comparison with LDA methods is competitive with similar methods that have been presented in the literature. The $\mathrm{ML_{LS}}$ method that they introduced in the paper outperforms all SVMs, as well as the additional methods that they considered, on all scores.

Performance of the LDA-based methods was generally worse than the best discriminative method ($\mathrm{ML_{LS}}$) presented in ~\citet{JiTangYuYe_2008}.  However, on the Yahoo! {\it Arts} dataset, Dependency-LDA outperformed all methods on the Macro-F1 scores (which, as a reminder, emphasizes the performance on the less frequent labels), and Prior-LDA performed as good as the best discriminative method.  On the Micro-F1 scores for Yahoo! {\it Arts}, Dependency-LDA performance was slightly worse than the $\mathrm{CCA+SVM}$ and tuned SVM methods, and was clearly worse than the $\mathrm{ML_{LS}}$ method, but did outperform the other three discriminative methods.  On the Yahoo! {\it Health} dataset---which has fewer labels, and more training data per label than the {\it Arts} dataset---Dependency-LDA fared worse relative to the discriminative methods.  Dependency-LDA scored better than or similarly to just three of the six methods for Macro-F1 scores, and was beaten by all methods for the Micro-F1 scores.

We note that, although overall performance the LDA-based methods is generally worse than it is for the best discriminative methods on the two Yahoo! datasets, this provides additional evidence that even on non power-law datasets, the LDA-based approaches show a particular strength in terms of performance on infrequent labels (as evidenced by the relatively good Macro-F1 scores for Dependency-LDA).  Furthermore, on these types of datasets, depending on the evaluation metrics being considered, and the exact statistics of the dataset, the Dependency-LDA method is in some cases competitive with or even better than SVMs and more advanced discriminative methods.

\begin{figure}[ht] % float placement: (h)ere, page (t)op, page (b)ottom, other (p)age
  \centering
  \includegraphics[width=.75\linewidth]{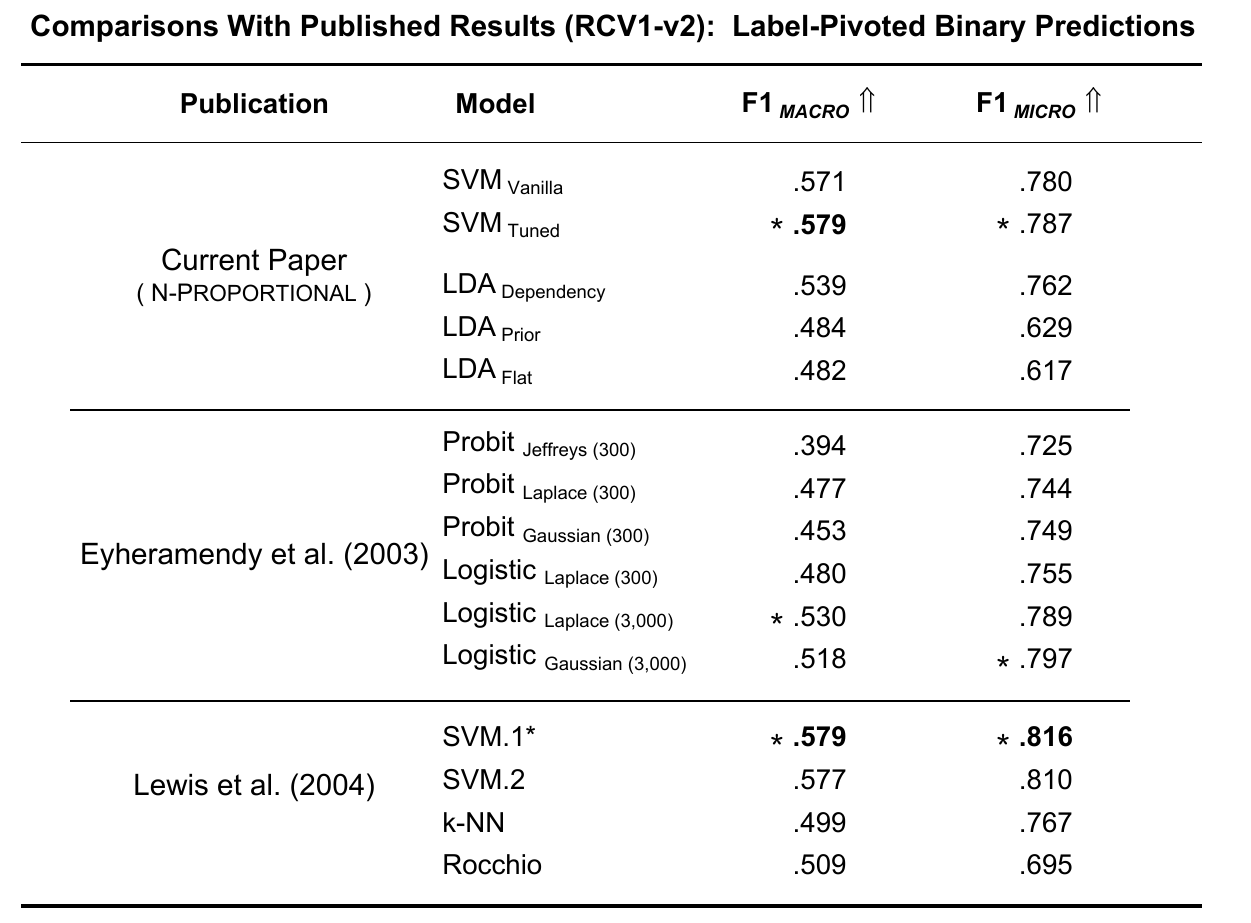}
  \caption{Comparison of Macro-F1 and Micro-F1 scores for the models utilized in the current paper with previously published results.}
  \label{fig:ComparisonWithPublished_Binary_RCV1}
\end{figure}

\subsection*{Comparison With Published scores on RCV1-v2 Datasets}

The RCV1-v2 dataset is a common multi-label benchmark, and numerous results on this dataset can be found in the literature.  We chose to compare with results from both \citet{Lewis_Etc_2004} and \citet{Eyheramendy_2003} since this provides us with a very wide range of algorithms for comparison (where the former paper considers several of the most popular discriminative classification methods, and the latter paper considers numerous Bayesian style regression methods).  Note that the Macro-F1 and Micro-F1 scores for the {\it SVM-1} algorithm presented in \citet{Lewis_Etc_2004} were the result of two distinct sets of predictions (where one set of SVM predictions was thresholded to optimize Micro-F1, and a separate set of predictions were optimized for Macro-F1).  Since all other methods presented in Figure~\ref{fig:ComparisonWithPublished_Binary_RCV1} (as well as throughout our paper) used a single set of predictions to compute all scores, we re-computed the Macro-F1 scores using the predictions optimized for Micro-F1\footnote{These were re-computed from the confusion matrices made available in the online appendix to their paper}, in order to be consistent across all results. The {\it SVM-1} algorithm nonetheless is tied for the best Macro-F1 score (with our own SVM results) and achieves the best Micro-F1 score overall.

In terms of the LDA-based methods, the Dependency LDA model clearly performs worse than SVMs on RCV1-v2.  However, it outperforms all non-SVM methods on Macro-F1 (including methods from both \citet{Lewis_Etc_2004} and \citet{Eyheramendy_2003}).  It additionally achieves a Micro-F1 score that is competitive with most of the non-SVM methods (although it is significantly worse than most logistic-regression methods, in addition to SVMs).

\vfill\eject

\bibliography{MLJ_Refs_V22}

\end{document}